%% file: iclr2025_conference.tex
\documentclass{article} 
\usepackage{iclr2025_conference,times}
\usepackage{graphicx}
\input{math_commands.tex}

\usepackage{hyperref}
\usepackage{url}

\usepackage{tikz}
\usepackage{booktabs}   
\usepackage{array}      
\usepackage{colortbl}
\usepackage{multirow}
\usepackage{enumitem}

\definecolor{visualcolor}{RGB}{230,240,255}
\definecolor{headercolor}{RGB}{70,130,180}
\definecolor{lightblue}{RGB}{240,248,255}
\definecolor{strengthcolor}{RGB}{230,255,230}
\definecolor{improvementcolor}{RGB}{255,240,230}
\definecolor{twitterblue}{RGB}{29,161,242}
\definecolor{instapink}{RGB}{225,48,108}
\definecolor{facebookblue}{RGB}{66,103,178}
\definecolor{emailgray}{RGB}{128,128,128}
\definecolor{persona1color}{RGB}{230,240,255}
\definecolor{persona2color}{RGB}{255,240,230}
\definecolor{persona3color}{RGB}{230,255,230}

\newenvironment{compactitemize}{
    \begin{itemize}[
        leftmargin=*,
        nosep,
        before=\vspace{-0.5\baselineskip},
        after=\vspace{-0.5\baselineskip}
    ]
}{\end{itemize}}

\definecolor{headercolor}{RGB}{70,130,180}
\definecolor{rowcolor}{RGB}{240,248,255}

\title{
    Agentic Multimodal AI for Hyper-Personalized B2B and B2C Advertising in  
    Competitive Markets: An AI-Driven Competitive Advertising Framework
}

\author{
  Sakhinana Sagar Srinivas\thanks{Equal contribution.} \thanks{Corresponding author.} \\
  Tata Research \\
  Bangalore, India \\
  \texttt{sagar.sakhinana@tcs.com} \\
  \And
  Akash Das\footnotemark[1] \\
  Tata Research \\
  Bangalore, India \\
  \texttt{akash@tcs.com} \\
  \And
  Shivam Gupta \\
  Tata Research \\
  New Delhi, India \\
  \texttt{shivam.gupta@tcs.com} \\
  \And
  Venkataramana Runkana \\
  Tata Research \\
  Pune, India \\
  \texttt{venkat.runkana@tcs.com} \\
}

\iclrfinalcopy 
\begin{document}

\vspace{-3mm}
\maketitle

\begin{abstract}
\vspace{-3mm}
The increasing deployment of foundation models (FMs) in real-world applications necessitates strategies to enhance their adaptivity, reliability, and efficiency in dynamic market environments. In the chemical industry, AI-discovered materials are driving innovation in new chemical products, but their commercial success depends on effective market adoption. This requires FM-driven advertising frameworks capable of operating in-the-wild, adapting to diverse consumer segments and competitive market conditions. We introduce an AI-driven, multilingual, multimodal framework that leverages foundation models for autonomous, hyper-personalized, and competitive advertising in both Business-to-Business (B2B) and Business-to-Consumer (B2C) markets. By integrating retrieval-augmented generation (RAG), multimodal reasoning, and adaptive persona-based targeting, our framework generates culturally relevant and market-aware advertisements tailored to dynamic consumer behaviors and competitive landscapes. Our approach is validated through a combination of real-world experiments using actual product data and a Simulated Humanistic Colony of Agents to model consumer personas and optimize ad strategies at scale while maintaining privacy compliance. This ensures market-grounded and regulatory-compliant advertising. Synthetic experiments are designed to mirror real-world scenarios, enabling the testing and optimization of advertising strategies by simulating market conditions, consumer behaviors, and product scenarios. This approach helps companies avoid costly real-world A/B tests while ensuring privacy compliance and scalability, allowing them to refine strategies through simulations before actual deployment. By combining structured retrieval-augmented reasoning with in-context learning (ICL) for adaptive ad generation, the framework enhances engagement, prevents market cannibalization, and optimizes Return on Ad Spend (ROAS). This work presents a scalable FM-driven solution that bridges AI-driven novel product innovation and market adoption, advancing the deployment of multimodal, in-the-wild AI systems for high-stakes decision-making environments such as commercial marketing.
\vspace{-3mm}
\end{abstract}

\section{Introduction}
\vspace{-3mm}
In recent years, generative AI has revolutionized materials discovery. For instance, MatterGen has enabled the design of novel materials, while MatterSim validates their performance under real-world conditions \cite{zeni2023mattergen, yang2024mattersim}. Complementing these efforts, autonomous labs like A-Lab integrate AI with robotics to propose and execute synthesis recipes, accelerating the discovery and development of innovative materials for applications such as energy storage and sustainability \cite{burger2024accelerating}. These materials form the foundation of a wide range of chemical products, with applications spanning fast-moving consumer goods (FMCG), the semiconductor industry, the electronics sector, and beyond. Recent advancements in AI-driven materials discovery, autonomous synthesis, and process automation \cite{zeni2023mattergen, yang2024mattersim, burger2024accelerating}, coupled with progress in scaling the production of chemical products from simulations and laboratory experiments to large-scale industrial manufacturing \cite{srinivas2024accelerating, gowaikar2024agentic}, have paved the way for commercialization. However, a significant bottleneck remains in translating these innovations into market adoption. Success depends not only on technical performance but also on stakeholders' ability to effectively communicate their value to diverse audiences—ranging from consumers and investors to regulatory bodies. The rapid evolution of digital platforms has transformed advertising, compelling businesses to adopt AI-driven programmatic advertising to stay competitive. Chemical product manufacturers increasingly leverage Demand-Side Platforms (DSPs) to participate in Real-Time Bidding (RTB) auctions across both business-to-business (B2B) and business-to-consumer (B2C) channels. The primary objective is to maximize Return on Ad Spend (ROAS) by optimizing ad visibility across various platforms. This optimization spans search engines (Google Ads, Bing Ads), where advertisers engage in Open Auctions or negotiate Programmatic Guaranteed deals; social media platforms such as LinkedIn for B2B engagement and Instagram/TikTok for consumer outreach; and e-commerce marketplaces like Amazon, Alibaba, and Knowde. At the core of these campaigns is RTB, a process where publishers (sellers) offer individual ad impressions for sale through Supply-Side Platforms (SSPs), while advertisers (buyers) compete to purchase these impressions via DSPs in real-time auctions. These auctions often follow a second-price bidding model or header bidding, ensuring that advertisers optimize ROAS while publishers maximize their ad revenue. SSPs facilitate auctions by passing user and page-level data—including geolocation, device type, search intent, and potentially Mobile Ad IDs (MAIDs)—to DSPs, ensuring compliance with privacy regulations such as GDPR and CCPA. To refine targeting, advertisers integrate first-party data, such as website visits, purchase history, and CRM records; third-party audience intelligence sourced via Data Management Platforms (DMPs), which provide insights into demographics, interests, and behaviors; and real-time contextual data, such as webpage content, device type, and live user intent, ensuring ads appear at the most relevant moments. These elements help DSPs optimize bids, enhancing targeting precision, while Ad Exchanges determine the winning bid. A critical factor in RTB success is Click-Through Rate (CTR) prediction, which selects the most relevant ad creative (often HTML5-based rich media) for display. This personalized approach enhances user engagement and mitigates ad fatigue, ensuring ads remain relevant, engaging, and likely to drive conversions. RTB platforms reward high-CTR ads with lower cost-per-click (CPC), reducing acquisition costs and improving conversion efficiency for advertisers. Given this dynamic, AI-driven Dynamic Creative Optimization (DCO) plays a crucial role in enabling hyper-personalized advertising, a key driver of ROAS improvement. DCO dynamically adjusts visuals, messaging, and calls-to-action based on user behavior, engagement metrics, and real-time contextual data, ensuring that ads resonate deeply with each audience segment. By leveraging DCO techniques, advertisers enhance hyper-personalization, drive higher engagement and conversions, and ultimately maximize ROAS. Despite advances in AI, existing advertising systems often fail to leverage contextual multimodal user behavior insights, limiting their effectiveness in competitive product marketing, where multimodal, multilingual, and persona-specific ad targeting is critical. Advertising multiple products within the same category—such as consumer goods—from different manufacturers presents additional challenges, including avoiding cannibalization (where ads for one product negatively impact another), tailoring messages to diverse user preferences, and emphasizing each product’s unique selling points (USPs), such as superior features, pricing, or brand perception. Moreover, creating culturally relevant ads for global audiences while addressing competition within similar product categories from different manufacturers remains a significant hurdle. To address these gaps, we propose an automated, data-driven, multilingual, and multimodal framework capable of delivering hyper-personalized, competitive advertisements tailored to diverse consumer personas and dynamic market conditions. This framework integrates multimodal data analysis by leveraging advanced large foundational models and persona-specific targeting to ensure ads resonate across cultural contexts and stakeholder groups. To validate this approach, we conduct experiments in both real-world and synthetic settings. Real-world experiments involve using data from actual companies, including their product information, advertising strategies, and consumer insights. To model target consumer behavior based on this real-world data, we leverage a Simulated Humanistic Colony of Agents instead of conducting expensive or complex human trials, allowing us to efficiently evaluate and refine advertising strategies before market launch. In contrast, synthetic experiments involve creating controlled environments to simulate various market conditions and using a Simulated Humanistic Colony of Agents to model personas (consumer segments) for personalized ad optimization of competing products from different manufacturers. This approach enables the systematic evaluation of ad strategies at scale while ensuring regulatory compliance (e.g., GDPR) and avoiding the costs and complexities associated with real-world data collection. By combining the robustness of real-world validation with the flexibility and scalability of synthetic testing, these experiments serve as both a proof of concept and a demonstration of the framework's ability to generate competitive, impactful advertisements. In real-world settings, the comprehensive framework integrates three interconnected systems: (1) The Multimodal Agentic Advertisement Market Survey (MAAMS) system (see Figure~\ref{fig:Figure1}), led by a Meta-Agent, combines insights from publicly available sources—such as social media, financial databases, and market research tools—to evaluate brand sentiment, visual identity, emotional engagement, financial performance, and market trends. This system generates a comprehensive overview of a product's market position, providing a detailed, multimodal analysis of how different companies present and position their chemical products. (2) The Personalized Market-Aware Targeted Advertisement Generation (PAG) system (see Figures~\ref{fig:Figure2}-\ref{fig:Figure3}) builds on MAAMS to create tailored, multilingual ads for diverse consumer personas. Using a Simulated Humanistic Colony of Agents, it aligns ads with specific user preferences. The Adv Curator Agent designs personalized ads, while the Social Media Agent optimizes them for platforms like Twitter and Instagram. (3) The Competitive Hyper-Personalized Advertisement System (CHPAS) (see Figure~\ref{fig:Figure4}) takes the personalized ads generated by PAG and enhances their competitiveness by differentiating them for competing products from different manufacturers. It strategically emphasizes unique selling points—such as affordability, functionality, and quality—to ensure relevance and engagement for target audiences. Together, these systems leverage advanced agentic architectures powered by LLMs to deliver data-driven, adaptive, and impactful advertising strategies. For ad evaluations, we utilize a multi-faceted approach that assesses clickability rates and ad quality through automated reward models, LLM evaluators, and human evaluators. For ad copy optimization, we evaluate various hyper-personalized versions using simulated human personas. This iterative, persona-driven testing and evaluation process helps maximize engagement and conversion rates. Additionally, we utilize Open-Domain Question Answering (ODQA) with Retrieval-Augmented Generation (RAG) to answer specific questions by searching and analyzing the market intelligence gathered by the MAAMS system on the product's market position. Furthermore, we conduct synthetic experiments, offering a powerful tool for advertisers to develop, test, and optimize AI-driven advertisement generation frameworks in a cost-effective and scalable manner. These experiments provide a controlled environment for precise testing, ensure inherent privacy compliance, enhance framework robustness by exposing it to a wide range of scenarios (including rare edge cases), and accelerate innovation—all while minimizing the risks associated with real-world testing. The synthetic, data-driven, hypothetical framework for optimizing multi-product, persona-specific advertising integrates seven key components: market research data, persona profiling, product analysis, competitive analysis, ad generation using foundational models, ad platform optimization, and ad evaluation. We discuss this framework in detail in the technical appendix. The experimental results demonstrate the potential of AI-driven, multimodal frameworks to transform advertising, providing businesses with a powerful tool to navigate modern marketing challenges.  This work establishes a foundation for future research in AI-driven advertising, offering a scalable solution for industries seeking to bridge product innovation and market communication. The paper is organized as follows: Section~\ref{sec:proposed_method} details our proposed method, Section~\ref{sec:experiments} describes the experimental setup, Section~\ref{sec:results} presents results and analysis, and Section~\ref{sec:conclusion} concludes with key findings and future directions.

 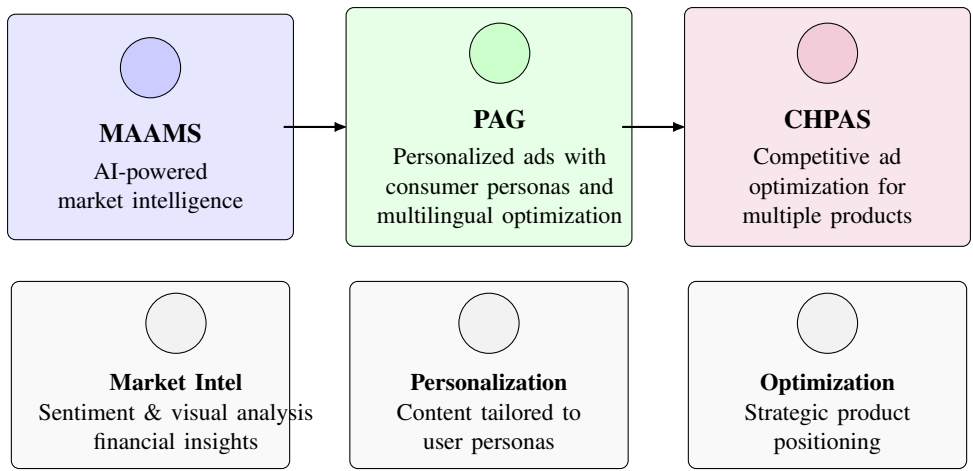
\begin{figure*}[ht!]
\centering
\begin{tikzpicture}[
    node distance=4cm,
    box/.style={
        draw,
        rounded corners=3pt,
        minimum width=3.6cm,
        minimum height=3cm,
        text width=3.4cm,
        align=center,
        inner sep=0.2cm
    },
    smallbox/.style={
        draw,
        rounded corners=3pt,
        minimum width=3.6cm,
        minimum height=2cm,
        text width=3.4cm,
        align=center,
        inner sep=0.15cm
    },
    arrow/.style={
        ->,
        >=latex,
        thick,
        shorten >=2pt,
        shorten <=2pt
    },
    icon/.style={
        circle,
        draw,
        minimum size=0.8cm,
        inner sep=0.1cm
    }
]

\node[box, fill=blue!10] (maams) at (-4.5,1.8) {
    \begin{tabular}{c}
        \tikz \node[icon, fill=blue!20] {}; \\[0.2cm]
        \textbf{MAAMS} \\[0.1cm]
        \small AI-powered  \\
        \small market intelligence 
    \end{tabular}
};

\draw[arrow] (-2.8,1.8) -- (-1.8,1.8);

\node[box, fill=green!10] (pag) at (0,1.8) {
    \begin{tabular}{c}
        \tikz \node[icon, fill=green!20] {}; \\[0.2cm]
        \textbf{PAG} \\[0.1cm]
        \small Personalized ads with \\
        \small consumer personas and \\
        \small multilingual optimization
    \end{tabular}
};

\draw[arrow] (1.7,1.8) -- (2.7,1.8);

\node[box, fill=purple!10] (chpas) at (4.5,1.8) {
    \begin{tabular}{c}
        \tikz \node[icon, fill=purple!20] {}; \\[0.2cm]
        \textbf{CHPAS} \\[0.1cm]
        \small Competitive ad \\
        \small optimization for \\
        \small multiple products
    \end{tabular}
};

\node[smallbox, fill=gray!5] (market) at (-4.5,-1.5) {
    \begin{tabular}{c}
        \tikz \node[icon, fill=gray!10] {}; \\[0.1cm]
        \small \textbf{Market Intel} \\
        \small Sentiment \& visual analysis \\
        \small financial insights
    \end{tabular}
};

\node[smallbox, fill=gray!5] (personalization) at (0,-1.5) {
    \begin{tabular}{c}
        \tikz \node[icon, fill=gray!10] {}; \\[0.1cm]
        \small \textbf{Personalization} \\
        \small Content tailored to \\
        \small user personas
    \end{tabular}
};

\node[smallbox, fill=gray!5] (competitive) at (4.5,-1.5) {
    \begin{tabular}{c}
        \tikz \node[icon, fill=gray!10] {}; \\[0.1cm]
        \small \textbf{Optimization} \\
        \small Strategic product \\
        \small positioning
    \end{tabular}
};

\end{tikzpicture}
\caption{AI framework architecture showing the main systems (MAAMS, PAG, CHPAS) and supporting features for chemical product advertising optimization.}
\label{fig:AIframework}
\vspace{-2mm}
\end{figure*}

\vspace{-3mm}
\section{Proposed Method}
\label{sec:proposed_method}
\begin{figure*}[htbp] 
    \centering 
    \includegraphics[
        width=1.0\textwidth, 
        height=1.00\textheight,  
        keepaspectratio, 
        trim=0mm 0mm 0mm 0mm, 
        clip  
    ]{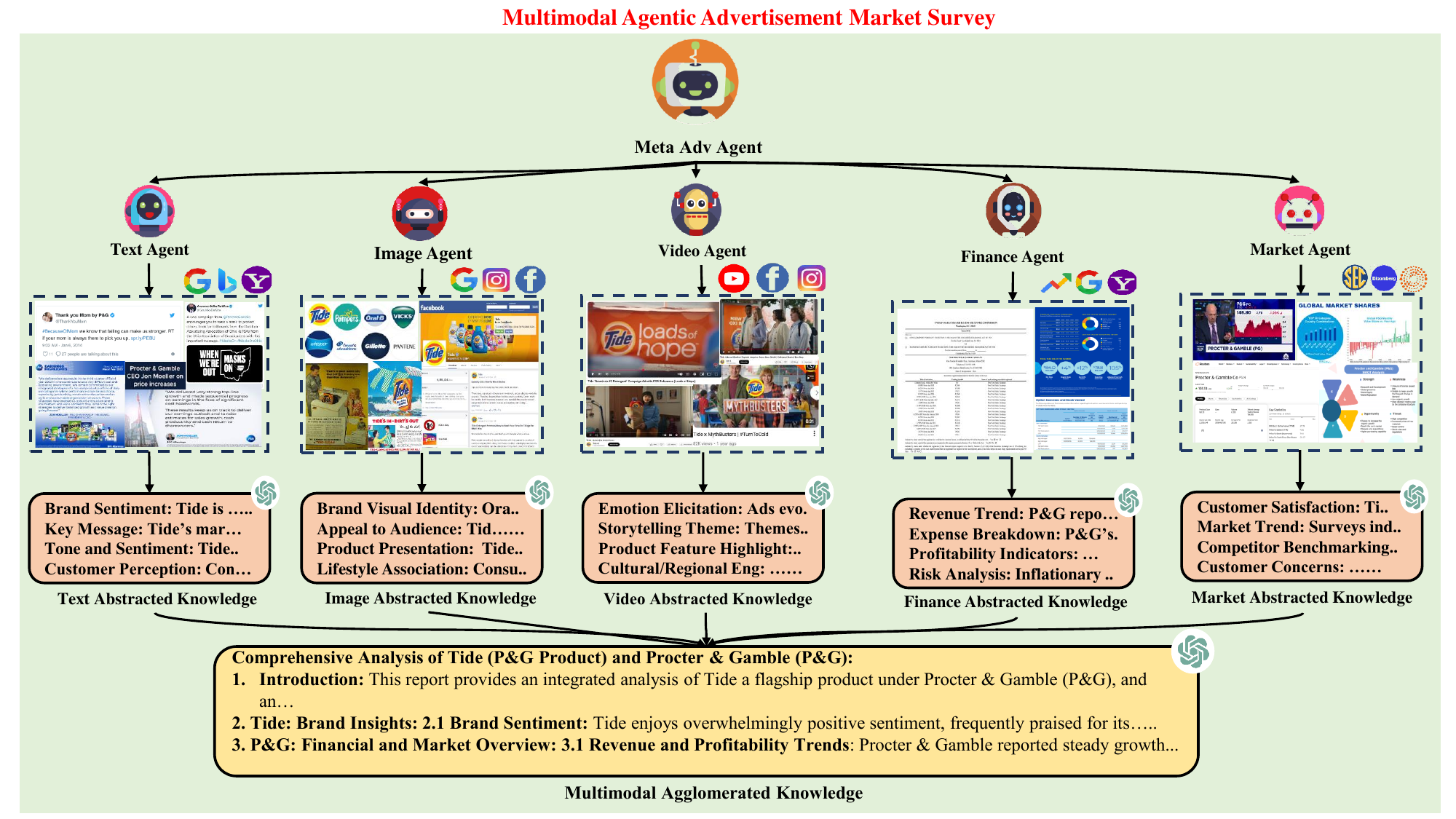} 
    \vspace{-6mm}
    \caption{The MAAMS system, led by a Meta-Agent, collects data from Text, Image, Video, Finance, and Market agents to analyze brand sentiment, visual identity, emotional engagement, financial performance, and market trends. It compiles these insights into a comprehensive report on a product’s overall performance, marketing effectiveness, financial health, and market standing.
    } 
    \label{fig:Figure1} 
    \vspace{-5mm}
\end{figure*}

\vspace{-5mm}
This section presents a comprehensive framework for multimodal analysis and personalized advertisement generation in the chemical industry (see Figure \ref{fig:AIframework}). The framework consists of three interconnected systems: (a) The Multimodal Agentic Advertisement Market Survey (MAAMS) system, which integrates insights across various modalities to assess product sentiment, branding, engagement, profitability, and competitive positioning. (b) The Personalized Market-Aware Targeted Advertisement Generation (PAG) system, designed to create tailored, multilingual advertisements for diverse consumer personas. (c) The Competitive Hyper-Personalized Advertisement System (CHPAS), which generates differentiated advertisements for competing products, ensuring that unique selling points are highlighted for specific user preferences. These systems employ advanced agentic architectures powered by Large Language Models (LLMs) to deliver data-driven, adaptive, and impactful advertising strategies.  The MAAMS system, orchestrated by a meta-agent, systematically gathers insights into a brand’s market presence, performance, and perception across multiple modalities. It employs specialized agents—Text, Image, Video, Finance, and Market—each powered by LLMs to retrieve and analyze multimodal data, synthesizing a comprehensive understanding of product positioning and consumer engagement. As illustrated in Figure \ref{fig:Figure1}, these agents extract insights to evaluate brand sentiment, visual identity, emotional engagement, financial performance, and market trends. The Text Agent, powered by LLMs, utilizes SerpAPI to retrieve data from search engines such as Google, Bing, and Yahoo, analyzing brand sentiment, customer perception, and messaging strategies. By leveraging advanced natural language processing, LLMs extract insights from unstructured text data, generating structured summaries of key findings. Additionally, the system assesses critical aspects such as chemical compositions, safety regulations (e.g., REACH, GHS labeling), and compliance messaging, ensuring that advertisements align with regulatory standards and industry best practices. The Image Agent, enhanced by LLMs, extracts visual data from platforms like Instagram and Pinterest to evaluate visual identity, audience appeal, and lifestyle associations. LLMs interpret visual content such as logos, packaging, hazard symbols, and barcodes, ensuring both aesthetic effectiveness and regulatory compliance. They generate descriptive summaries that connect visual elements to textual context and brand messaging. The Video Agent, supported by LLMs, aggregates content from platforms like YouTube and TikTok to assess emotional appeal, storytelling themes, product highlights, and cultural relevance. By processing video transcripts, subtitles, and metadata, LLMs provide nuanced insights into how safety protocols, environmental impact, and sustainability practices are communicated, evaluating the effectiveness of video advertisements. The Finance Agent, utilizing LLMs, gathers financial data from platforms like Bloomberg and Yahoo Finance, analyzing revenue trends, expense breakdowns, profitability indicators, and risk factors. Through financial reports, earnings calls, and market research, LLMs evaluate financial performance and the cost-effectiveness of advertising strategies, generating actionable insights. The Market Agent, powered by LLMs, collects market data from sources such as Statista and Google Trends, focusing on customer satisfaction, market trends, competitor benchmarking, and consumer concerns. By processing survey results, customer feedback, and competitor data, LLMs identify emerging trends, measure customer loyalty, and provide deep insights into consumer behavior. The Meta-Agent, leveraging LLMs as its core intelligence layer, integrates insights from all specialized agents into a unified knowledge base. It synthesizes textual, visual, financial, and market data to generate a comprehensive report that aligns with consumer expectations, regulatory requirements, and industry standards. This unified knowledge directly supports two key systems: (a) The PAG system for creating personalized multilingual advertisements. (b) The CHPAS system for generating differentiated ads for competing products. The unified knowledge is particularly valuable as it combines traditional market data with specialized chemical industry requirements, such as safety regulations and compliance standards (e.g., REACH, GHS labeling).

\vspace{-4mm}

\begin{figure*}[htbp] 
    \centering 
    \includegraphics[
        width=1.0\textwidth, 
        height=1.00\textheight,  
        keepaspectratio, 
        trim=0mm 0mm 0mm 0mm, 
        clip  
    ]{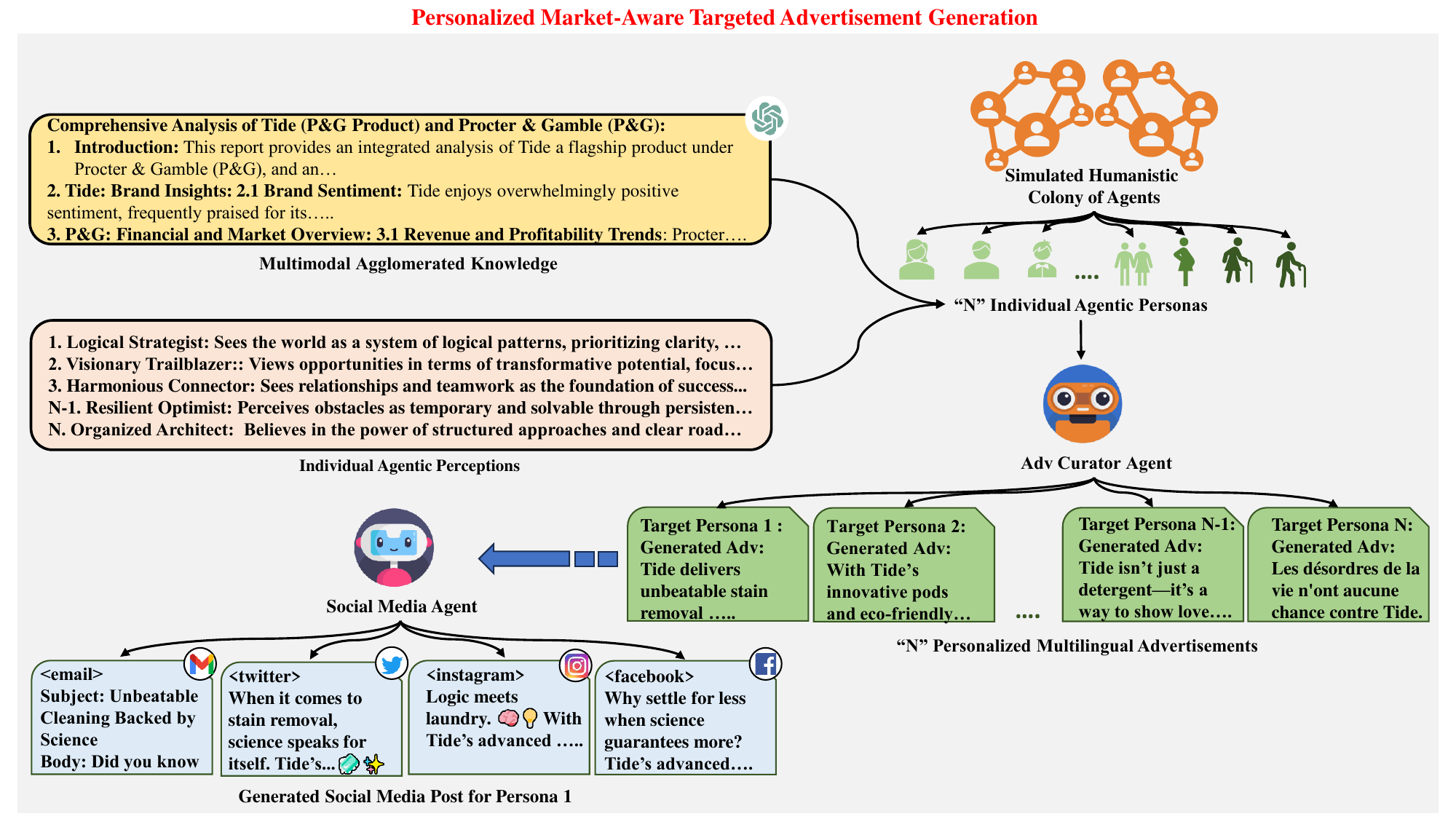} 
    \vspace{-6mm}
    \caption{The figure illustrates the workflow of the PAG system, designed to generate personalized, multilingual advertisements tailored to diverse consumer personas. Leveraging multimodal agglomerated knowledge, a simulated humanistic colony of agents mimics consumer personas (e.g., Logical Strategist, Visionary Trailblazer) to align ads with specific preferences, such as innovation, sustainability, or emotional connections. The Adv Curator Agent creates personalized multilingual advertisements, ensuring cultural and linguistic appropriateness for global relevance and engagement. The Social Media Agent further optimizes these ads for platforms like Twitter, Instagram, and Facebook, maximizing their impact across diverse audiences.} 
    \label{fig:Figure2} 
    \vspace{-2mm}
\end{figure*}

\begin{figure*}[htbp] 
    \centering 
    \includegraphics[
        width=1.00\textwidth, 
       height=1.00\textheight,  
        keepaspectratio, 
        trim=0mm 0mm 0mm 0mm, 
        clip  
    ]{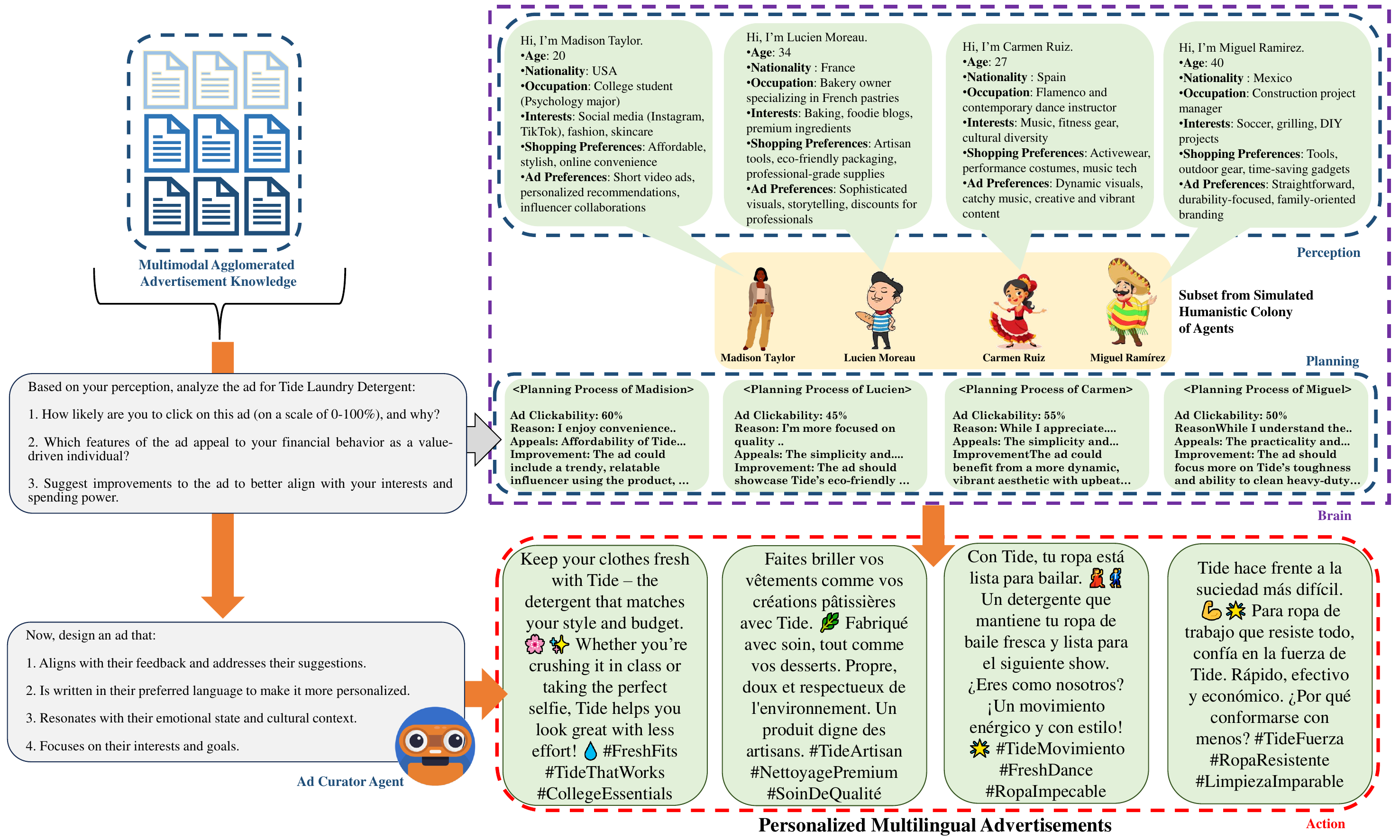} 
    \vspace{-7mm}
    \caption{The figure demonstrates the PAG system for creating customized ads tailored to diverse consumer personas. Using multimodal agglomerated knowledge, the Adv Curator Agent evaluates individual preferences, cultural contexts, and feedback to produce personalized multilingual advertisements for specific consumer segments. This approach ensures that the ads are engaging, emotionally resonant, and effective in driving consumer action, highlighting their value for modern personalized advertising.} 
    \label{fig:Figure3} 
    \vspace{-5mm}
\end{figure*}

\begin{figure*}[htbp] 
    \centering 
    \includegraphics[
        width=1.00\textwidth, 
        height=1.00\textheight,  
        keepaspectratio, 
        trim=0mm 0mm 0mm 0mm, 
        clip  
    ]{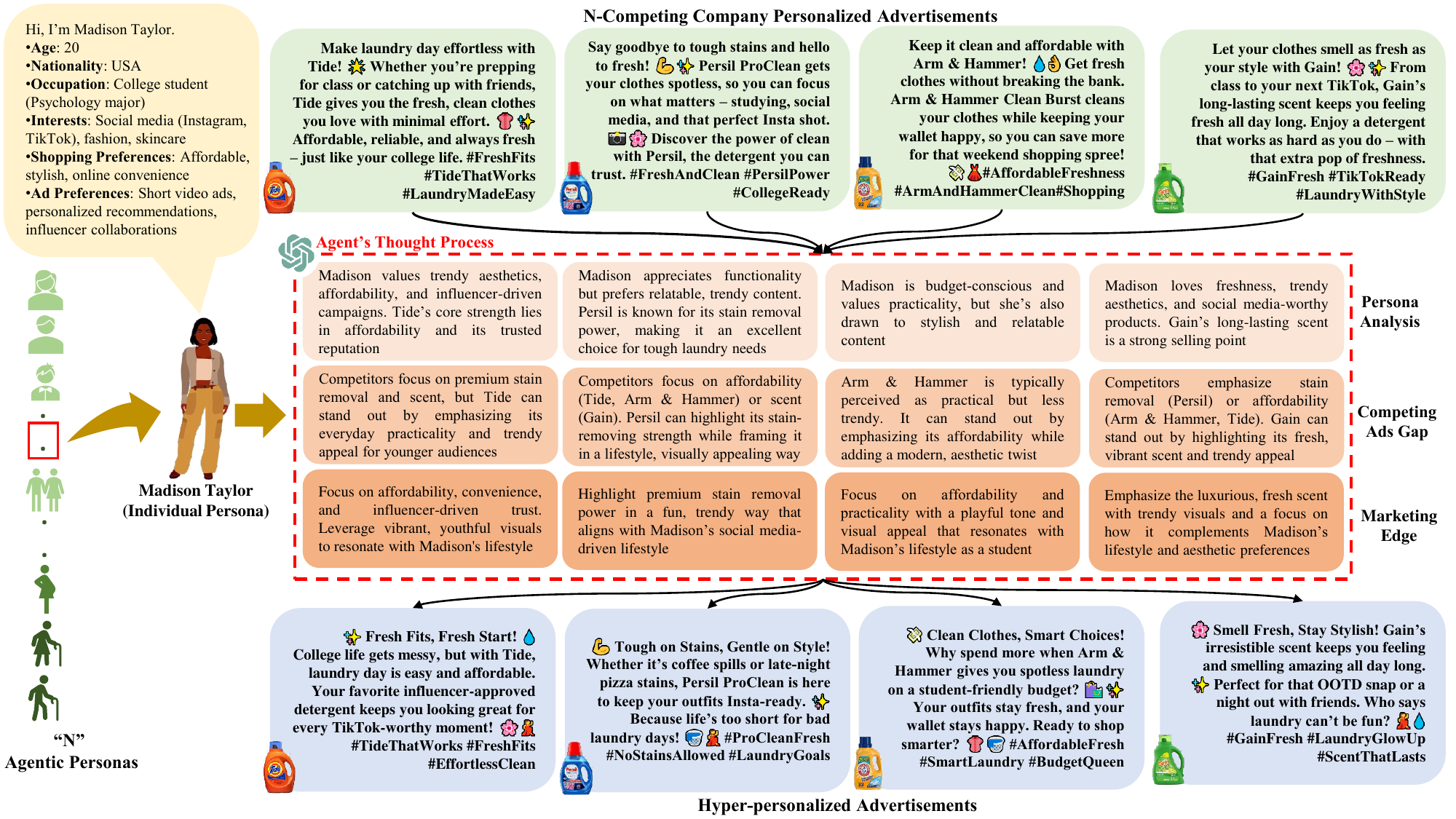} 
    \vspace{-7mm}
    \caption{The figure demonstrates the generation of hyper-personalized advertisements for competing products within the same category from different manufacturers, tailored to a specific consumer persona. Each advertisement emphasizes unique selling points—such as affordability, functionality, freshness, and trendy appeal—while aligning with the persona's preferences, lifestyle, and shopping behaviors. The system strategically highlights each product's competitive advantages, ensuring relevance and engagement for the target audience. This approach showcases the system’s ability to create differentiated advertisements for competing products, effectively positioning each manufacturer’s unique strengths while maintaining personalization and platform optimization.} 
    \label{fig:Figure4} 
    \vspace{-1mm}
\end{figure*}

This integrated approach ensures that the resulting advertisements are not only engaging and culturally relevant but also technically accurate and compliant with industry standards, ultimately supporting the creation of better-targeted and more effective advertising strategies for chemical products. In summary, the MAAMS system provides an automated, multimodal approach to analyzing chemical product advertising and market positioning, delivering deep insights into brand performance, consumer engagement, and regulatory compliance. We present the PAG system, designed to achieve the goal of personalized advertising by delivering highly relevant and engaging messages to individual consumers or specific segments. As illustrated in Figure \ref{fig:Figure2}, the system leverages aggregated knowledge from the MAAMS system at its core. It integrates data on product information—such as unique features, usage instructions, and regulatory compliance—alongside brand sentiment, consumer behavior, and financial performance to create tailored advertising strategies. A simulated humanistic colony of agents enables the emulation of individuals with specific personalities, interests, and goals through diverse individual agentic personas (e.g., Logical Strategist, Visionary Trailblazer, Harmonious Connector; see Figure \ref{fig:Figure2}). These personas represent distinct consumer characteristics and preferences, enabling the creation of highly tailored advertisements. For example: The Logical Strategist values clarity, logic, and evidence-based decision-making, responding well to ads that emphasize scientific innovation and data-driven benefits. The Visionary Trailblazer focuses on transformative potential and innovation, preferring ads that highlight sustainability and eco-friendly practices. The Harmonious Connector values relationships and emotional connections, engaging with ads that evoke family values and community impact. The Resilient Optimist views challenges as solvable and responds to ads that inspire confidence and problem-solving. The Organized Architect prioritizes structure and efficiency, preferring ads that emphasize ease of use and dependability. By leveraging these personas, the system ensures that advertisements are highly relevant, emotionally engaging, and action-oriented, aligning with the unique preferences and values of diverse consumer segments. The Adv Curator Agent generates personalized multilingual advertisements by optimizing ads for the characteristics and preferences of the personas. It ensures that the advertisements are culturally and linguistically appropriate while aligning with each persona's distinct traits. Simultaneously, the Social Media Agent tailors these ads for platforms like Twitter, Instagram, and Facebook, creating platform-specific posts that resonate with the target personas' audiences. Together, these agents ensure that the advertisements are not only personalized but also optimized for diverse cultural contexts and platforms. By combining data-driven insights from the MAAMS system, human-like creativity via multi-agent personas, and platform-specific optimization  the PAG system creates persuasive advertising campaigns to drive consumer purchasing decisions.  The system leverages multimodal agglomerated knowledge to optimize ads for specific segments, emphasizing benefits while building brand image, emotional connections, quality, and reliability. Figure \ref{fig:Figure3} demonstrates the PAG system’s capability to generate tailored, multilingual advertisements that align with consumer preferences, behaviors, and emotional triggers across diverse cultural contexts while maintaining regulatory compliance. Unlike the PAG system, which focuses on product-centric messaging and tailoring ads for specific user personas, the CHPAS system (refer to Figure \ref{fig:Figure4}) is designed for competitive personalized ad optimization across multiple products from different manufacturers within the same category. For example, it can effectively promote competing chemical products such as laundry detergents from brands like Tide, Persil, Arm \& Hammer, and Gain by tailoring advertisements to individual users' preferences, demographics, and behaviors. Leveraging personalized advertisements from the PAG system (see Figures \ref{fig:Figure2}-\ref{fig:Figure3}), the CHPAS system creates competitive personalized ads that are tailored for specific user personas while strategically highlighting each product’s competitive advantages. The system ranks products for each persona according to affinity and competitive strengths, ensuring that the most suitable products are highlighted to each user. For example, for a college student interested in affordability and trendy aesthetics, Tide might emphasize convenience and influencer-driven trust, while Gain could focus on long-lasting freshness and social media appeal. By tailoring ads to individual preferences and strategically positioning each product’s unique selling points, the system avoids cannibalization and maximizes ad effectiveness. This results in a data-driven, persona-centric strategy that delivers highly personalized, competitive, and high-performing advertisements. By aligning ads with user preferences and optimizing for engagement, CHPAS enhances consumer action and brand loyalty across diverse product categories.

\vspace{-2mm}
\section{Experiments}
\vspace{-1mm}
\label{sec:experiments}
To validate the effectiveness of our proposed framework, we conducted a series of experiments in both real-world and synthetic settings. These experiments were designed to evaluate the performance of our multimodal, AI-driven systems in generating impactful, personalized advertisements. We discuss synthetic data experiments in the technical appendix. Below, we describe the datasets, evaluation metrics, experimental setup, and key findings.

\vspace{-2mm}
\subsection{Datasets}
\vspace{-1mm}
Developing hyper-personalized and competitive advertisements for chemical products requires addressing two key challenges: (a) Modeling diverse consumer behaviors at scale while ensuring compliance with privacy regulations such as GDPR. (b) Accurately representing chemical products’ technical specifications, safety requirements, and value propositions within regulatory constraints. To overcome these challenges, our framework integrates two complementary data sources: (a) A Simulated Humanistic Colony of Agents, which serves as a privacy-compliant, AI-driven model of real-world consumer behavior. (b) Real-world company and product category data, ensuring AI-generated advertisements remain market-grounded, commercially competitive, and regulatory-compliant. The simulated colony enables the PAG and CHPAS systems to test, optimize, and refine advertisements across diverse consumer segments, product categories, and competitive market conditions while maintaining strict privacy compliance. It does so by simulating consumers along four key behavioral dimensions: (i) Occupational diversity: Covers office support, management, sales, healthcare, education, and engineering (Figure \ref{fig:agent_demographics}), ensuring advertisements align with profession-specific purchasing decisions, industry needs, and professional priorities. 
(ii) Emotional state diversity: Agents exhibit a spectrum of psychological states, ranging from neutral to complex emotions (Figure \ref{fig:emotional_state}), allowing AI to evaluate advertisement resonance, engagement levels, and psychological impact. (iii) Multilingual representation: Supports English, Spanish, Asian, European, and Middle Eastern languages (Figure \ref{fig:language_distribution}), enabling global cultural adaptability, linguistic optimization, and localized messaging strategies. (iv) Socioeconomic stratification: Categorizes agents into lower, middle, and upper income classes (Figure \ref{fig:socioeconomic_class}), facilitating targeted ad messaging based on spending behavior, price sensitivity, and purchasing power. By integrating these demographic, emotional, linguistic, and financial variables, the simulated colony provides privacy-compliant, synthetic behavioral insights, ensuring AI-generated advertisements are precisely targeted, engagement-driven, and adaptable to dynamic market conditions. Complementing this behavioral simulation, our framework incorporates real-world company and product category data, capturing competitive positioning, market dynamics, and regulatory constraints. The dataset spans 50+ globally recognized FMCG companies and 2,000+ products, encompassing household and personal care items, specialty cleaning products, cosmetics, and many others. This integration enables advertisements to be aligned with real-world industry benchmarks, ensuring technically accurate product representation, competitive differentiation, and optimized pricing strategies. The MAAMS system serves as the central intelligence layer, aggregating market intelligence—such as brand sentiment, financial performance, competitive analysis, and regulatory compliance insights—to provide a data-driven foundation for PAG and CHPAS. By leveraging retrieval-augmented multimodal analysis, the framework synthesizes market intelligence with AI-driven behavioral modeling, ensuring AI-generated advertisements are hyper-personalized, commercially competitive, linguistically and culturally adaptable, and technically robust. This dual-source approach, combining synthetic consumer behavior modeling and real-world market intelligence, allows businesses to refine advertising strategies at scale, delivering personalized, engagement-driven, and compliance-aware advertisements while eliminating privacy risks and reducing the high costs associated with traditional consumer trials.

\vspace{-4mm}

\vspace{-2mm}

\vspace{2mm}
\section{Evaluation Metrics}
\vspace{-1mm}
We evaluated our framework using two key metrics: Clickability Rates and Ad Quality Scores. Clickability Rates measure engagement levels by calculating the ratio of ad clicks to impressions for initial, personalized, and hyper-personalized advertisements (Figure \ref{fig:clickability_rates}). Ad Quality Scores assess advertisement quality across five dimensions: Helpfulness (usefulness and actionable information), Correctness (accuracy and factual consistency), Coherence (logical flow and clarity), Complexity (level of detail and sophistication), and Verbosity (conciseness and comprehensiveness). These scores are derived from three evaluation methods: Reward Model Scoring, using models such as NVIDIA Nemotron-4-340b-reward; LLM-as-Judge, where large language models such as GPT-4o evaluate ads based on predefined criteria; and Human Evaluation, providing a benchmark assessment (Figure \ref{fig:product_comparison_scores}). Together, these metrics ensure that advertisements are engaging, accurate, clear, well-structured, and appropriately detailed.

\vspace{-3mm}
\section{Results and Analysis}
\label{sec:results}
\begin{figure*}[htbp]
    \centering  
    \includegraphics[width=0.8\textwidth,trim=0 0 0 30,clip]{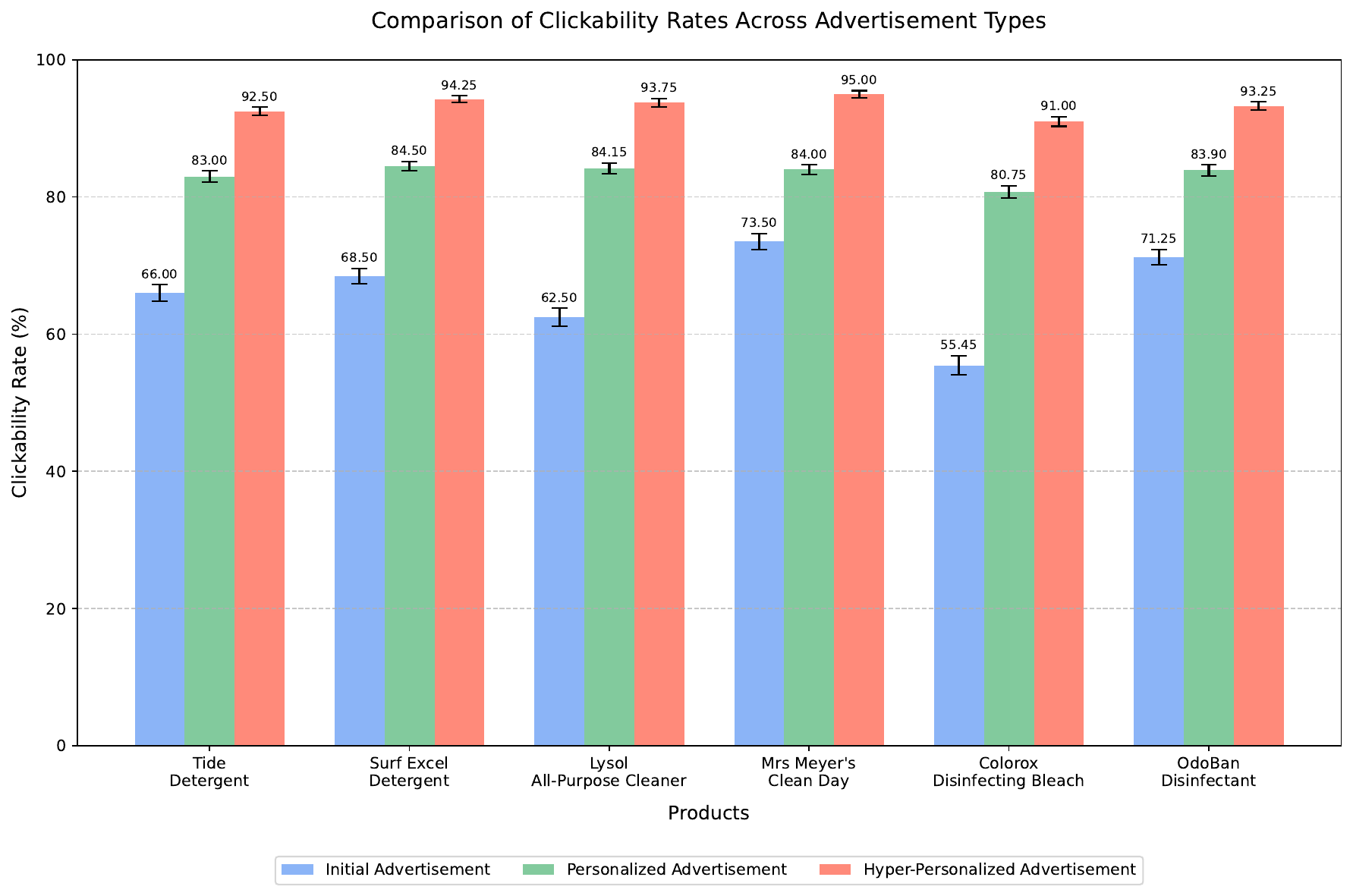}
    \vspace{-2mm}
    \caption{Comparison of average clickability rates across different advertisement types for various products. The advertisement types include initial, personalized, and hyper-personalized ads, with clickability rates calculated across multiple consumer personas within the simulated humanistic colony of agents. Hyper-personalized advertisements consistently achieve the highest clickability rates, followed by personalized ads, while initial ads show the lowest engagement. Error bars indicate variability in the data across different personas, highlighting the effectiveness of tailored advertising strategies in driving consumer engagement.}
    \vspace{-3mm}
    \label{fig:clickability_rates}
\end{figure*}

\begin{figure*}[htbp]
    \centering 
    \includegraphics[width=0.815\textwidth,trim=0 0 0 30,clip]{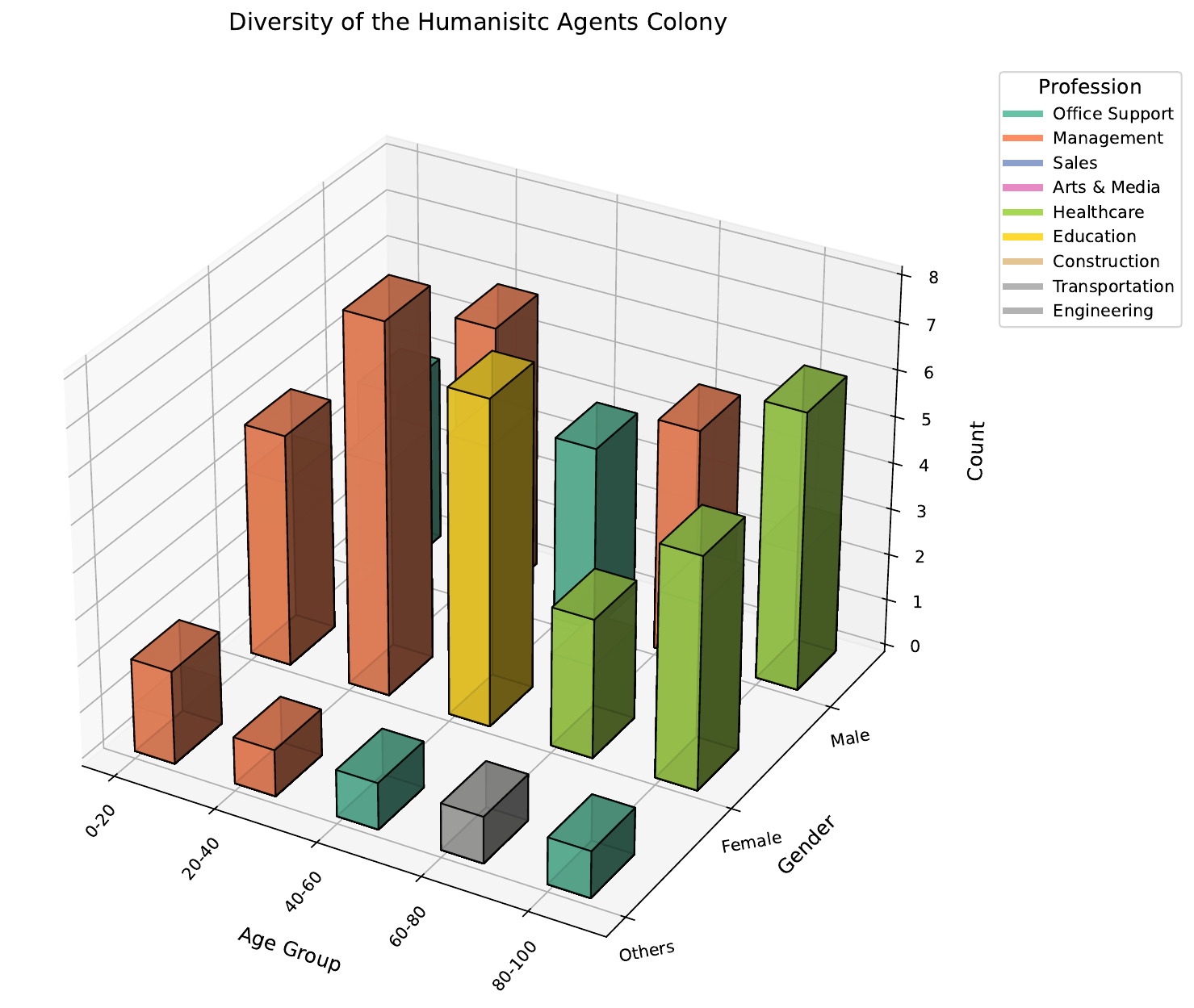}
    \vspace{-5mm}
    \caption{Demographic Distribution of Humanistic Agents by Profession: This figure illustrates the demographic diversity of agents within the Humanistic Agents Colony, focusing on the intersection of age group, gender, and profession. The colony spans a wide range of professions, including Office Support, Management, and others, ensuring broad representation for consumer research and marketing. This diversity enables the generation of highly tailored advertisements and marketing content targeted at various occupational backgrounds, ensuring relevance across different professional contexts and improving the precision of outreach efforts.}
    \label{fig:agent_demographics}
    \vspace{-12mm}
\end{figure*}

\begin{figure*}[htbp]
    \centering   
    \includegraphics[width=0.665\textwidth,trim=0 0 0 45,clip]{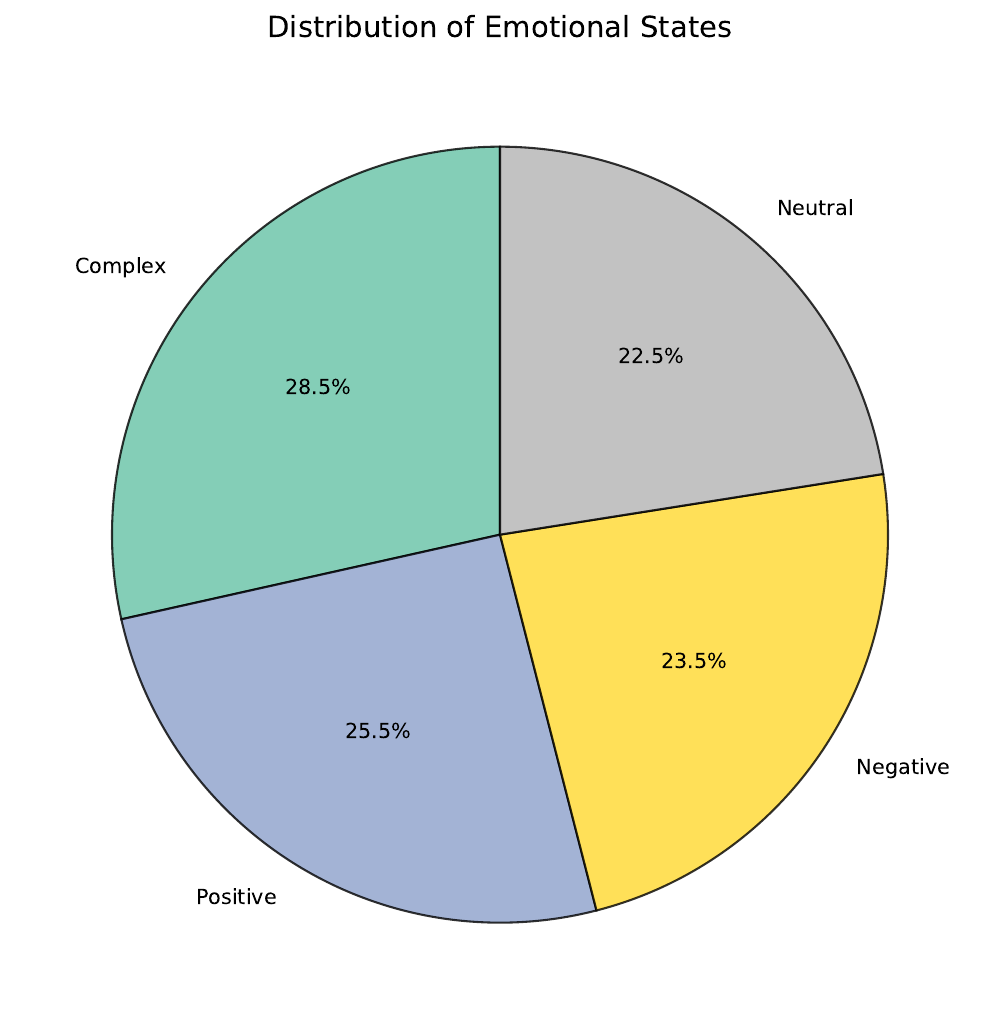}
    \vspace{-12mm}
    \caption{The figure displays the distribution of four distinct emotional states—Complex, Positive, Negative, and Neutral—observed within a group of simulated agents. Understanding these varied emotional profiles is essential for evaluating advertising campaign effectiveness across a range of psychological responses. By assessing ad performance against agents exhibiting these emotional states, we can refine messaging and build campaigns that resonate deeply with audiences, aligning with their emotional contexts, from simple neutrality to complex emotional experiences.}
    \label{fig:emotional_state}
\end{figure*}

\begin{figure*}[htbp]
    \centering      
    \includegraphics[width=0.675\textwidth,trim=0 0 0 30,clip]{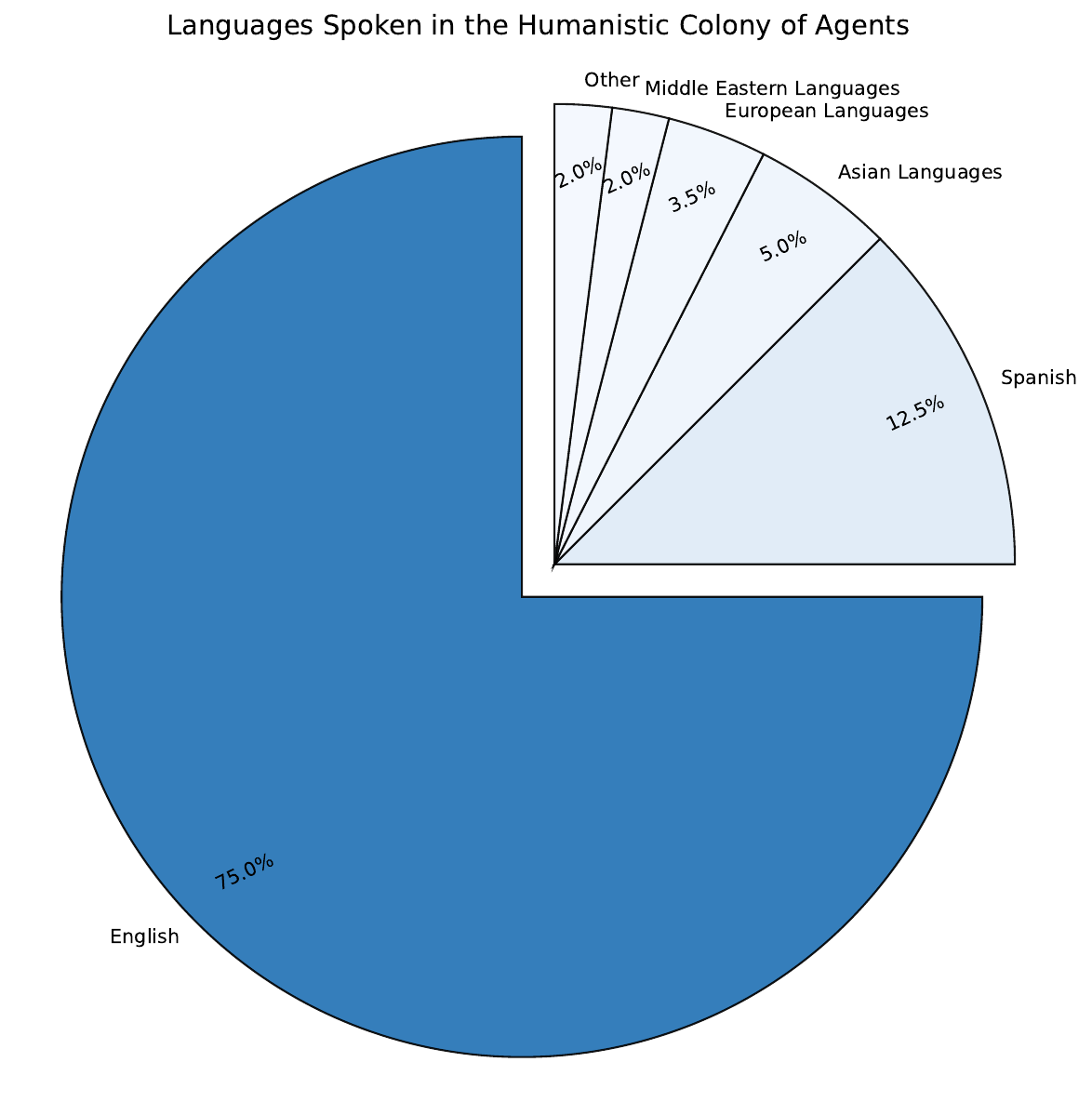}
    \vspace{-5mm}
    \caption{The figure depicts the distribution of languages spoken by agents in the Humanistic Colony. The personas represent a multilingual population, with English as the dominant language ($75.0\%$), followed by Spanish ($12.5\%$), and smaller representations of Asian languages ($5.0\%$), European languages ($3.5\%$), Middle Eastern languages ($2.0\%$), and other languages ($2.0\%$), ensuring adaptability for global advertising campaigns. This linguistic diversity supports the creation of culturally relevant messaging for diverse audiences.
}
    \label{fig:language_distribution}
    \vspace{-20mm}
\end{figure*}

\begin{figure*}[htbp]
    \centering        
    \includegraphics[width=0.675\textwidth,trim=0 0 0 50,clip]{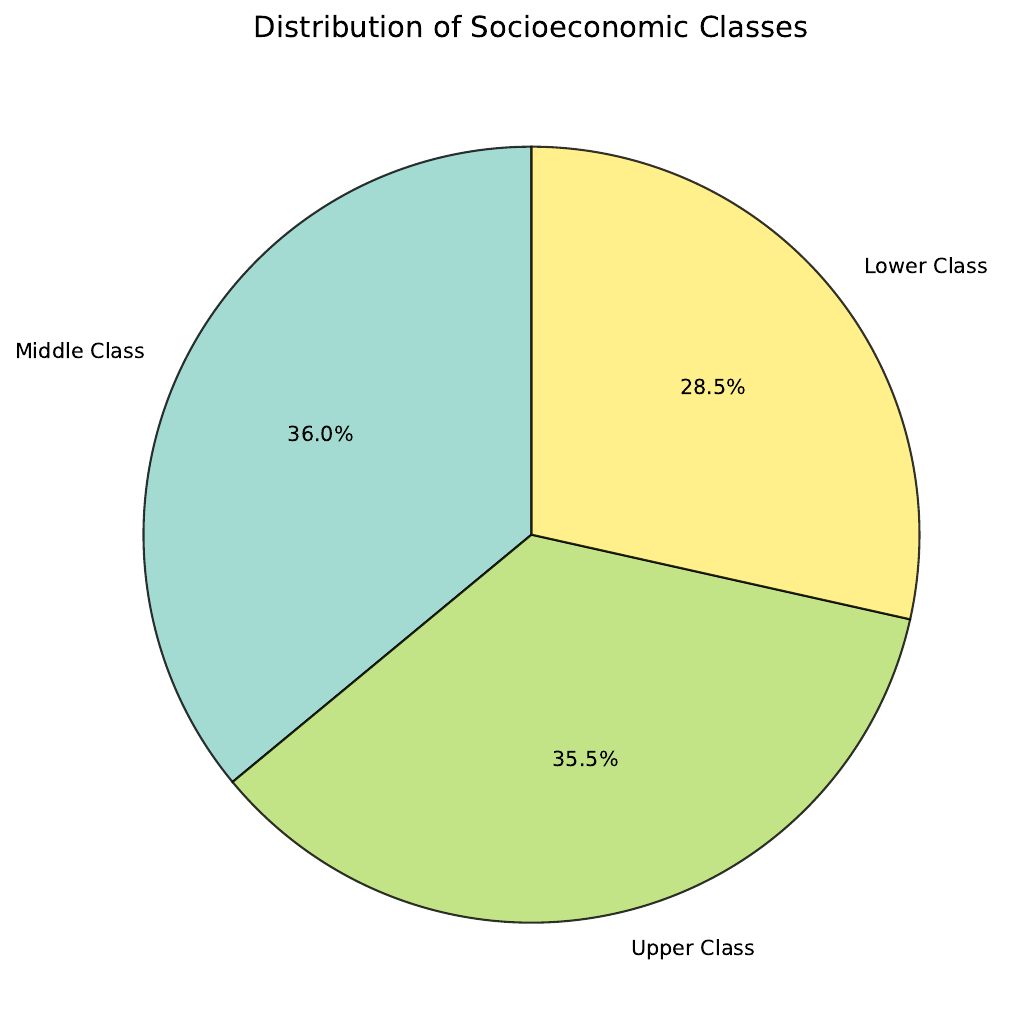}
    \vspace{-10mm}
    \caption{The figure depicts the socioeconomic class distribution of agents. The personas are segmented into Middle Class (36\%), Upper Class (35.5\%), and Lower Class (28.5\%), ensuring broad representation across various financial backgrounds. This segmentation allows for targeted messaging, enabling advertisements to resonate with diverse financial priorities and spending behaviors across different income levels.}
    \label{fig:socioeconomic_class}
\end{figure*}

\begin{figure*}[htbp]
    \centering  
    \includegraphics[width=0.75\textwidth,trim=0 5 0 20,clip]{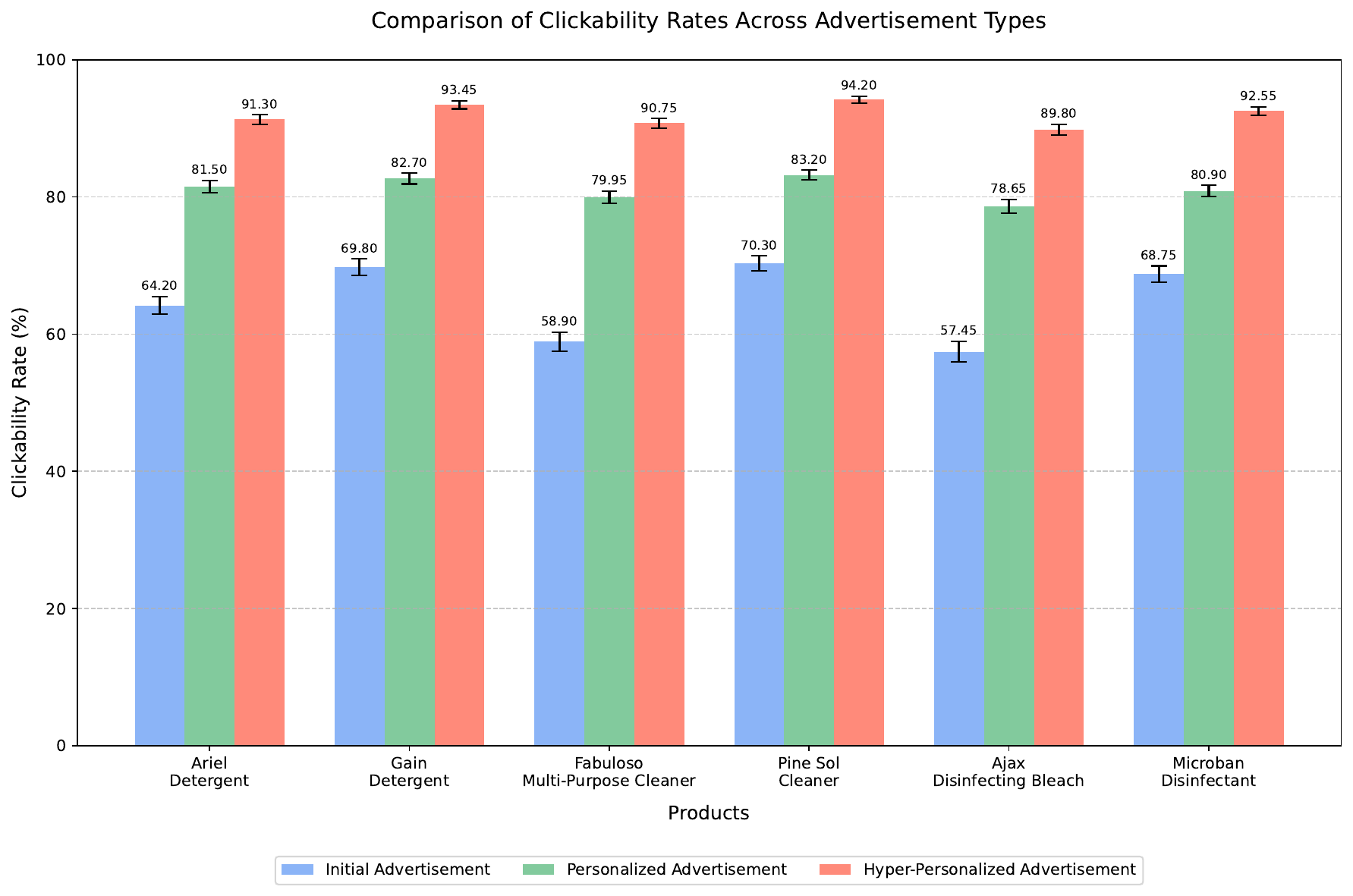}
    \vspace{-2mm}
    \caption{Comparison of clickability rates across advertisement types for household cleaning products. The figure illustrates the average clickability rates (in percentages) for three advertisement types—initial, personalized, and hyper-personalized—calculated across multiple consumer personas within the simulated humanistic colony of agents. The products analyzed include Ariel Detergent, Gain Detergent (Laundry Detergents), Fabuloso Multi-Purpose Cleaner, Pine-Sol Cleaner (Multi-Purpose Cleaners), Ajax Disinfecting, and Microban Disinfectant (Antibacterial Cleaner). The results demonstrate that hyper-personalized advertisements consistently achieved the highest clickability rates, followed by personalized ads, while initial ads showed the lowest engagement. Error bars indicate variability in the data across different personas.}    
    \label{fig:clickability_rates_new_products}
    \vspace{-6mm}
\end{figure*}

\begin{figure*}[htbp]
    \centering  
    \includegraphics[width=0.875\textwidth,trim=0 10 0 20,clip]{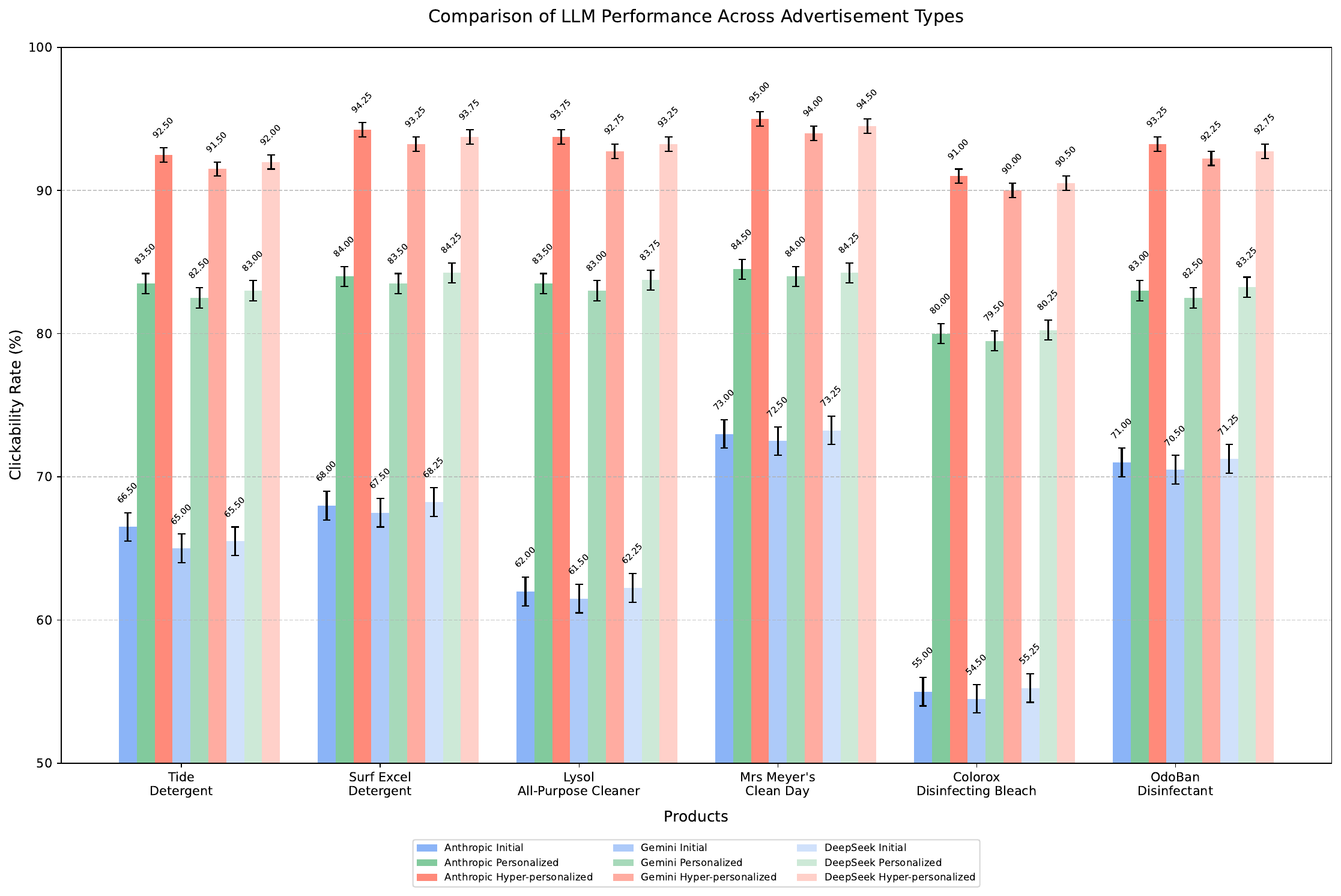}
    \vspace{-2mm}
    \caption{This figure compares the average clickability rates across three AI models—Anthropic, Gemini, and DeepSeek—for initial, personalized, and hyper-personalized advertisements across seven different product categories: Tide Detergent, Surf Excel Detergent (Laundry Detergents), Lysol All-Purpose Cleaner, Mrs. Meyer’s Clean Day (Multi-Purpose Cleaners), Clorox Disinfecting Bleach, and OdoBan Disinfectant (Antibacterial Cleaner). Clickability rates are calculated as the average across multiple consumer personas within the simulated humanistic colony of agents. The results indicate that hyper-personalized ads consistently achieve the highest engagement across all AI models, followed by personalized ads, while initial ads perform the worst. The relatively small differences between AI models suggest that the level of personalization is a more significant factor in ad effectiveness than the specific model used. Clickability rates generally range from the low 60\% to 95\%, with variation between products.}
    \label{fig:llm_ad_comparison}
\end{figure*}

\begin{figure*}[htbp]
    \centering
    \includegraphics[
        width=1.25\textwidth,
        trim=75mm 0mm 0mm 20mm, 
        clip
    ]{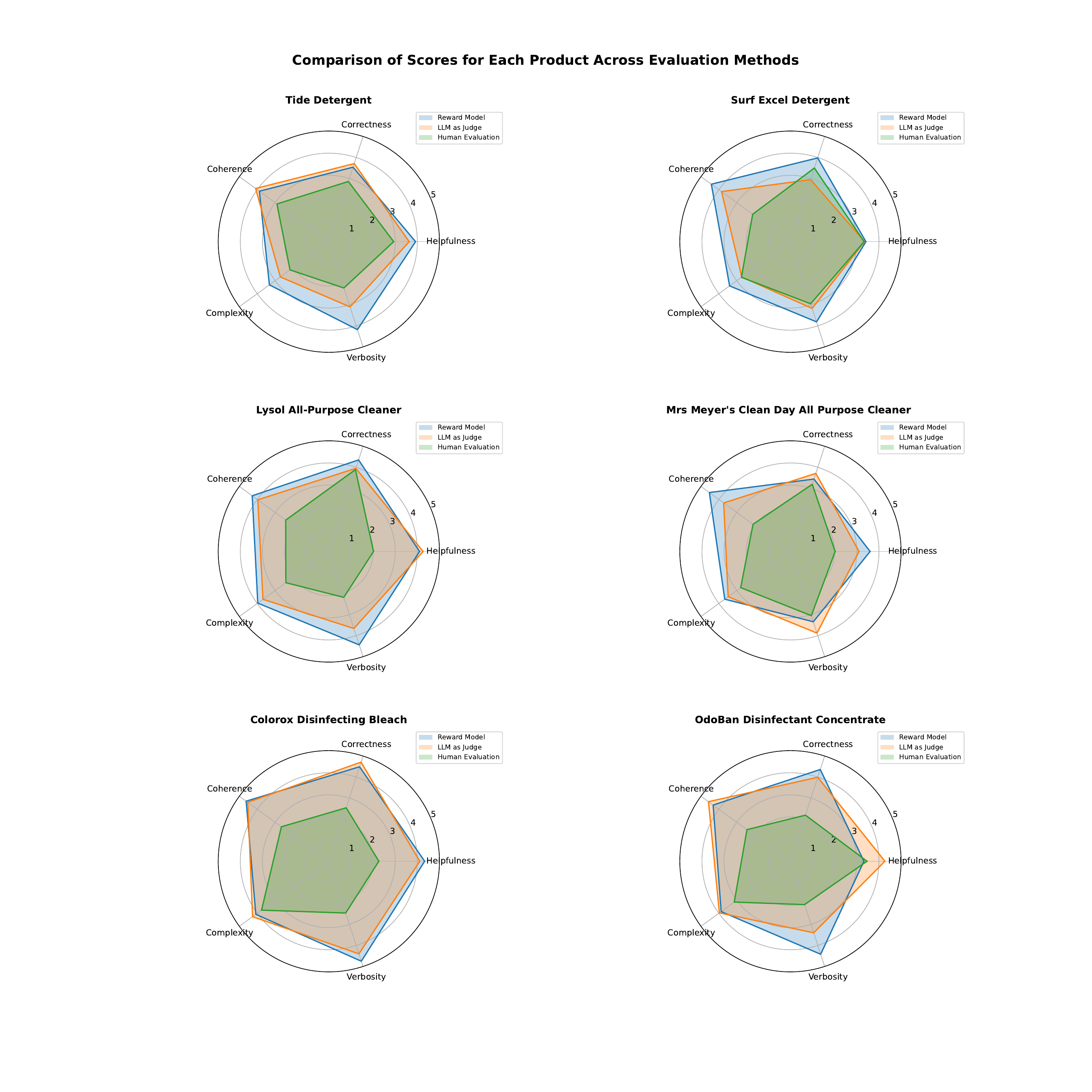}
    \vspace{-25mm}
    \caption{Evaluation of AI-generated advertisements for cleaning and detergent products, including Tide Detergent, Surf Excel Detergent (Laundry Detergents), Lysol All-Purpose Cleaner, Mrs. Meyer’s Clean Day All-Purpose Cleaner (Multi-Purpose Cleaners), Clorox Disinfecting Bleach, and OdoBan Disinfectant Concentrate (Antibacterial Cleaner). 
    The advertisements are assessed using three evaluation methods: Reward Model Scoring, LLM-as-Judge Evaluation, and Human Evaluation. Each product is evaluated across five key metrics: Helpfulness, Correctness, Coherence, Complexity, and Verbosity. The figure highlights variations in scores across products and evaluation methods, illustrating how AI-generated advertisements perform in different aspects of clarity, engagement, and informativeness. These insights support the optimization of competitive, persona-specific advertisements tailored to consumer preferences.}
    \label{fig:product_comparison_scores}
\end{figure*}

\vspace{-2mm}
The experimental results highlight the potential of AI-driven, multimodal frameworks to revolutionize the advertising landscape. Figures \ref{fig:clickability_rates} and \ref{fig:clickability_rates_new_products} illustrate clickability rates, which measure the percentage of times users click on an advertisement when it is displayed. This metric is computed as the ratio of clicks to ad impressions. The figures compare clickability rates for three advertisement types—initial, personalized, and hyper-personalized—across different product categories. Specifically, Figure \ref{fig:clickability_rates} evaluates clickability rates for household cleaning and detergent brands such as Tide, Surf Excel, Lysol, Mrs. Meyer’s Clean Day, Clorox, and OdoBan, while Figure \ref{fig:clickability_rates_new_products} presents results for Ariel, Gain, Fabuloso, Pine-Sol, Ajax, and Microban. Across all products, hyper-personalized advertisements consistently achieved the highest clickability rates, followed by personalized ads, while initial ads recorded the lowest engagement. For instance, in Figure \ref{fig:clickability_rates}, hyper-personalized ads for Tide Detergent reached a clickability rate of approximately 92.5\%, compared to 83.0\% for personalized ads and 66.0\% for initial ads. A similar pattern is observed in Figure \ref{fig:clickability_rates_new_products}, where hyper-personalized ads for Ariel Detergent achieved a clickability rate of 91.3\%, outperforming personalized ads at 81.5\% and initial ads at 64.2\%. These results demonstrate that generic, untargeted advertisements, or initial ads, are significantly less effective in capturing consumer interest compared to persona-specific, AI-driven strategies. Figure \ref{fig:llm_ad_comparison} extends this analysis by comparing clickability rates across three AI models—Anthropic, Gemini, and DeepSeek—for each advertisement type. The results show a consistent trend: hyper-personalized ads perform best across all models, followed by personalized ads, while initial ads remain the least effective. Notably, the differences between AI models are marginal, suggesting that the degree of personalization is a more significant factor in ad effectiveness than the specific AI model used. For instance, across different cleaning products, the clickability rates of hyper-personalized ads generated by Anthropic, Gemini, and DeepSeek remain within a close range, reinforcing the conclusion that hyper-personalization is a key driver of engagement, regardless of the underlying AI model. These findings underscore the importance of AI-driven, persona-specific advertising strategies in modern digital marketing. By tailoring advertisements to individual consumer preferences, businesses can significantly enhance engagement, optimize advertising expenditures, and improve overall market competitiveness. In addition, radar charts (Figures \ref{fig:product_comparison_scores} and \ref{fig:product_comparison_scores2}) compare evaluation scores for six cleaning products—Tide Detergent, Surf Excel Detergent (Detergents), Lysol All-Purpose Cleaner, Mrs. Meyer’s Clean Day All-Purpose Cleaner (Multi-Purpose Cleaners), Clorox Disinfecting Bleach, and OdoBan Disinfectant Concentrate (Antibacterial Cleaner)—with Ariel Detergent, Gain Detergent (Laundry Detergents), Fabuloso Multi-Purpose Cleaner, Pine-Sol Cleaner (Multi-Purpose Cleaners), Ajax Disinfecting, and Microban Disinfectant (Antibacterial Cleaner)—across five dimensions: Helpfulness, Correctness, Coherence, Complexity, and Verbosity. Each product’s performance is assessed using Reward Model Scoring, LLM-as-Judge evaluations, and Human Evaluation methods on a scale of 1 to 5, revealing distinct patterns between automated metrics and human judgment.

\vspace{-2mm}

\vspace{-3mm}
\section{Conclusion}
\label{sec:conclusion}
\vspace{-2mm}
Our framework demonstrates the transformative potential of AI-driven, multimodal systems in revolutionizing personalized advertising for competitive markets. By integrating advanced market analysis, persona-specific targeting, and competitive differentiation, we enable the creation of hyper-personalized advertisements that enhance consumer engagement. The results validate the framework's scalability and compliance with privacy regulations through real-world experiments with simulated persona behavior. Future work could extend this offline framework to real-time operations, enabling dynamic optimization based on live market data and consumer interactions. This work paves the way for data-driven, AI-powered advertising in the evolving digital landscape.

\bibliography{iclr2025_conference}
\bibliographystyle{iclr2025_conference}

\begin{figure*}[htbp]
    \centering
    \includegraphics[
        width=1.25\textwidth,
        trim=75mm 0mm 0mm 20mm, 
        clip
    ]{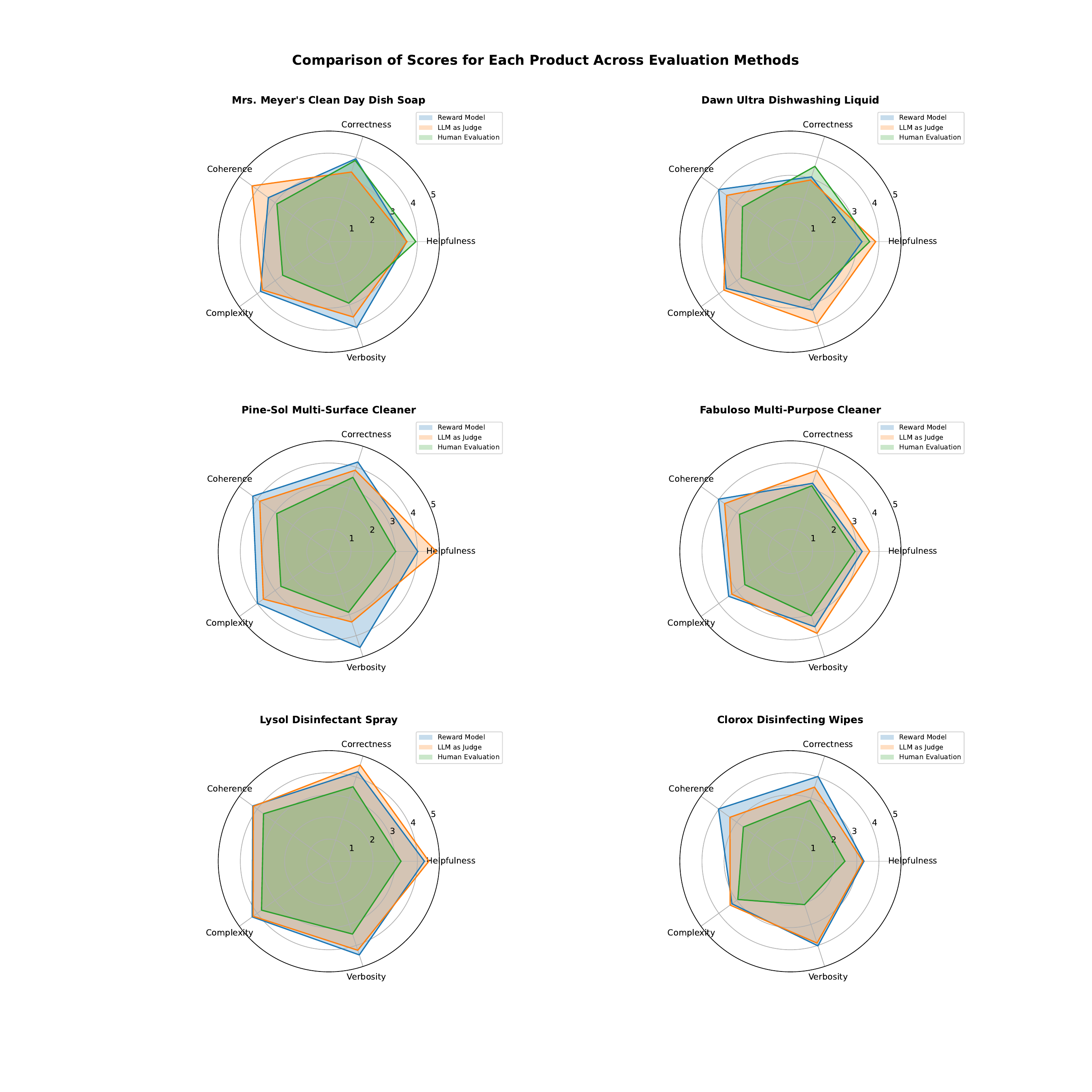}
    \vspace{-25mm}
    \caption{Comparison of AI-generated advertisements for cleaning and detergent products, including Mrs. Meyer’s Clean Day Dish Soap, Dawn Ultra Dishwashing Liquid (Dishwashing Liquids), Pine-Sol Multi-Surface Cleaner, Fabuloso Multi-Purpose Cleaner (Multi-Purpose Cleaners), Lysol Disinfectant Spray, and Clorox Disinfecting Wipes (Disinfectants). The advertisements are evaluated across multiple dimensions—Helpfulness, Correctness, Coherence, Complexity, and Verbosity—using Reward Model Scoring, LLM-as-Judge Evaluation, and Human Evaluation. The figure highlights variations in scores across products and evaluation methods, providing insights into the effectiveness of AI-generated advertisements tailored to different consumer preferences and product categories.}
    \label{fig:product_comparison_scores2}
\end{figure*}

\newpage
\clearpage

Example of a Personalized Advertising Campaign for a Simulated Consumer Persona: Table~\ref{tab:persona_example} introduces the simulated persona ``Alice", while Tables~\ref{tab:hyper_personalized_example}, \ref{tab:Clickability_Score_example}, and \ref{tab:Social_media_posts_example} illustrate her hyper-personalized advertisement, ad clickability analysis, and AI-generated social media posts, respectively. As shown in Table~\ref{tab:persona_example}, the simulated persona “Alice” represents a middle-income consumer with value-driven financial behavior. To effectively engage Alice and similar consumer segments, the Personalized Market-Aware Targeted Advertisement Generation (PAG) system leverages AI-driven hyper-personalization to optimize advertisements based on her demographic profile, interests, and spending habits. Table~\ref{tab:hyper_personalized_example} illustrates an example of a hyper-personalized advertisement generated for Alice. The ad aligns with her interests in fashion and travel while emphasizing sustainability and affordability. It dynamically adjusts messaging, visuals, and call-to-action strategies to enhance engagement, reflecting a personalized marketing approach tailored to her lifestyle preferences. To assess the effectiveness of such targeted advertisements, Table~\ref{tab:Clickability_Score_example} presents an analysis of ad clickability, evaluating strengths such as direct personalization, emotional appeal, and visually engaging content. The results indicate a high engagement score, with key areas for improvement identified to further optimize consumer response, such as stronger fashion-centric messaging and promotional incentives. Beyond static advertisements, Table~\ref{tab:Social_media_posts_example} highlights AI-generated, platform-specific social media posts designed for Twitter, Instagram, Facebook, and email marketing. These posts incorporate tailored messaging, audience-specific engagement strategies, and platform-adapted content to ensure a seamless, cross-channel advertising experience. By integrating hyper-personalization with audience insights, the system enhances brand engagement and maximizes advertising effectiveness across diverse digital ecosystems. Together, these components demonstrate the power of AI-driven, data-informed advertising strategies in creating highly relevant, impactful marketing campaigns. By modeling consumer personas like Alice and continuously optimizing advertisements through engagement metrics, businesses can enhance consumer interactions, improve conversion rates, and drive targeted outreach at scale.

\begin{table}[htbp]
    \centering
    \renewcommand{\arraystretch}{1.2} 
    \begin{tabular}{ll}
        \toprule
        \textbf{Attribute} & \textbf{Value} \\
        \midrule
        Name & Alice \\
        Age & 30 \\
        Gender & Female \\
        Interests & Fashion, Travel \\
        Socioeconomic Class & Middle-income \\
        Spending Power & Moderate \\
        Financial Behavior & Value-driven \\
        Language & English \\
        Cultural Context & Western Europe \\
        Emotional State & Excited \\
        Goal & Explore \\
        \bottomrule
    \end{tabular}
    \caption{Example of a simulated consumer persona, ``Alice”, used in the Simulated Humanistic Colony of Agents within the Personalized Market-Aware Targeted Advertisement Generation (PAG) system. This persona represents a 30-year-old female from Western Europe with moderate spending power and a value-driven financial approach. By modeling attributes such as interests, emotional state, and cultural context, AI-driven systems optimize hyper-personalized advertisements tailored to distinct consumer segments.}
    \label{tab:persona_example}
\end{table}

\begin{table}[htbp]
\centering
\begin{tabular}{>{\raggedright\arraybackslash}p{0.2\textwidth} >{\raggedright\arraybackslash}p{0.75\textwidth}}
\toprule
\rowcolor{headercolor}\multicolumn{2}{l}{\textcolor{white}{\textbf{Personalized Advertisement}}} \\
\midrule
\rowcolor{lightblue}
\textbf{Ad Title} & \textbf{Experience the Magic of Family Moments with Surf Excel} \\
\midrule
\rowcolor{visualcolor}
\textbf{Visual Description} & A vibrant, colorful image of a family enjoying a sunny picnic in a beautiful park. The family members are dressed in trendy outfits, and there are playful stains on their clothes visible as they laugh and share joyful moments together. In the background, a stylish travel bag is placed nearby, hinting at their adventurous spirit. \\
\midrule
\rowcolor{lightblue}
\textbf{Headline} & ``Adventure Awaits! Keep Your Family's Style Spotless with Surf Excel'' \\
\midrule
\rowcolor{lightblue}
\textbf{Body Text}  & 
Hey Alice! 

Life's too short to worry about stains when you're making unforgettable memories with your loved ones! At Surf Excel, we believe in celebrating life and embracing every moment, whether you're traveling to breathtaking destinations or enjoying a lovely day outdoors.

\textbf{What Makes Surf Excel Perfect for You:}
\begin{itemize}
    \item \textbf{Eco-Friendly Cleaning} : Our products are crafted with care for both your clothes and the planet. Enjoy stain-free adventures while doing your part for the environment! 
    \item \textbf{Stylish Solutions} : With our sleek packaging and color-friendly formulas, keeping your favorite outfits looking fresh and fashionable has never been easier.
    \item \textbf{Real Family Stories} : Join families who trust Surf Excel to tackle stains from every adventure. Check out testimonials that highlight how our detergent has become a travel companion for lasting memories!
\end{itemize}

\textbf{Cost-Effective Performance} : Get the quality and reliability you deserve without breaking the bank. Invest in products that enrich your lifestyle and keep your fashion game strong! \\
\midrule
\rowcolor{lightblue}
\textbf{Call to Action} & ``Shop Now and Get Ready for Your Next Adventure!'' \\
\midrule
\rowcolor{lightblue}
\textbf{Footer} & Include social media icons and a hashtag: \#SurfExcelMoments | Visit Us: www.surfexcel.com \\

\midrule
\rowcolor{visualcolor}
\textbf{Visual Note} & Include earthy tones and trendy design elements to appeal visually while integrating eco-conscious graphics, like soft leaves or water droplets, to emphasize sustainable practices. \\
\midrule
\rowcolor{lightblue}
\textbf{Closing Note} & This ad resonates with your excitement for exploring life's moments, aligns with your interests in fashion and travel, and emphasizes eco-friendly values, making it not just an advertisement but an invitation to a lifestyle you love! \\
\bottomrule
\end{tabular}
\caption{Example of a hyper-personalized advertisement generated by the Personalized Market-Aware Targeted Advertisement Generation (PAG) system. This advertisement is tailored for ``Alice," a simulated persona with interests in fashion and travel, moderate spending power, and an eco-conscious mindset. It incorporates dynamic content, including personalized messaging, tailored visuals, and platform-specific adaptations, to enhance engagement and drive conversions.}
 \label{tab:hyper_personalized_example}
\end{table}

\begin{table}[htbp]
\centering
\begin{tabular}{>{\raggedright\arraybackslash}p{0.3\textwidth} >{\raggedright\arraybackslash}p{0.65\textwidth}}
\toprule
\rowcolor{headercolor}\multicolumn{2}{l}{\textcolor{white}{\textbf{Ad Clickability Analysis}}} \\
\midrule
\rowcolor{lightblue}
\textbf{Overall Rating} & \textbf{85\%} Clickability Score \\
\midrule
\multicolumn{2}{l}{\textbf{Strengths}} \\
\rowcolor{strengthcolor}
1. \textbf{Personalization} & The ad directly addresses ``Alice", which creates a personal connection and makes it feel tailored to the recipient. This can significantly enhance engagement. \\
\rowcolor{strengthcolor}
2. \textbf{Visual Appeal} & The description of the visual elements, such as a vibrant family picnic and trendy outfits, aligns well with Alice's interests in fashion and travel. A colorful and joyful image can attract attention and elicit positive emotions, which may lead to increased click-through rates. \\
\rowcolor{strengthcolor}
3. \textbf{Clear Value Proposition} & The ad effectively communicates several benefits of using Surf Excel, including eco-friendliness, stylish packaging, and testimonials from real families. This not only highlights the product's features but also addresses a potential concern for consumers—keeping clothes clean while having fun. \\
\rowcolor{strengthcolor}
4. \textbf{Emotional Appeal} & The ad taps into the excitement of making memories and adventures, which resonates with Alice's emotional state of being excited. It invites her to embrace life and its messes, framing stains as a natural part of family fun. \\
\rowcolor{strengthcolor}
5. \textbf{Strong Call to Action} & The call to action (``Shop Now and Get Ready for Your Next Adventure!") is inviting and motivating. It encourages immediate action, which can lead to higher click rates. \\
\midrule
\multicolumn{2}{l}{\textbf{Areas for Improvement}} \\
\rowcolor{improvementcolor}
1. \textbf{Focus on Fashion} & While fashion is mentioned, more emphasis on how Surf Excel preserves trendy clothing could enhance resonance with fashion-conscious consumers. Specific examples of fashionable brands or styles that work well with Surf Excel might strengthen the appeal further. \\
\rowcolor{improvementcolor}
2. \textbf{Additional Incentives} & Introducing a limited-time offer, discount, or free sample could improve clickability. While the ad mentions cost-effectiveness, a specific incentive could create urgency and better encourage clicks. \\
\rowcolor{improvementcolor}
3. \textbf{Visual Enhancements} & The visual note provides good suggestions, but incorporating a more specific visual example, such as showing the product being used in the context of a fashionable outing or adventure, might enhance interest. \\
\midrule
\rowcolor{lightblue}
\textbf{Conclusion} & Overall, the ad is well-crafted and effectively targets Alice's interests and emotional state, resulting in a high clickability score. However, slight adjustments could make it even more compelling. \\
\bottomrule
\end{tabular}
\caption{Clickability analysis for a personalized advertisement targeting the simulated persona ``Alice." The table highlights key strengths, including personalization, visual appeal, and emotional engagement, contributing to an 85\% clickability score. Areas for improvement, such as a stronger emphasis on fashion-related messaging and additional incentives, are identified to further optimize performance.}
\label{tab:Clickability_Score_example}
\end{table}

\begin{table}[htbp]
\centering
\begin{tabular}{>{\raggedright\arraybackslash}p{0.15\textwidth} >{\raggedright\arraybackslash}p{0.8\textwidth}}
\toprule
\rowcolor{headercolor}\multicolumn{2}{l}{\textcolor{white}{\textbf{\ Social Media Posts}}} \\
\midrule

\rowcolor{twitterblue!15}
\textbf{\ Twitter} & 
"Life's too short to worry about stains while making memories! 
Embrace every adventure with your loved ones and let Surf Excel keep your family stylish \& eco-friendly. Ready for a stain-free journey? 
\textbf{\#SurfExcelMoments}" \\
\midrule

\rowcolor{instapink!15}
\textbf{\ Instagram} & 
\textbf{Family Adventures Await!} 

Dive into a world of fun and unforgettable moments with your loved ones! At Surf Excel, we celebrate life's joys while ensuring your family's favorite outfits stay spotless and stylish. Our eco-friendly cleaning solutions mean you can adventure freely—no worrying about stains!

Picture this: a sunny picnic in the park, laughter echoing around, and beautiful outfits shining bright, even with a few playful stains! 

Join the Surf Excel family and share in the real stories of families who trust us to keep their clothes fresh. Your journey to stain-free memories starts now! 

\textbf{Shop Now for Your Next Adventure!}

\textbf{\#SurfExcelMoments \#EcoFriendly \#FamilyFun \#StainFreeMemories} \\
\midrule

\rowcolor{facebookblue!15}
\textbf{Facebook} & 
\textbf{Hey friends!} 

Are you ready to create beautiful memories with your family without worrying about those pesky stains? At Surf Excel, we believe every adventure deserves to be celebrated! Whether you're stylishly picnicking in the park or exploring breathtaking destinations, we've got your back!

\textbf{Why Choose Surf Excel?}
\begin{itemize}
\item  \hspace{1mm} Eco-Friendly Cleaning
\item  \hspace{1mm} Stylish Solutions
\item  \hspace{1mm} Cost-Effective Performance
\end{itemize}

Join thousands of families who trust Surf Excel for their stain-fighting needs. Check out our powerful products, designed to keep your clothes looking fresh and vibrant through all of life's adventures!

\textbf{Shop Now and Get Ready for Your Next Adventure!}

\textbf{\#SurfExcelMoments \#Family \#Adventure \#EcoFriendly} \\
\midrule

\rowcolor{emailgray!15}
\textbf{Email} & 
\textbf{Subject:} Embrace Every Adventure with Surf Excel! 

Dear [Recipient's Name],

Life's most precious moments are often filled with laughter, joy, and the occasional mess! At Surf Excel, we understand the importance of creating unforgettable memories with your loved ones without the worry of stains ruining your favorite outfits.

\textbf{Why Surf Excel is Right for You:}
\begin{itemize}
\item \textbf{Eco-Friendly Cleaning}: Make a positive impact on the planet while keeping your clothes looking pristine.
\item \textbf{Stylish Solutions}: Our product packaging is not only sleek but ensures your clothes remain fresh and fashionable.
\item \textbf{Real Family Stories}: Join the community of families who rely on Surf Excel to conquer stains from their most cherished adventures.
\end{itemize}

And let's not forget our cost-effective performance that doesn't compromise on quality.

Ready to keep your family stylish and those memories stain-free? 

\textbf{Shop now and embrace every adventure with confidence and style!} [Link to shop] \\
\bottomrule
\end{tabular}
\caption{Example of AI-generated social media posts optimized for Twitter, Instagram, Facebook, and email marketing. These posts leverage personalized messaging, audience-specific engagement strategies, and platform-tailored content to maximize ad effectiveness. By aligning with the persona's interests and emotional state, the social media campaign enhances engagement, click-through rates, and brand perception.}
\label{tab:Social_media_posts_example}
\end{table}

\newpage
\clearpage

\appendix
\section{Appendix}

\subsection{Ad Copy Optimization} 
Ad Copy Optimization is a critical component of modern marketing strategies, designed to maximize the effectiveness of advertising campaigns by refining and testing multiple ad variations. This process builds upon the Competitive Hyper-Personalized Advertisement System (CHPAS), which generates persona-specific, competitive advertisements for products within the same category from different manufacturers. Ad Copy Optimization enhances these ads by systematically evaluating their effectiveness across diverse consumer personas, ensuring they resonate with target customers while maintaining competitive differentiation. This optimization process is essential for improving key performance metrics such as click-through rates (CTR), which measure the percentage of users who click on an ad after viewing it, and conversions, which track the number of users who take a desired action, such as making a purchase or signing up for a service. By addressing diverse consumer preferences, platform-specific engagement requirements, and rapidly evolving market dynamics, Ad Copy Optimization ensures that marketing messages maximize engagement and drive meaningful customer action. To ground our experimental evaluation, we employ a diverse colony of Simulated Humanistic Agentic Personas (SHAP), which assess CHPAS-generated ads. These personas represent distinct consumer archetypes with well-defined demographic profiles, behavioral patterns, and decision-making frameworks, including: (a) Analytical Evaluator, which focuses on technical accuracy, factual consistency, and logical appeal; (b) Emotional Resonance Assessor, which measures emotional impact, brand connection, and relatability; (c) Cultural Context Validator, which ensures cultural relevance, sensitivity, and appropriateness for diverse audiences; and (d) Consumer Preference Analyzer, which evaluates alignment with consumer preferences and behavioral patterns.  The experimental framework systematically evaluates CHPAS-generated ad variations by aligning them with the unique preferences of these personas and the unique selling points (USPs) of the products. Audiences are segmented for randomized testing across platforms, ensuring a comprehensive evaluation of ad effectiveness through multi-dimensional analysis. Each agentic persona rates advertisements across multiple dimensions using a standardized evaluation framework, including: (a) Messaging Clarity (0–10); (b) Emotional Resonance (0–10) (c) Technical Accuracy (0–10); (d) Cultural Fit (0–10). To enhance the robustness of the evaluation process, inter-persona validation is employed, where multiple agents cross-validate each other's assessments using sophisticated validation protocols. This ensures reliability and minimizes bias in the results while maintaining statistical significance. Persona ratings are then analyzed with statistical rigor to identify the most effective ads, ensuring that the findings are both significant and actionable for real-world implementation. The integration of these persona-based evaluations provides a deep understanding of ad effectiveness, enabling businesses to iteratively refine advertisements and improve engagement metrics through data-driven optimization. By leveraging these multi-dimensional insights, chemical product manufacturers can reduce guesswork, adapt to evolving market trends, and allocate resources more effectively to maximize return on investment (ROI) and market penetration. This approach underscores the critical role of data-driven, persona-validated advertising in modern marketing strategies and competitive differentiation. In summary, Ad Copy Optimization serves as the refinement layer for CHPAS-generated advertisements, enhancing their effectiveness by incorporating simulated humanistic feedback, standardized evaluation metrics, and rigorous statistical analysis. This iterative approach ensures that marketing efforts remain competitive, personalized, and aligned with the dynamic needs of diverse consumer segments, ultimately driving higher engagement and improved conversion rates across multiple market segments.

\subsection{Open-Domain Question Answering (ODQA)}
This section presents an advanced Optimized Retrieval-Augmented Generation (RAG) system \cite{lewis2020retrieval, gupta2024comprehensive, gao2023retrieval, han2025graphrag}, designed to process and utilize multimodal agglomerated knowledge from the Multimodal Agentic Advertisement Market Survey (MAAMS) system for question-answering (QA) tasks. The knowledge synthesized by MAAMS comprises market intelligence from diverse modalities, including text, images, video, financial data, and market data, which is consolidated into a document-specific text output format for each product. While the system primarily processes textual knowledge extracted from documents, it enables semantic search and contextually accurate response generation for QA tasks. A QA system built on this agglomerated knowledge base acts as a crucial interface between customers and technical product information, allowing users to receive precise and relevant answers without navigating lengthy documentation. The system excels in four key areas:  (a) Providing direct access to product information; (b) Offering comparative product insights; (c) Translating technical specifications into user-friendly language; (d) Delivering specific usage guidance. By extracting and presenting relevant information in an accessible format, the QA system enhances the customer experience, making technical product information more actionable and user-friendly. The system begins by processing documents using an automated text extraction method, ensuring the structured retrieval of textual content and metadata. The extracted text is segmented into manageable, semantically coherent chunks with overlapping regions to maintain contextual continuity. To improve context awareness, each chunk is summarized for relevance in relation to other chunks within the same document, a process known as Contextual Relevance Summarization (CRS). These augmented chunks prioritize semantically aligned content, enhance long-context handling, and strengthen chain-of-thought reasoning. This refined retrieval in RAG pipelines minimizes noise, improves precision, and enhances multi-hop retrieval and source attribution. While summarization increases computational costs, the benefits of improved retrieval accuracy and reasoning outweigh the trade-offs.  Each augmented chunk is converted into a numerical vector representation using OpenAI's text-embedding-3-small model and indexed in a vector store for efficient cosine similarity-based retrieval. User queries are embedded into the same vector space and matched against stored embeddings to retrieve the most contextually relevant chunks. Responses are synthesized using OpenAI’s GPT-4o-mini model, integrating the top-K retrieved chunks to enhance factual accuracy, contextual relevance, and coherence through iterative refinement. To further improve response quality, the system incorporates an iterative self-reflection mechanism, where LLM-as-Judge (using OpenAI’s GPT-4o) critiques generated responses for potential weaknesses, such as factual inaccuracies, missing details, or lack of clarity. Based on these critiques, responses are iteratively revised until they meet either a predefined number of iterations or a quality threshold, ensuring optimal output. To evaluate performance, reference responses are generated using GPT-4o, serving as a benchmark for assessing alignment and correctness. A comprehensive evaluation framework measures retrieval and response quality using Precision, Recall, Mean Reciprocal Rank (MRR), Mean Average Precision (MAP), BLEU, METEOR, ROUGE, and cosine similarity. Faithfulness is assessed by comparing the semantic similarity between generated responses and retrieved chunks, ensuring alignment with source content, while relevance is evaluated by comparing the embeddings of generated responses with original query embeddings. Additionally, a state-of-the-art reward model (NVIDIA Nemotron-4-340b-reward) evaluates and scores the final responses, optimizing for user satisfaction and practical utility. The reward model assesses responses based on five key metrics: (a) Helpfulness (effectiveness in addressing the question); (b) Correctness (accuracy and inclusion of pertinent facts); (c) Coherence (clarity and logical flow); (d) Complexity (intellectual depth and domain expertise); (e) Verbosity (appropriate level of detail relative to the question). By combining retrieval-augmented generation, self-refinement, and multimodal processing, this AI-powered knowledge engine serves as a strategic decision-support system, enabling companies to extract, synthesize, and optimize market insights. It automates competitive analysis, facilitates data-driven decision-making, and retrieves multimodal knowledge for benchmarking products, refining branding, generating targeted ad content, and ensuring regulatory compliance.  Marketing executives, product managers, brand analysts, and compliance teams leverage the system as an enterprise chatbot, API, semantic search engine, and advertising optimization platform to enhance knowledge retrieval, consumer engagement, and strategic effectiveness. For evaluation, a benchmark QA dataset of 2,500 QA pairs is created using GPT-4o, refined through human review to ensure accuracy and relevance, serving as a benchmark for the Optimized RAG system.  In summary, this AI-powered engine serves as a strategic decision-support tool, automating competitive analysis, improving branding, and enabling data-driven marketing, making it a scalable and intelligent solution for enterprise use in marketing, compliance, and consumer engagement.

\begin{figure*}[htbp]
    \centering
    \includegraphics[
        width=0.725\textwidth,
        trim=0mm 8mm 0mm 15mm, 
        clip
    ]{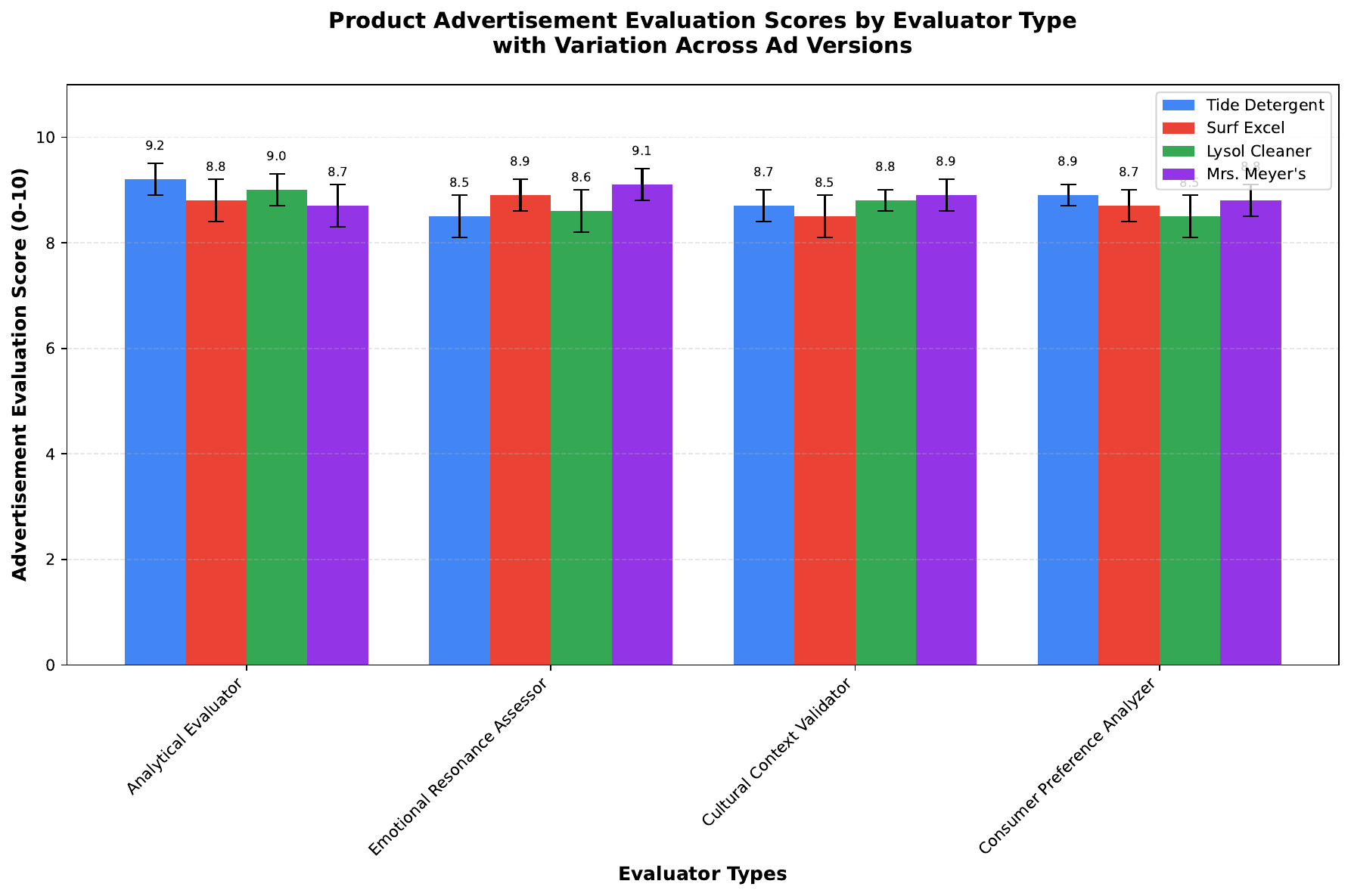}
    \vspace{-4mm}
    \caption{A comparative analysis of product advertisements across different evaluator types, showing mean evaluation scores and variations across multiple ad versions. Error bars represent standard deviations in scores across different advertisement variations for each product-evaluator combination. The plot illustrates how each evaluator persona (Analytical Evaluator, Emotional Resonance Assessor, Cultural Context Validator, and Consumer Preference Analyzer) assesses different versions of advertisements for each product, offering insights into both the average effectiveness and consistency of advertising approaches.}
    \label{fig:product_comparison_scores2}
    \vspace{-5mm}
\end{figure*}

\begin{figure*}[htbp] 
    \centering 
    \includegraphics[
        width=1.02\textwidth, 
        height=1.5\textheight,  
        keepaspectratio, 
        trim=0mm 0mm 0mm 0mm, 
        clip  
    ]{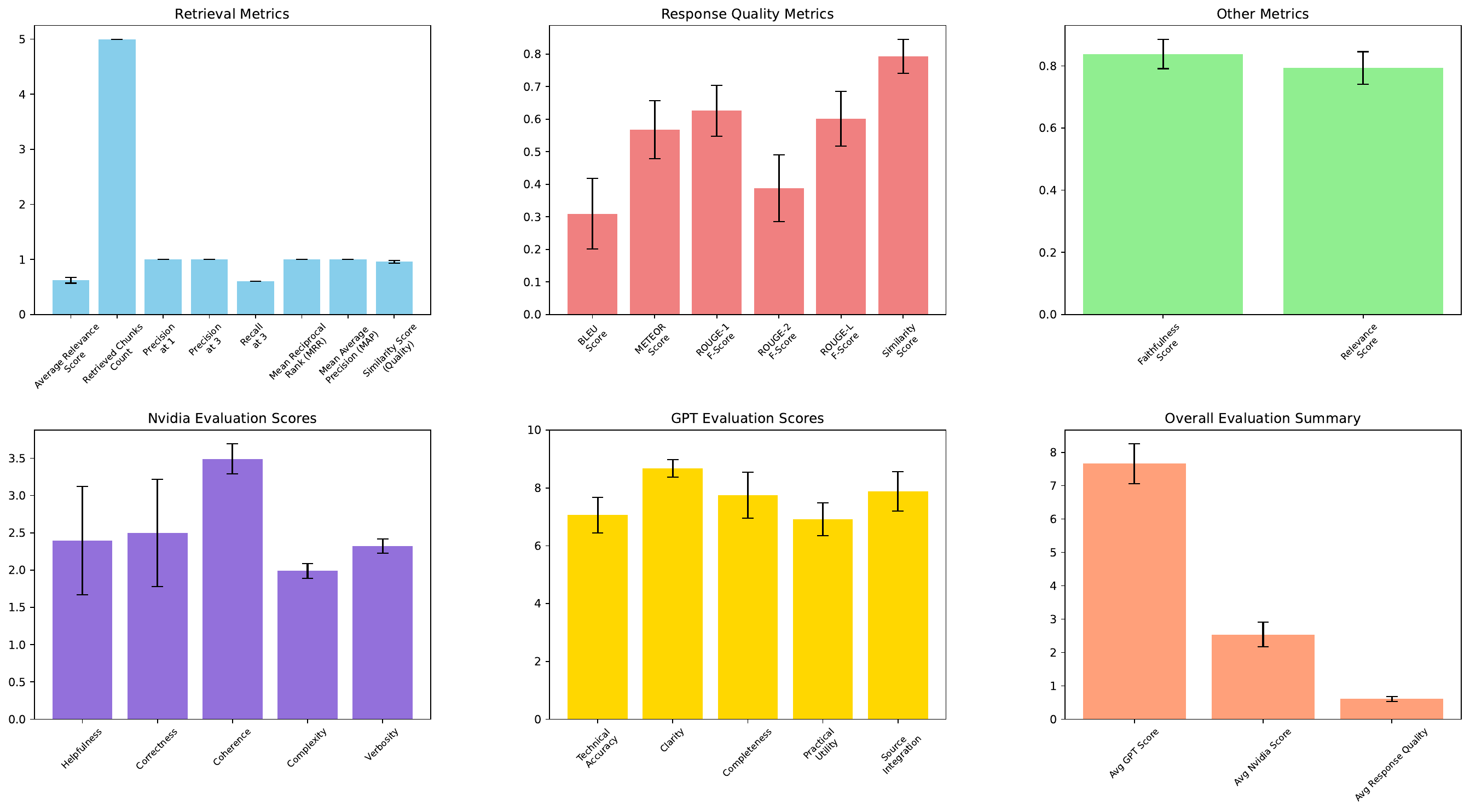} 
    \vspace{-7mm}
    \caption{A comprehensive evaluation of the Optimized RAG System. Subfigures 1–3 analyze retrieval performance through precision, recall, and relevance, ensuring reliable and contextually relevant responses. Subfigures 4–5 present AI-driven evaluations using NVIDIA's Nemotron and GPT-based scoring, assessing correctness, completeness, and practical utility. Subfigure 6 summarizes overall performance, highlighting the balance between accurate retrieval and high-quality response generation.} 
    \label{fig:RAGREsults} 
\end{figure*}

\subsection{Results}
Figure \ref{fig:RAGREsults} presents the retrieval metrics used to evaluate the performance of the Optimized RAG system across six key indicators. The Average Relevance Score (scale: 0-1) quantifies the alignment of retrieved chunks with the query, where higher values indicate better relevance. The Retrieved Chunk Count measures the number of chunks retrieved per query, which varies based on retrieval settings. Precision@1 and Precision@3 (scale: 0-1) assess the proportion of relevant chunks within the top-1 and top-3 retrieved results, respectively, with Precision@1 focusing on the relevance of the first result and Precision@3 evaluating the top three results. Recall@3 (scale: 0-1) measures the proportion of all relevant chunks retrieved within the top-3 results compared to the total relevant chunks in the dataset, emphasizing retrieval coverage. Mean Reciprocal Rank (MRR) (scale: 0-1) evaluates the rank position of the first relevant chunk, with higher values indicating faster retrieval of relevant information. Mean Average Precision (MAP) (scale: 0-1) aggregates precision scores across multiple recall levels, providing a comprehensive measure of retrieval performance. High values in the Average Relevance Score, Precision@k, MRR, and MAP indicate effective retrieval, ensuring high-quality context for response generation. While a high Retrieved Chunk Count improves recall, it may introduce noise, whereas a low count risks limiting information coverage. This figure demonstrates the system’s ability to optimize retrieval accuracy and relevance, which is crucial for improving response synthesis.  Response quality metrics assess the coherence, accuracy, and informativeness of generated responses by comparing them to benchmark responses from GPT-4o. The BLEU Score (scale: 0-1) measures n-gram overlap with reference answers, where higher values indicate better alignment. The METEOR Score (scale: 0-1) accounts for synonym matching, fluency, and recall, improving upon BLEU by considering linguistic variations. ROUGE-1, ROUGE-2, and ROUGE-L F-Scores (scale: 0-1) assess token overlap, capturing unigram and bigram precision-recall relationships. The Similarity Score (scale: 0-1) quantifies the semantic alignment between generated and reference responses, ensuring contextual consistency. High values across these metrics indicate superior response generation with enhanced readability and factual alignment. Faithfulness and Relevance Scores measure response reliability and query alignment. The Faithfulness Score (scale: 0-1) assesses how well the generated response adheres to retrieved source content, ensuring factual integrity and reducing hallucination risks. The Relevance Score (scale: 0-1) evaluates the semantic similarity between the generated response and the original query, determining how well the system addresses user intent. Higher scores in both metrics indicate that the RAG system remains grounded in retrieved evidence, improving trustworthiness and response credibility. The NVIDIA evaluation scores present response assessments from NVIDIA’s Nemotron-4-340b-reward model, which scores responses on a scale of 0-4, where higher values indicate superior quality. The key evaluation metrics include Helpfulness, Correctness, Coherence, Complexity, and Verboseness. Helpfulness measures how well responses address queries, while Correctness evaluates factual accuracy and completeness. Coherence assesses logical structuring and readability, while Complexity considers the depth of reasoning and domain knowledge integration. Verboseness ensures a balance between conciseness and necessary detail. Observed scores for individual responses range from approximately 0.5 to 3.5 across these dimensions. Higher values indicate well-structured, informative, and contextually accurate responses, while lower values suggest deficiencies in clarity, completeness, or factual grounding.  The GPT evaluation scores assess response quality using GPT-4-based metrics, covering Technical Accuracy, Clarity, Completeness, Practical Utility, and Source Integration. Technical Accuracy ensures domain-specific precision, Clarity measures readability and logical flow, and Completeness evaluates the coverage of key aspects. Practical Utility assesses real-world applicability, while Source Integration verifies grounding in retrieved evidence. Scores range from 0 (poor) to approximately 10 (excellent), with higher values indicating more coherent, informative, and factually robust responses. Lower scores suggest weaker grounding, incomplete information, \subsection{Open-Domain Question Answering (ODQA)}
This section presents an advanced Optimized Retrieval-Augmented Generation (RAG) system \cite{lewis2020retrieval, gupta2024comprehensive, gao2023retrieval, han2025graphrag}, designed to process and utilize multimodal agglomerated knowledge from the Multimodal Agentic Advertisement Market Survey (MAAMS) system for question-answering (QA) tasks. The knowledge synthesized by MAAMS comprises market intelligence from diverse modalities, including text, images, video, financial data, and market data, which is consolidated into a document-specific text output format for each product. While the system primarily processes textual knowledge extracted from documents, it enables semantic search and contextually accurate response generation for QA tasks. A QA system built on this agglomerated knowledge base acts as a crucial interface between customers and technical product information, allowing users to receive precise and relevant answers without navigating lengthy documentation. The system excels in four key areas:  (a) Providing direct access to product information; (b) Offering comparative product insights; (c) Translating technical specifications into user-friendly language; (d) Delivering specific usage guidance. By extracting and presenting relevant information in an accessible format, the QA system enhances the customer experience, making technical product information more actionable and user-friendly. The system begins by processing documents using an automated text extraction method, ensuring the structured retrieval of textual content and metadata. The extracted text is segmented into manageable, semantically coherent chunks with overlapping regions to maintain contextual continuity. To improve context awarenesor reduced practical relevance.  The overall evaluation summary consolidates the performance of the Optimized RAG system, displaying key evaluation metrics. The Avg GPT Score (scale: 0-10) reflects language model-based evaluation, ensuring linguistic and contextual quality. The Avg NVIDIA Score (scale: 0-4) aggregates reward model evaluations, capturing user-centric response effectiveness. The Avg Response Quality (scale: 0-1) combines retrieval precision and response coherence, offering a comprehensive measure of system performance. Higher averages across these metrics indicate a balanced trade-off between accurate retrieval and high-quality response generation. In summary, this set of figures presents a comprehensive evaluation of the Optimized RAG system, covering retrieval accuracy, response quality, faithfulness, relevance, and benchark model-based assessments. In Figure \ref{fig:RAGREsults}, Subfigures 1–3 analyze retrieval precision, response coherence, and factual grounding, ensuring the system’s reliability and contextual relevance. Subfigures 4–5 showcase AI-driven evaluations from NVIDIA’s Nemotron and GPT-based scoring, emphasizing correctness, completeness, and utility. Subfigure 6 integrates these insights into an overall performance summary, highlighting the balance between accurate retrieval and high-quality response generation.

\subsection{Mathematical Modeling}
The proposed framework for multimodal analysis and personalized advertisement generation in the chemical industry is formalized through a structured mathematical representation, capturing the key components of the Multimodal Agentic Advertisement Market Survey (MAAMS), Personalized Market-Aware Targeted Advertisement Generation (PAG), and Competitive Hyper-Personalized Advertisement System (CHPAS). The MAAMS system integrates insights from multiple specialized agents—Text, Image, Video, Finance, and Market—represented as \( \mathcal{M} = \{ \text{Text}, \text{Image}, \text{Video}, \text{Finance}, \text{Market} \} \). Each agent \( m \in \mathcal{M} \) processes domain-specific raw data \( \mathcal{D}_m = \{ d_1^m, d_2^m, \dots, d_{N_m}^m \} \), such as search engine text, social media images, and financial reports, into structured insights via a transformation function \( f_m: \mathcal{D}_m \rightarrow \mathcal{I}_m \), where \( \mathcal{I}_m = f_m(\mathcal{D}_m) \). The aggregated insights form a multimodal knowledge base \( \mathcal{K} = \bigcup_{m \in \mathcal{M}} \mathcal{I}_m \), which is further synthesized by a Meta-Agent \( \mathcal{A}_{\text{meta}} \), leveraging large language models (LLMs), to produce a market intelligence report \( \mathcal{R} = \mathcal{A}_{\text{meta}}(\mathcal{K}) \). This structured synthesis enables efficient data collection and decision-oriented insights, driving personalized advertisement generation and competitive ad differentiation in the PAG and CHPAS systems. The PAG system utilizes the market intelligence report \( \mathcal{R} \), synthesized from the multimodal knowledge base \( \mathcal{K} \), to generate personalized advertisements for \( N \) consumer personas, denoted as \( \mathcal{P} = \{ p_1, p_2, \dots, p_N \} \). Each persona \( p \in \mathcal{P} \) is associated with a preference set \( \mathcal{C}_p \), capturing its traits, interests, and goals (e.g., Logical Strategist, Visionary Trailblazer). The Advertisement Curator Agent \( \mathcal{A}_{\text{curator}} \) generates persona-specific advertisements \( \mathcal{AD}_p \) using structured insights from \( \mathcal{R} \), formulated as \( \mathcal{A}_{\text{curator}}(\mathcal{R}, \mathcal{C}_p) = \mathcal{AD}_p \). These advertisements are further optimized for specific platforms \( \mathcal{S} = \{ \text{Twitter}, \text{Instagram}, \text{Facebook} \} \) by the Social Media Agent \( \mathcal{A}_{\text{social}} \), producing platform-adapted advertisements \( \mathcal{AD}_{p, \mathcal{S}} \), represented as \( \mathcal{A}_{\text{social}}(\mathcal{AD}_p, \mathcal{S}) = \mathcal{AD}_{p, \mathcal{S}} \). This ensures that advertisements are persona-specific, leveraging refined insights from \( \mathcal{R} \), and optimized for platform constraints, cultural relevance, and linguistic appropriateness. The CHPAS system extends personalization to competing products, denoted as \( \mathcal{Q} = \{ q_1, q_2, \dots, q_M \} \), where each \( q_i \) represents a specific product (e.g., laundry detergents from different manufacturers). For each persona \( p \), the system generates hyper-personalized advertisements \( \mathcal{AD}_{p,q} \) based on affinity scores \( \alpha_{p,q} \) and competitive strengths \( \beta_q \). The affinity score \( \alpha_{p,q} \) quantifies the alignment between persona preferences \( \mathcal{C}_p \) and product features, computed as \( \alpha_{p,q} = f_{\text{affinity}}(\mathcal{C}_p, \mathcal{R}, q) \), where \( f_{\text{affinity}} \) evaluates the relevance of product \( q \) to persona \( p \). The competitive strength \( \beta_q \) captures the product’s unique selling points, computed as \( \beta_q = f_{\text{strength}}(\mathcal{R}, q) \), where \( f_{\text{strength}} \) assesses the competitiv\subsection{Open-Domain Question Answering (ODQA)}
This section presents an advanced Optimized Retrieval-Augmented Generation (RAG) system \cite{lewis2020retrieval, gupta2024comprehensive, gao2023retrieval, han2025graphrag}, designed to process and utilize multimodal agglomerated knowledge from the Multimodal Agentic Advertisement Market Survey (MAAMS) system for question-answering (QA) tasks. The knowledge synthesized by MAAMS comprises market intelligence from diverse modalities, including text, images, video, financial data, and market data, which is consolidated into a document-specific text output format for each product. While the system primarily processes textual knowledge extracted from documents, it enables semantic search and contextually accurate response generation for QA tasks. A QA system built on this agglomerated knowledge base acts as a crucial interface between customers and technical product information, allowing users to receive precise and relevant answers without navigating lengthy documentation. The system excels in four key areas:  (a) Providing direct access to product information; (b) Offering comparative product insights; (c) Translating technical specifications into user-friendly language; (d) Delivering specific usage guidance. By extracting and presenting relevant information in an accessible format, the QA system enhances the customer experience, making technical product information more actionable and user-friendly. The system begins by processing documents using an automated text extraction method, ensuring the structured retrieval of textual content and metadata. The extracted text is segmented into manageable, semantically coherent chunks with overlapping regions to maintain contextual continuity. To improve context awarenes\subsection{Open-Domain Question Answering (ODQA)}
This section presents an advanced Optimized Retrieval-Augmented Generation (RAG) system \cite{lewis2020retrieval, gupta2024comprehensive, gao2023retrieval, han2025graphrag}, designed to process and utilize multimodal agglomerated knowledge from the Multimodal Agentic Advertisement Market Survey (MAAMS) system for question-answering (QA) tasks. The knowledge synthesized by MAAMS comprises market intelligence from diverse modalities, including text, images, video, financial data, and market data, which is consolidated into a document-specific text output format for each product. While the system primarily processes textual knowledge extracted from documents, it enables semantic search and contextually accurate response generation for QA tasks. A QA system built on this agglomerated knowledge base acts as a crucial interface between customers and technical product information, allowing users to receive precise and relevant answers without navigating lengthy documentation. The system excels in four key areas:  (a) Providing direct access to product information; (b) Offering comparative product insights; (c) Translating technical specifications into user-friendly language; (d) Delivering specific usage guidance. By extracting and presenting relevant information in an accessible format, the QA system enhances the customer experience, making technical product information more actionable and user-friendly. The system begins by processing documents using an automated text extraction method, ensuring the structured retrieval of textual content and metadata. The extracted text is segmented into manageable, semantically coherent chunks with overlapping regions to maintain contextual continuity. To improve context awarenese advantages of product \( q \). The CHPAS system generates hyper-personalized advertisements as \( \mathcal{A}_{\text{comp}}(\mathcal{R}, \mathcal{C}_p, q) = \mathcal{AD}_{p,q} \), ensuring that advertisements are not only tailored to individual preferences but also strategically highlight each product's competitive advantages. The effectiveness of the generated advertisements is evaluated using AI-driven and human-aligned metrics to ensure they are engaging, relevant, and persuasive. Specifically, the evaluation incorporates helpfulness \( H(\mathcal{AD}_{p,q}) \), correctness \( C(\mathcal{AD}_{p,q}) \), and coherence \( R(\mathcal{AD}_{p,q}) \), while faithfulness \( F(\mathcal{AD}_{p,q}) \) measures alignment with retrieved market intelligence, and relevance \( V(\mathcal{AD}_{p,q}) \) quantifies how well the advertisement matches consumer expectations. These scores are derived from NVIDIA Nemotron-4-340b-reward and LLM-as-Judge, evaluation models. The optimization objective is to maximize the weighted sum of these metrics:  
\[
\max_{\mathcal{AD}_{p,q}} \left( w_1 H(\mathcal{AD}_{p,q}) + w_2 C(\mathcal{AD}_{p,q}) + w_3 R(\mathcal{AD}_{p,q}) + w_4 F(\mathcal{AD}_{p,q}) + w_5 V(\mathcal{AD}_{p,q}) \right),
\]
where \( w_1, w_2, w_3, w_4, w_5 \) are hyperparameters reflecting the relative importance of each factor. By integrating multimodal AI analysis, retrieval-augmented market insights, and persona-driven optimization, the framework ensures that advertisements are not only technically accurate and engaging but also strategically competitive and culturally relevant.

\vspace{-2mm}
\subsection{Experimental Setup} 
\vspace{-1mm}
The experimental validation integrates real-world environments to evaluate our three-system framework: (1) MAAMS (Multimodal Agentic Advertisement Market Survey) for market analysis, (2) PAG (Personalized Market-Aware Targeted Advertisement Generation) for personalized multilingual advertising, and (3) CHPAS (Competitive Hyper-Personalized Advertisement System) for competitive ad differentiation. The MAAMS system is orchestrated by a Meta-Agent powered by GPT-4o, responsible for cross-modal synthesis and market intelligence aggregation through specialized agents. The Text Agent (GPT-4o + SerpAPI) retrieves and analyzes consumer sentiment, brand messaging, and regulatory compliance insights from Google, Bing, and Yahoo. The Image Agent (GPT-4o + SerpAPI + Instagram and Pinterest APIs) conducts visual content analysis, fetching images from search queries to evaluate brand identity, packaging, and product perception. The Video Agent (GPT-4o + YouTube API) extracts keyframes and transcripts from YouTube videos to assess emotional appeal, storytelling effectiveness, and product highlights in advertisements. The Finance Agent (GPT-4o + Bloomberg and Yahoo Finance) processes financial performance indicators, revenue trends, and advertising expenditures using Bloomberg and Yahoo Finance data pipelines. The Market Agent (GPT-4o + Statista and Google Trends via Pytrends) analyzes consumer behavior trends, competitor benchmarking, and market demands using Statista and Google Trends APIs. The PAG system generates multilingual, persona-driven advertisements using a Simulated Humanistic Colony of Agents built on GPT-4o, simulating diverse consumer segments, including but not limited to: the Logical Strategist, who prioritizes scientific innovation; the Visionary Trailblazer, who focuses on sustainability and eco-conscious branding; the Harmonious Connector, who values emotional connections and community impact; the Resilient Optimist, who responds to confidence-building and problem-solving messaging; and the Organized Architect, who prefers efficiency and structured communication. The system incorporates an Advanced Curator Agent (GPT-4o) that creates multilingual advertisements, ensuring cultural and linguistic alignment, and a Social Media Agent (GPT-4o) that optimizes content for platform-specific engagement, refining advertisements for Twitter, Instagram, and Facebook. The CHPAS system incorporates product differentiation from various manufacturers, increases affinity scoring for persona-product matching, implements cannibalization prevention mechanisms, and optimizes unique selling proposition (USP) emphasis for competing products, such as tailoring detergent advertisements differently for specific user segments. Real-world testing encompasses data from over 50 fast-moving consumer goods (FMCG) companies across 50+ product categories, with 2,000+ products spanning household care, cosmetics, and pharmaceuticals. The system simulates diverse personas across occupational groups, emotional profiles, language preferences (English, Spanish, Asian, European, Middle Eastern), and socioeconomic classes. The evaluation framework employs comprehensive metrics, including clickability rate tracking across initial, personalized, and hyper-personalized advertisements. Quality assessment is conducted using the NVIDIA Nemotron-4-340b-reward model, LLM-as-Judge evaluation with GPT-4o, and human evaluator benchmarking, analyzing five key dimensions: helpfulness (actionable information), correctness (factual accuracy), coherence (logical flow), complexity (level of detail), and verbosity (conciseness). To enhance computational efficiency in AI-driven advertisement generation, we adopt a structured pipeline where MAAMS is executed first to generate and store market intelligence before subsequent PAG and CHPAS stages. This decoupled approach minimizes redundant processing, reduces latency, and optimizes resource utilization by avoiding repeated execution of MAAMS during each ad iteration. Instead, MAAMS outputs serve as a cached knowledge base, allowing PAG and CHPAS to efficiently generate persona-specific and competitively optimized advertisements. This separation ensures scalable, high-performance ad generation while maintaining adaptability to evolving market conditions.

\vspace{-2mm}
\subsection{Related Work}
\vspace{-1mm}
Advancements in artificial intelligence have significantly transformed personalized advertising, enabling the creation of highly targeted and engaging campaigns. However, advertising in specialized domains, such as chemical products, presents unique challenges that remain underexplored. This section reviews recent contributions to personalized advertising, multimodal data analysis, and AI-driven consumer persona modeling, highlighting their relevance to our work.

\vspace{-2mm}
\subsubsection{Personalized Advertising}
\vspace{-1mm}
Personalized advertising has evolved from traditional demographic-based targeting to AI-driven methods that enable real-time personalization. Recent work by \citet{adSformers} introduced the adSformers framework, which utilizes transformer-based architectures to dynamically model short-term user behaviors and optimize ad recommendations. Similarly, \citet{UniMP} proposed a unified multimodal personalization system that integrates text, image, and interaction data for generative recommendations. While these approaches excel in consumer goods markets, their applicability to technical domains—such as chemical products, where regulatory compliance and technical accuracy are critical—has not been thoroughly investigated. Advertisements for chemical products must effectively communicate complex technical details, including safety protocols and environmental impact, while maintaining a connection with consumers. This work addresses this gap by introducing a framework that combines technical accuracy with personalized, engaging messaging.

\vspace{-2mm}
\subsubsection{Multimodal Data Analysis for Marketing}
\vspace{-1mm}
The integration of multimodal data has enabled more comprehensive analyses of consumer behavior. \citet{Niimi2024} demonstrated the effectiveness of combining textual and demographic data for predicting customer ratings, while \citet{li2024multimodal} highlighted the importance of fusing visual and textual modalities for sentiment analysis. Despite these advances, existing methods often treat modalities independently, missing opportunities to derive holistic insights. For example, while text data can reveal consumer sentiment and image data can assess visual appeal, integrating these insights with video data for emotional engagement and financial data for market trends provides a more complete picture of advertising effectiveness. This framework addresses this limitation by integrating text, images, and video data alongside financial and market insights to comprehensively evaluate advertising effectiveness. This multimodal approach ensures that advertisements are not only visually appealing and impactful but also aligned with market dynamics and consumer preferences.

\vspace{-2mm}
\subsubsection{AI-Driven Consumer Persona Modeling}
\vspace{-1mm}
Persona modeling is essential for effective personalized advertising. \citet{PersonaDoc} proposed a framework for generating personas using qualitative user data and AI, while \citet{RPLA_Survey} explored role-playing language models to simulate diverse personas. Building upon these approaches, this work utilizes a simulated humanistic colony of agents representing distinct consumer personas with well-defined characteristics, enabling the creation of highly targeted advertisements that align with specific consumer preferences and decision-making patterns. By utilizing distinct personas such as the Logical Strategist, Visionary Trailblazer, and Harmonious Connector, this framework captures diverse consumer preferences and characteristics, allowing for the generation of tailored advertisements that align with specific consumer segments.

\vspace{-2mm}
\subsubsection{Challenges in Specialized Domains}
\vspace{-1mm}
Specialized markets, such as chemical products, require precise communication of technical details and adherence to stringent regulatory requirements, including compliance with GHS and REACH standards \citep{hellier2012}. Prior studies have primarily focused on packaging and labeling to address these challenges but often overlook the emotional and cultural dimensions of advertising. For example, while hazard symbols and usage instructions are critical for regulatory compliance, they must be complemented by compelling messaging to effectively engage consumers. This work bridges this gap by integrating regulatory compliance with personalized, culturally relevant messaging, ensuring that advertisements fulfill both technical and consumer-centric objectives. This approach enhances consumer trust and engagement while maintaining adherence to industry standards and regulations. In summary, while existing research has made significant strides in personalized advertising, multimodal analysis, and persona modeling, several gaps remain. The proposed framework addresses these challenges by combining persona modeling, multimodal analysis, and domain-specific considerations, advancing the state of personalized advertising in technical domains such as chemical products. By integrating technical accuracy, consumer engagement, and regulatory compliance, this work provides a comprehensive solution for creating effective and engaging advertisements in specialized markets.

\subsection{Synthetic Experimentation for Scalable Ad Personalization: A Data-Driven Approach to Competitive Advertising}
The proliferation of digital platforms has heightened the demand for hyper-personalized advertising, especially when promoting multiple, same-category products from different companies to the same user. This presents a challenge: how to balance personalization, avoid brand or product cannibalization, and effectively highlight each product's unique selling points (USPs). To address this, we leverage synthetic experiments—a data-driven, scalable, and privacy-compliant approach—to systematically test and optimize advertising strategies. These experiments offer a structured framework for simulating real-world complexities while maintaining control over critical variables. Our synthetic experimentation framework supports competitive advertising by enabling companies to test diverse ad configurations, customer responses, and market scenarios without the costs and risks inherent in large-scale, real-world testing. This approach provides four primary advantages: (a) Controlled Environment for Systematic Testing: Synthetic experiments create a controlled environment where diverse market conditions, consumer personas, and product categories can be accurately simulated. AI-powered optimization allows for dynamic evaluation of different ad strategies, ensuring scalability and adaptability. Unlike real-world campaigns, where external noise (competitor actions, seasonality, or economic fluctuations) complicates causal attribution, synthetic testing isolates variables, allowing for precise measurement of their impact on ad performance. By modeling interactions between competing ads, companies can optimize their delivery strategies, mitigating negative consequences such as audience fatigue or brand dilution. (b) Cost-Effective and Scalable Competitive Testing: Testing competitive advertising strategies in the real world is prohibitively expensive, often requiring extensive A/B testing across multiple market segments. Synthetic experiments eliminate this constraint by simulating and evaluating different advertising permutations at scale. Companies can systematically assess how various combinations of competing products impact consumer decision-making without incurring massive testing costs or exposing campaigns to real-world risks. This is particularly valuable in industries where strategically nuanced product differentiation (e.g., pricing, feature sets, or branding) is critical for market success. (c) Privacy Compliance and Ethical Considerations: With increasing global data regulations such as GDPR and CCPA, real-world consumer data collection poses significant compliance risks. Synthetic data generation enables rigorous testing of advertising models while preserving privacy and security. By eliminating the need for direct reliance on actual consumer data, synthetic experiments allow brands to ethically refine targeting strategies, ensuring compliance with data protection laws. Additionally, they mitigate potential biases and ethical concerns associated with real-world consumer profiling, fostering more responsible AI-driven advertising. (d) Accelerated AI Model Development and Edge Case Testing: The development of AI-powered advertising systems demands rapid iteration, extensive testing, and robust validation across diverse scenarios. Synthetic experiments generate thousands of test cases, including rare but impactful edge cases—such as sudden market disruptions, shifts in consumer sentiment, or emerging trends—that would take years to observe in real-world data. By accelerating testing cycles, companies can refine AI-driven ad personalization models, ensuring their robustness before real-world deployment. This significantly enhances adaptability, allowing brands to respond proactively to evolving market conditions. Our framework ensures that synthetic experiments preserve key statistical properties and behavioral patterns observed in real advertising data, while also enabling controlled manipulation of crucial variables. This methodology provides a strong validation foundation prior to real-world deployment, where mistakes could be costly and irreversible. Ultimately, these synthetic experiments serve as a proof of concept, demonstrating the framework’s capability to generate personalized, competitive ad campaigns tailored to specific customer segments. By bridging the gap between controlled testing and real-world adaptation, synthetic experimentation paves the way for more effective, data-driven advertising strategies that optimize engagement, relevance, and return on investment.

\subsubsection{Framework Architecture}  
The hyper-personalized competitive ad optimization framework is designed as a modular and scalable system that integrates multiple components to tackle the challenge of advertising multiple same-category products from competing companies. This framework is built on synthetic experimentation, incorporating synthetic personas, AI-driven ad generation, and market-aware optimization to create and refine competitive ad strategies. The system consists of seven key components, each playing a crucial role in optimizing ad effectiveness.  
The first component, Market Research Data, provides foundational demographic insights (including information on income range, price sensitivity, and tech-savviness across different age groups) to assess how different customer segments respond to advertising strategies, product category attributes (outlining consumer purchasing behavior, loyalty trends, and the importance of eco-friendly considerations for various products), and competitor intelligence (offering insights into brand perception, market share distribution, and sustainability positioning for leading industry players). The second component, Persona Profiling, defines synthetic customer personas based on a combination of demographic attributes, interests, values, and purchasing behaviors, enabling targeted ad personalization. This module categorizes consumers into distinct lifestyle segments (e.g., eco-conscious, luxury seeker, budget-conscious individuals), ensuring that advertisements are tailored to resonate with their unique preferences. Furthermore, an AI-driven persona affinity scoring system determines how well a specific product aligns with a given persona by analyzing factors such as interest alignment, value compatibility, and income-based price sensitivity.  The third component, Product Analysis, systematically evaluates product features, pricing strategies, and brand positioning to inform competitive advertising strategies. By assessing price sensitivity, consumer loyalty, and brand positioning, this module enables companies to refine their messaging to highlight the most compelling attributes of their offerings. Additionally, this component examines feature uniqueness by comparing a product’s specifications against those of competing products to identify key differentiators.  The fourth component, Competitive Analysis, focuses on identifying unique selling points, price positioning, and brand strength relative to competitors. This module employs market positioning assessment techniques such as the Herfindahl-Hirschman Index (HHI) to determine market concentration levels, competitive advantage scoring to evaluate a brand’s relative strength w.r.t competitors, and category growth simulations to predict long-term trends in consumer demand.  The fifth component, AI-Powered Ad Generation, is responsible for leveraging large language models to generate product advertisements at two distinct levels. The first level involves the creation of base advertisements, which are generic product-focused ad copies highlighting fundamental features and benefits. The second level introduces an optimized ad generation process, where base ads are refined and personalized based on competitive intelligence and persona affinity analysis. This ensures that advertisements are not only feature-driven but also tailored to the interests, values, and pain points of specific consumer segments.  The sixth component, AI-Based Ad Evaluation, is designed to assess the performance of generated ads using advanced AI-driven scoring mechanisms. This evaluation is conducted through two primary methods. The first method utilizes large language models acting as evaluators to score advertisements based on key criteria such as relevance, persuasiveness, emotional impact, clarity, and call-to-action effectiveness. The second method incorporates an AI-powered evaluation model that measures additional attributes such as helpfulness, correctness, coherence, complexity, and verbosity. This dual evaluation system provides a comprehensive assessment of how well an advertisement performs in engaging consumers and driving conversions. The seventh and final component, Simulation and Iterative Ad Optimization, is dedicated to running synthetic market simulations that assess and refine ad strategies be \subsection{Open-Domain Question Answering (ODQA)}
This section presents an advanced Optimized Retrieval-Augmented Generation (RAG) system \cite{lewis2020retrieval, gupta2024comprehensive, gao2023retrieval, han2025graphrag}, designed to process and utilize multimodal agglomerated knowledge from the Multimodal Agentic Advertisement Market Survey (MAAMS) system for question-answering (QA) tasks. The knowledge synthesized by MAAMS comprises market intelligence from diverse modalities, including text, images, video, financial data, and market data, which is consolidated into a document-specific text output format for each product. While the system primarily processes textual knowledge extracted from documents, it enables semantic search and contextually accurate response generation for QA tasks. A QA system built on this agglomerated knowledge base acts as a crucial interface between customers and technical product information, allowing users to receive precise and relevant answers without navigating lengthy documentation. The system excels in four key areas:  (a) Providing direct access to product information; (b) Offering comparative product insights; (c) Translating technical specifications into user-friendly language; (d) Delivering specific usage guidance. By extracting and presenting relevant information in an accessible format, the QA system enhances the customer experience, making technical product information more actionable and user-friendly. The system begins by processing documents using an automated text extraction method, ensuring the structured retrieval of textual content and metadata. The extracted text is segmented into manageable, semantically coherent chunks with overlapping regions to maintain contextual continuity. To improve context awarenesfore they are deployed in real-world campaigns. This simulation process leverages AI-driven iteration cycles to optimize ad performance by testing various combinations of ad messaging, pricing strategies, and competitive positioning. The system continuously tracks performance metrics, allowing advertisers to iteratively refine their campaigns to maximize impact.  By integrating synthetic personas, AI-driven insights, and competitive benchmarking, this framework provides an automated, data-driven, and continuously optimized approach to advertising. It enables product manufacturers to develop highly personalized and competitive ads that drive higher engagement, conversion rates, and stronger market positioning. The framework ensures privacy compliance by relying on synthetic data rather than real consumer information, making it an ethical and scalable solution for modern advertising challenges. The combination of AI-generated ad content, competitive market analysis, and real-time performance tracking enables companies to refine their messaging dynamically, ensuring that advertisements remain relevant in rapidly evolving market conditions. By systematically simulating consumer responses and testing ad variations, this approach minimizes the risks associated with traditional A/B testing and allows companies to experiment with innovative marketing strategies in a controlled environment. This comprehensive, modular, and scalable architecture ensures that businesses can continuously refine their advertising strategies, optimize their market positioning, and maximize the return on investment from their marketing efforts.

\subsubsection{Methodology}
The hyper-personalized competitive ad optimization framework operates as an offline, synthetic experimental setup, where each stage follows a structured pipeline to simulate and analyze ad effectiveness in a controlled environment. The process begins with persona profiling, where synthetic user personas are generated based on demographic attributes (age, income, lifestyle), psychographic traits (interests, values), and behavioral tendencies (purchase frequency, price sensitivity, brand loyalty). These personas serve as structured input representations that guide subsequent stages. Next, product analysis evaluates each product’s characteristics, including functional features (e.g., ``organic ingredients”, ``long-lasting fragrance”), price positioning (premium, mid-range, budget), and brand perception (e.g., sustainability, innovation, quality), yielding a product-persona affinity score that determines how well a product aligns with different personas. The competitive analysis stage then integrates synthetic market data—including competing products, price trends, brand positioning, and market share—to generate a competitive positioning report highlighting differentiation opportunities and pricing strategies tailored to each persona. This analysis feeds into ad generation, where two types of synthetic ads are created: base ads, which focus on general product attributes, and optimized ads, which incorporate persona-specific customization, addressing user pain points (e.g., concerns about harsh chemicals), reinforcing values (e.g., eco-friendliness), and highlighting competitive advantages (e.g., larger volume or better pricing). These synthetic ads undergo AI-powered evaluation using GPT-4 Omni and NVIDIA Nemotron-4-340B, which assess each ad based on relevance, persuasiveness, emotional impact, clarity, and call-to-action effectiveness, producing a scoring matrix ranking ads based on their effectiveness for different personas. Finally, the ad ranking and deployment phase prioritizes ads based on their evaluation scores, persona-product affinity, and competitive positioning, ensuring that each persona is matched with the most compelling advertisement. As this is a synthetic experiment rather than a real-time system, ad performance feedback is not derived from live deployment but is instead simulated using synthetic persona responses and AI-based evaluation metrics. The system iterates offline, refining persona profiles, product analysis, and ad optimization in a batch-based process, ensuring that insights from synthetic experiments can inform future strategies for real-world deployment.

\subsubsection{Example Scenario}
To illustrate the framework’s functionality, consider a scenario involving three shampoos from competing companies, each targeting distinct customer personas:  
\textbf{PureGlow Shampoo} (premium, eco-friendly),  
\textbf{EconoFresh Shampoo} (budget-friendly), and  
\textbf{LuxeSilk Shampoo} (luxury, high-quality).  The framework systematically applies \textit{synthetic experimentation, competitive analysis, and AI-driven ad optimization} to create targeted messaging that maximizes engagement for each persona while ensuring competitive differentiation.

\begin{itemize}
    \item \textbf{Persona A: Eco-Conscious Professional}  
    \begin{itemize}
        \item \textbf{Demographic \& Lifestyle Insights}: Values sustainability, prefers plant-based products, and seeks a shampoo free from harsh chemicals.
        \item \textbf{Competitive Positioning}:  
        Compared to standard shampoos, PureGlow Shampoo offers an \textbf{eco-friendly, sulfate-free formula} that outperforms competitors by using \textbf{100\% biodegradable packaging and ethically sourced ingredients}. Unlike \textit{EconoFresh}, which contains synthetic cleansers, and \textit{LuxeSilk}, which focuses on premium aesthetics, PureGlow directly aligns with this persona's sustainability values.
        \item \textbf{Generated Ad:}  
        \begin{quote}
            \textit{"Choose a shampoo that cares for your hair—and the planet. PureGlow Shampoo is crafted with organic botanicals and zero synthetic additives, delivering a natural shine while protecting the environment. Make the switch today!"}
        \end{quote}
    \end{itemize}

    \item \textbf{Persona B: Budget-Conscious Student}  
    \begin{itemize}
        \item \textbf{Demographic \& Lifestyle Insights}: Prefers affordable products that offer long-lasting use and strong cleansing power.
        \item \textbf{Competitive Positioning}:  
        EconoFresh Shampoo delivers \textbf{cost-effective, high-foaming cleansing} that lasts 30\% longer than comparable brands. While \textit{PureGlow} is focused on premium eco-conscious buyers and \textit{LuxeSilk} on luxury aesthetics, EconoFresh prioritizes \textbf{affordability and value for money}.
        \item \textbf{Generated Ad:}  
        \begin{quote}
            \textit{"Stretch your budget further without compromising quality! EconoFresh Shampoo provides a deep clean with a refreshing scent, designed for daily use at an unbeatable price. Get more washes for less!"}
        \end{quote}
    \end{itemize}

    \item \textbf{Persona C: Luxury-Seeking Professional}  
    \begin{itemize}
        \item \textbf{Demographic \& Lifestyle Insights}: Prefers high-quality formulations enriched with premium ingredients for a refined haircare experience.
        \item \textbf{Competitive Positioning}:  
        LuxeSilk Shampoo is \textbf{infused with silk proteins and botanical extracts}, offering \textbf{superior hydration and shine} compared to competitors. Unlike \textit{PureGlow}, which prioritizes natural ingredients, or \textit{EconoFresh}, which focuses on affordability, LuxeSilk is designed for consumers who seek an \textbf{indulgent, high-performance formula}.
        \item \textbf{Generated Ad:}  
        \begin{quote}
            \textit{"Unleash the power of luxury with LuxeSilk Shampoo. Infused with silk proteins, this rich formula deeply nourishes each strand, leaving your hair irresistibly soft, radiant, and elegantly fragrant. Elevate your routine today!"}
        \end{quote}
    \end{itemize}
\end{itemize}

This scenario demonstrates how the framework \textbf{dynamically optimizes ad personalization} using \textit{persona-product affinity analysis, competitive positioning, and AI-driven ad evaluation}. By integrating these insights, the system \textbf{maximizes ad relevance and effectiveness}, ensuring each product is strategically positioned against its competitors.

\subsubsection{Results}
\begin{figure*}[htbp] 
    \centering 
    \includegraphics[
        width=1\textwidth, 
        keepaspectratio, 
        trim=0mm 0mm 0mm 20mm, 
        clip  
    ]{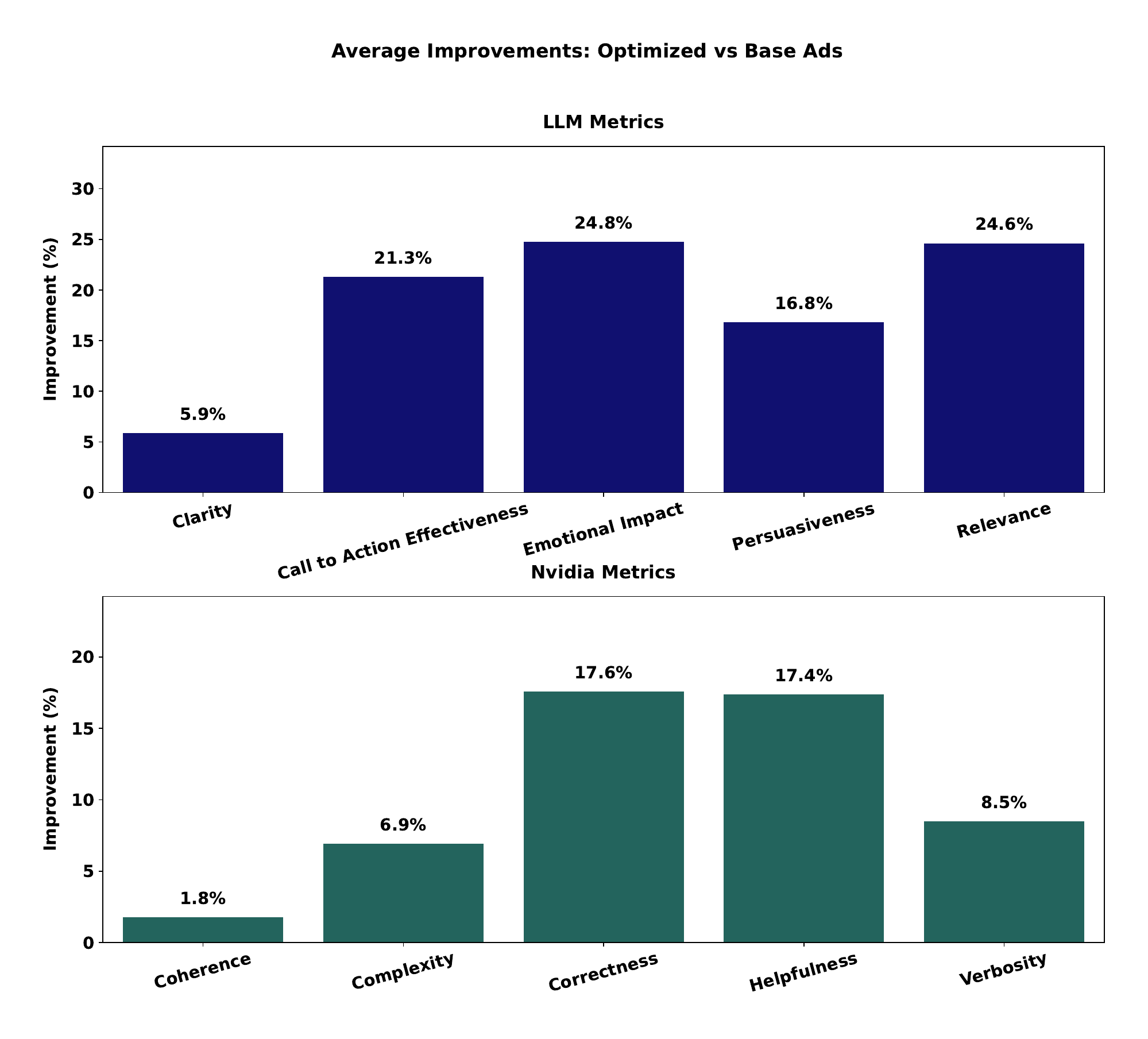} 
     \vspace{-10mm}
    \caption{Average percentage improvements of hyper-personalized ads over base ads across all products and personas. The figure presents the evaluation results from LLM-as-a-Judge (GPT-4o) and reward model (Nemotron-4-340b-reward), showing improvements in key performance indicators such as \textbf{clarity}, \textbf{call-to-action effectiveness}, \textbf{emotional impact}, \textbf{persuasiveness}, and \textbf{relevance} (LLM metrics) as well as \textbf{coherence}, \textbf{complexity}, \textbf{correctness}, \textbf{helpfulness}, and \textbf{verbosity} (reward model metrics). Hyper-personalized ads demonstrate consistent improvements across all dimensions.}
    \label{fig:FigureAppendix1} 
\end{figure*}

\begin{figure*}[htbp] 
    \centering 
    \includegraphics[
        width=1\textwidth, 
        keepaspectratio, 
        trim=0mm 0mm 0mm 18mm, 
        clip  
    ]{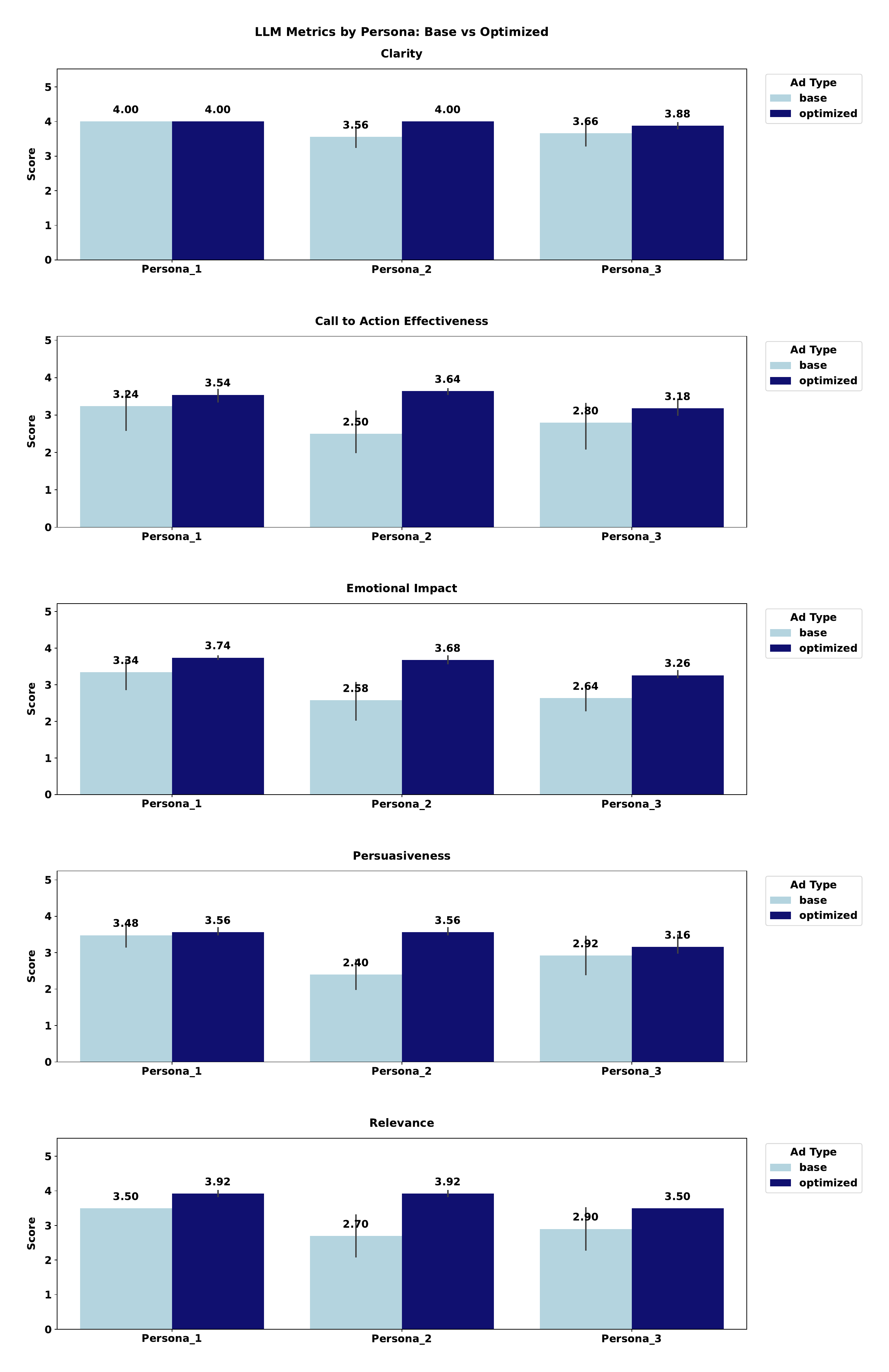} 
    \vspace{-10mm}
    \caption{Comparison of base vs. hyper-personalized ads using LLM-as-a-Judge evaluation (GPT-4o) across a subset of personas. This figure presents the performance comparison between base and hyper-personalized ads assessed by GPT-4o, measuring clarity, call-to-action effectiveness, emotional impact, persuasiveness, and relevance. Results are shown for a subset of personas to illustrate the impact of hyper-personalization across different user segments.}
    \label{fig:FigureAppendix2} 
\end{figure*}

\begin{figure*}[htbp] 
    \centering 
    \includegraphics[
        width=1\textwidth, 
        keepaspectratio, 
        trim=0mm 0mm 0mm 18mm, 
        clip  
    ]{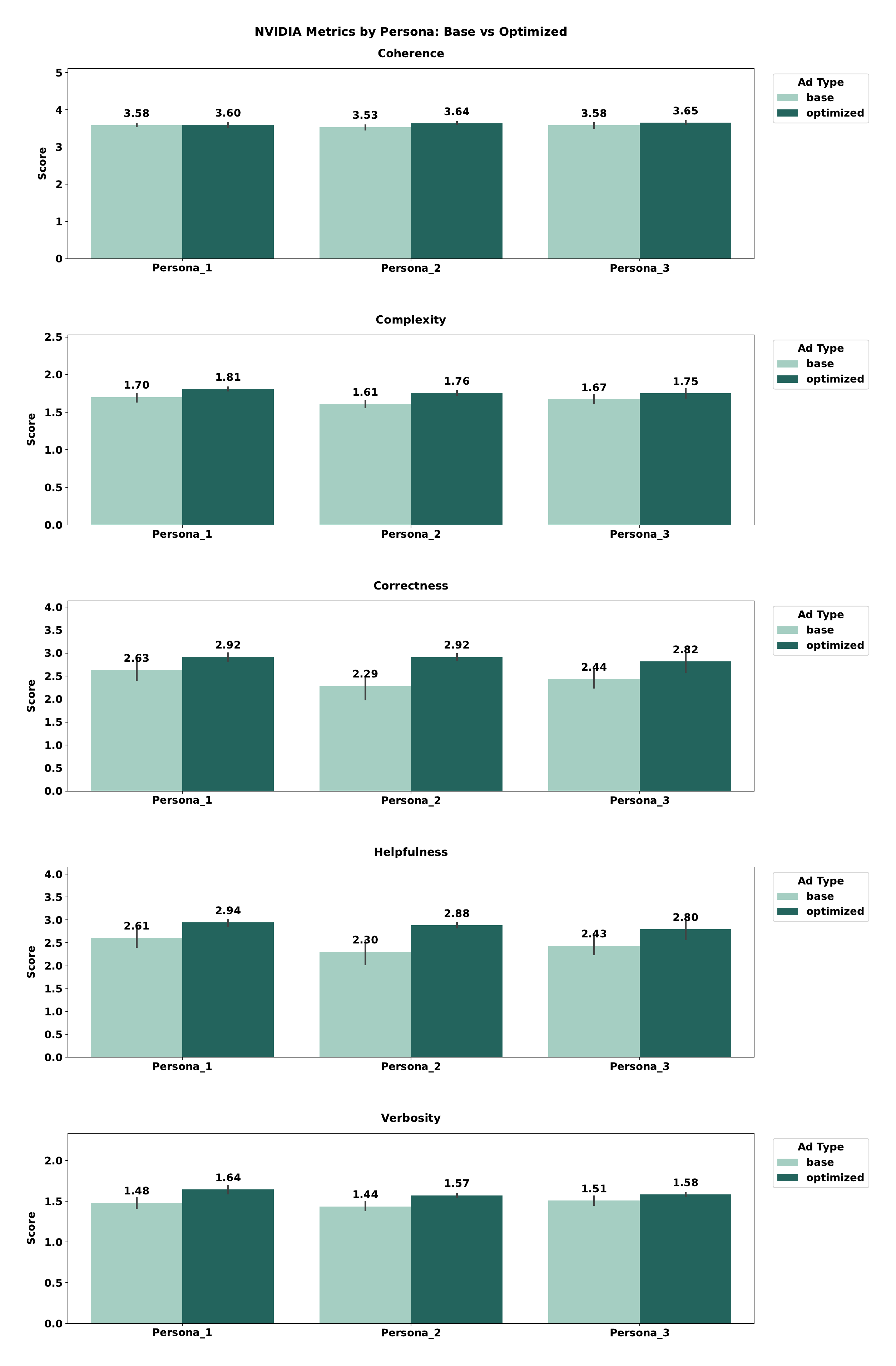} 
    \vspace{-5mm}
    \caption{Comparison of base vs. hyper-personalized ads using reward model evaluation (Nemotron-4-340b-reward) across a subset of personas. Evaluated using the Nemotron-4-340b-reward model, this figure compares coherence, helpfulness, and correctness across personas. The results represent a subset of personas to highlight how AI-driven ad personalization improves engagement and message effectiveness.}
    \label{fig:FigureAppendix3} 
\end{figure*}

\begin{figure*}[htbp] 
    \centering 
    \includegraphics[
        width=1\textwidth, 
        keepaspectratio, 
        trim=0mm 0mm 0mm 18mm, 
        clip  
    ]{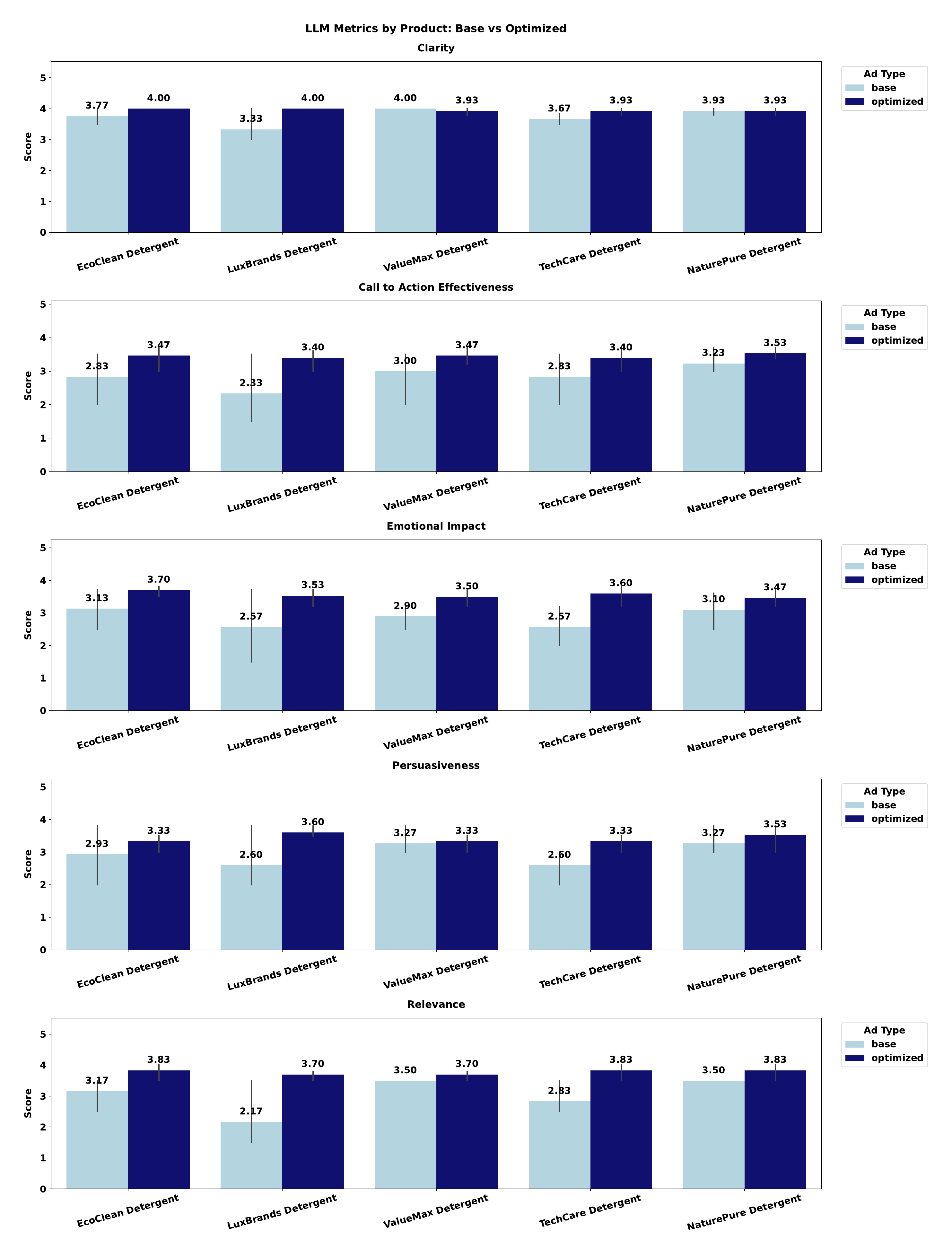} 
    \vspace{-10mm}
    \caption{Comparison of base vs. hyper-personalized ads using LLM-as-a-Judge evaluation (GPT-4o) across a subset of products. This figure evaluates base and hyper-personalized ads across selected product categories using GPT-4o. Improvements in clarity, call-to-action effectiveness, emotional impact, persuasiveness, and relevance reinforce the benefits of AI-driven hyper-personalization for product-specific advertising strategies. The results focus on a subset of products to provide a representative analysis.}
    \label{fig:FigureAppendix4} 
\end{figure*}

\begin{figure*}[htbp] 
    \centering 
    \includegraphics[
        width=1\textwidth, 
        keepaspectratio, 
        trim=0mm 0mm 0mm 18mm, 
        clip  
    ]{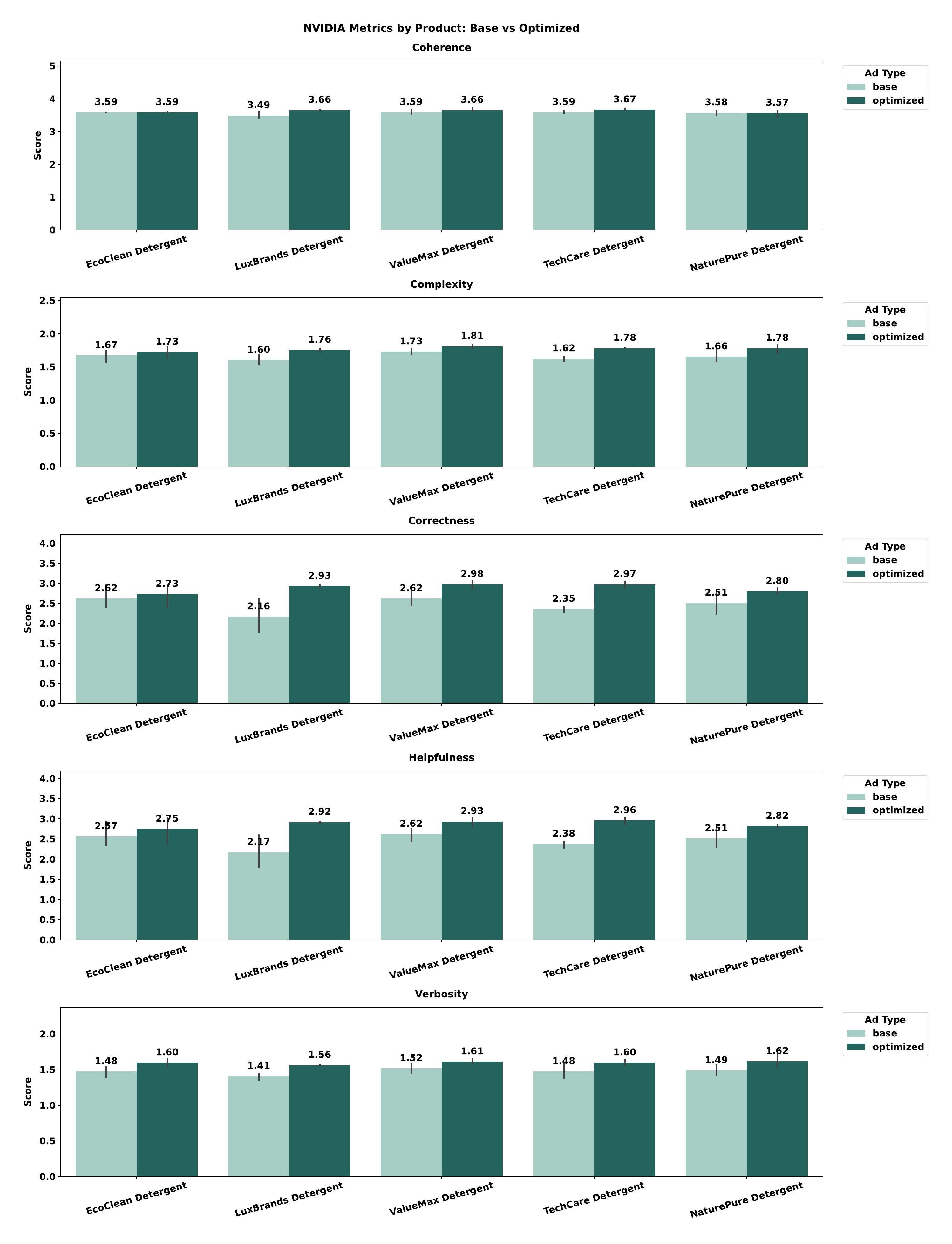} 
    \vspace{-10mm}
    \caption{Comparison of base vs. hyper-personalized ads using reward model evaluation (Nemotron-4-340b-reward) across a subset of products. This figure presents performance differences between base and hyper-personalized ads evaluated with the Nemotron-4-340b-reward model. Improvements in coherence, helpfulness, and correctness indicate the effectiveness of AI-driven hyper-personalized ad strategies tailored to selected product categories. Results are presented for a subset of products to illustrate the general trends in ad performance.}
    \label{fig:FigureAppendix5} 
\end{figure*}

We evaluate the AI-driven hyper-personalized competitive ad optimization framework for creating highly personalized and effective advertisements and present the empirical results in Figures \ref{fig:FigureAppendix1}--\ref{fig:FigureAppendix5}. Figures \ref{fig:FigureAppendix2}--\ref{fig:FigureAppendix3} provide a high-level comparison of base ads versus optimized ads across multiple personas, evaluated using the Nvidia Nemotron-4-340b-reward model (e.g., in terms of attributes like \textbf{coherence}, \textbf{complexity}, \textbf{correctness}, \textbf{helpfulness}, and \textbf{verbosity}) and LLM-as-a-Judge Metrics (e.g., \textbf{clarity}, \textbf{call-to-action effectiveness}, \textbf{emotional impact}, \textbf{persuasiveness}, and \textbf{relevance}) to assess and score the quality of generated advertisements. The results demonstrate that optimized ads consistently outperform base ads across all metrics, highlighting the effectiveness of hyper-personalization and competitive optimization. In Figure \ref{fig:FigureAppendix2}, metrics such as \textbf{Clarity} (4.00 vs. 3.56), \textbf{Call-to-Action Effectiveness} (3.54 vs. 2.50), and \textbf{Emotional Impact} (3.74 vs. 2.58) show notable improvements for optimized ads (see; Figure \ref{fig:FigureAppendix2}) . For instance, \textbf{Emotional Impact} scores are higher, indicating that optimized ads resonate more deeply with the personas' values and pain points. \textbf{Persuasiveness} (3.56 vs. 2.40) and \textbf{Relevance} (3.92 vs. 2.70) also see improvements, demonstrating that optimized ads are more compelling and better tailored to the target audience. In Figure \ref{fig:FigureAppendix3}, the \textbf{Coherence} metric (3.65 vs. 3.53) shows significant improvement for optimized ads, indicating enhanced clarity and logical flow. Similarly, other Nvidia evaluation criteria such as \textbf{Helpfulness} (2.94 vs. 2.30) and \textbf{Correctness} (2.92 vs. 2.29) also reflect higher scores for optimized ads, suggesting better alignment with personas' needs and expectations (refer; Figure \ref{fig:FigureAppendix3}). These results underscore the value of hyper-personalization and competitive optimization in ad creation, as optimized ads consistently deliver superior performance across both the Nvidia Reward model and LLM-as-a-Judge evaluation frameworks. Figures \ref{fig:FigureAppendix4}-\ref{fig:FigureAppendix5} analyze ad performance across different products, while earlier Figures \ref{fig:FigureAppendix2}-\ref{fig:FigureAppendix3} focused on ad performance across different personas. The core message remains consistent: optimized ads consistently outperform base ads across all metrics, delivering clearer, more compelling, and more relevant messaging, which drives better engagement and performance. This consistency across both personas and products demonstrates the robustness and scalability of the hyper-personalization and competitive optimization framework. The product-level analysis provides additional insights into how different products benefit from optimization, further validating the effectiveness of the approach. Figure \ref{fig:FigureAppendix1} compares the performance of optimized ads versus base ads across LLM Metrics and Nvidia Metrics. In the LLM Metrics section, \textbf{Relevance} (24.8\%) and \textbf{Emotional Impact} (24.6\%) show the highest improvements, indicating that optimized ads better align with audience needs and evoke stronger emotional responses. \textbf{Call to Action Effectiveness} improves by 16.8\%, making optimized ads more compelling, while \textbf{Clarity} has the smallest improvement at 5.9\%, suggesting base ads were already clear (see; Figure \ref{fig:FigureAppendix1}). In the Nvidia Metrics section, \textbf{Correctness} improves the most at 17.6\%, reflecting improved factual accuracy, followed by \textbf{Helpfulness} (8.5\%) and \textbf{Verbosity} (7.4\%). \textbf{Complexity} and \textbf{Coherence} show the lowest improvements at 6.9\% and 1.8\%, respectively, indicating base ads were already coherent and appropriately complex (see; Figure \ref{fig:FigureAppendix1}).  Overall, optimized ads perform better, with the most significant gains in relevance, emotional impact, and correctness, while clarity and coherence remain strong areas with limited room for improvement. This highlights the effectiveness of optimization in enhancing audience engagement and action.  The hyper-personalized competitive ad optimization framework offers several critical insights for companies:
\textbf{Personalization is key}: Tailoring ads to each persona’s unique characteristics ensures maximum relevance and engagement. \textbf{Leveraging competitive advantages}: Highlighting unique selling points (e.g., sustainability for eco-conscious personas or affordability for budget-conscious ones) helps products stand out. \textbf{AI-powered ad evaluation}: Using AI-driven metrics (relevance, persuasiveness, emotional impact) ensures continuous improvement. \textbf{Strategic deployment}: Prioritizing the most relevant ads for each persona optimizes engagement. \textbf{Competitive differentiation}: Companies can emphasize product advantages to maximize positioning in the market. By following these principles, companies can effectively advertise multiple same-category products, minimizing cannibalization and maximizing ad performance.

\begin{table}[htbp]
\caption{Comparative Analysis of EcoClean Detergent Advertisements: Content Personalization Across Three Target Demographics}
\label{tab:ad-comparison}
\small
\begin{tabular}{>{\raggedright\arraybackslash}p{0.12\textwidth} >{\raggedright\arraybackslash}p{0.32\textwidth} >{\raggedright\arraybackslash}p{0.32\textwidth} >{\raggedright\arraybackslash}p{0.30\textwidth}}
\toprule
\multicolumn{1}{c}{\cellcolor{headercolor}\textcolor{white}{\textbf{JSON Field}}} & 
\multicolumn{1}{c}{\cellcolor{headercolor}\textcolor{white}{\textbf{Eco-Conscious Luxury (Persona 1)}}} & 
\multicolumn{1}{c}{\cellcolor{headercolor}\textcolor{white}{\textbf{Active Lifestyle (Persona 2)}}} & 
\multicolumn{1}{c}{\cellcolor{headercolor}\textcolor{white}{\textbf{Premium Innovation (Persona 3)}}} \\
\midrule

\cellcolor{persona1color}\textbf{headline} & 
\cellcolor{persona1color}Transform Your Laundry with EcoClean Detergent - The Luxury Choice for the Eco-Conscious & 
\cellcolor{persona2color}Affordable EcoClean Detergent: Perfect for Your Active Lifestyle & 
\cellcolor{persona3color}Elevate Your Laundry Routine with EcoClean Detergent - Luxury Meets Sustainability \\
\midrule

\cellcolor{persona1color}\textbf{subheadline} & 
\cellcolor{persona1color}Discover an efficient, eco-friendly cleaning solution that saves money and protects the planet. & 
\cellcolor{persona2color}Clean clothes, eco-friendly choices, all within your budget! & 
\cellcolor{persona3color}Innovative cleaning that safeguards your wardrobe and the environment. \\
\midrule

\cellcolor{persona1color}\textbf{mainBody} & 
\cellcolor{persona1color}Introducing EcoClean Detergent - where luxury meets sustainability. Designed for discerning individuals like you, EcoClean offers unparalleled cleaning performance without compromising on your values. Our eco-friendly formula not only penetrates deep into your favorite fabrics, ensuring they stay vibrant and fresh, but it also comes in minimal, recyclable packaging to reduce waste. Enjoy the peace of mind that comes with knowing you're making a smart choice for both your wardrobe and our environment. Priced at just \$20.00, EcoClean embodies efficiency and luxury, making it the perfect fit for your eco-conscious lifestyle and commitment to savings. & 
\cellcolor{persona2color}As someone who values health and sustainability, you deserve a detergent that aligns with your lifestyle. EcoClean Detergent offers premium cleaning power at just \$20.00, making it a smart investment for your budget-conscious routine. Our eco-friendly formula not only cares for your clothes but also protects the planet, ensuring you can focus on your fitness journey without compromising on quality or performance. Experience the difference of natural ingredients that keep your activewear fresh and ready for your next workout while being gentle on both your family's health and the environment. & 
\cellcolor{persona3color}Meet EcoClean Detergent, the premium choice for conscious consumers who value both luxury and innovation. Our advanced formula not only delivers unbeatable cleaning power but also provides long-lasting protection for your favorite fabrics, ensuring they look new wash after wash. You deserve a detergent that not only meets your high standards but also aligns with your commitment to sustainability. With EcoClean, every wash is a step towards a cleaner planet, without compromising on quality. Experience the perfect blend of performance and environmental responsibility—because you shouldn't have to choose between luxury and sustainability. \\
\midrule

\cellcolor{persona1color}\textbf{callToAction} & 
\cellcolor{persona1color}Join the movement towards a cleaner, greener future. Order EcoClean today and elevate your laundry experience! & 
\cellcolor{persona2color}Make the switch to EcoClean today and enjoy a cleaner, greener laundry experience that fits your budget! & 
\cellcolor{persona3color}Invest in Quality for Your Home - Shop Now for Only \$20.00! \\
\midrule

\cellcolor{persona1color}\textbf{targetedBenefits} & 
\cellcolor{persona1color}
\begin{compactitemize}
\item Exceptional cleaning power that outperforms traditional detergents, giving you more value for your money
\item Eco-friendly ingredients and minimal packaging that align with your values and concern for the environment
\item Long-lasting protection for your clothes, helping you save on replacements while keeping them looking new
\end{compactitemize} & 
\cellcolor{persona2color}
\begin{compactitemize}
\item Exceptional cleaning power that beats traditional detergents while being easy on your wallet
\item Eco-friendly ingredients that prioritize your health and the well-being of the planet
\item Long-lasting freshness for your workout gear, so you can feel confident and ready for anything
\end{compactitemize} & 
\cellcolor{persona3color}
\begin{compactitemize}
\item Unmatched cleaning performance that rivals even the strongest brands
\item Long-lasting fabric protection that saves you money over time
\item Eco-friendly ingredients that resonate with your values of innovation and luxury
\end{compactitemize} \\
\bottomrule
\end{tabular}
\end{table}

\clearpage
\newpage

\subsubsection{Visualization Analysis of Synthetic Marketing Data}
\begin{figure}[htbp]
    \centering
    \includegraphics[width=0.8\textwidth]{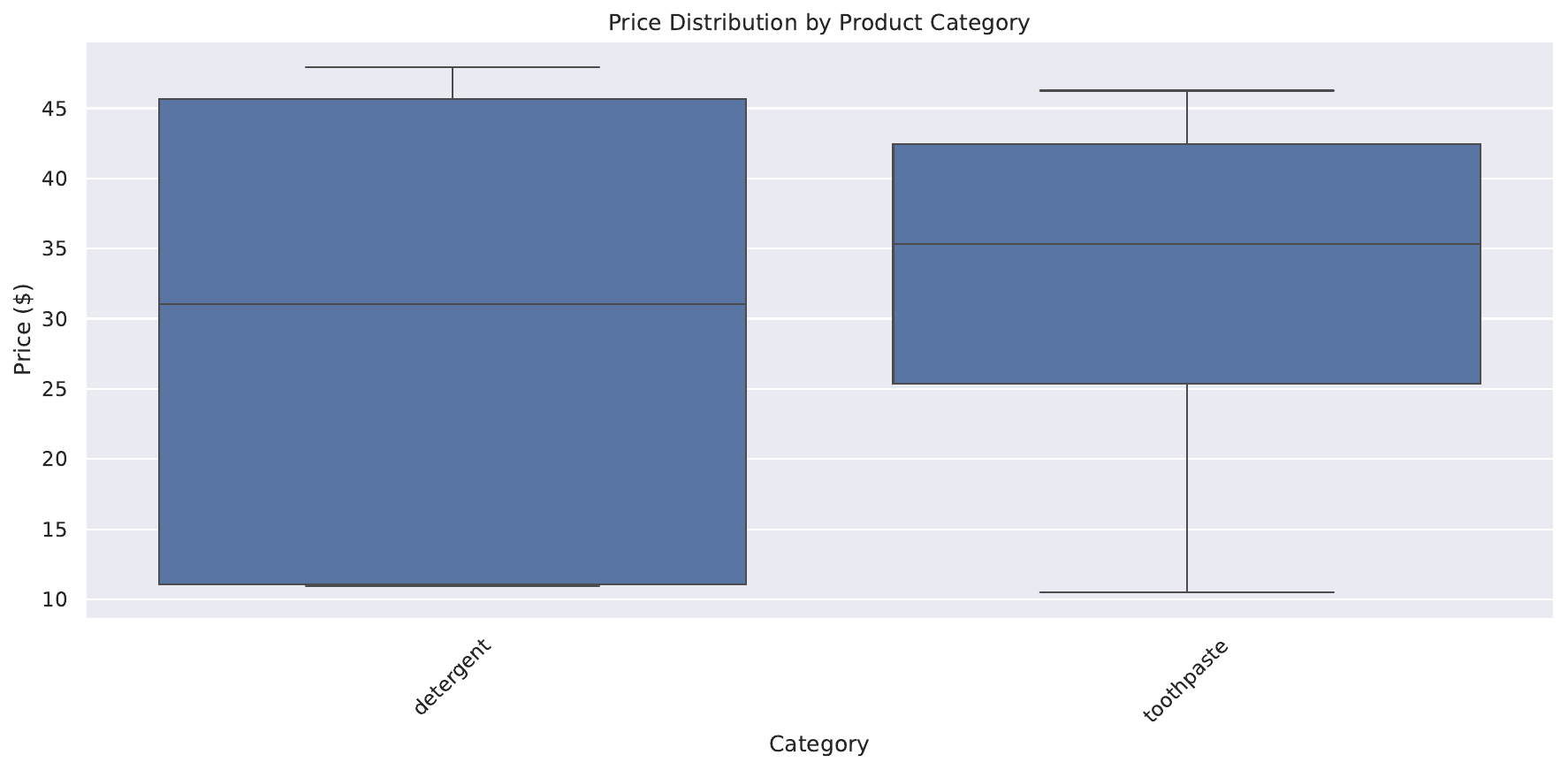}
    \caption{Price Distribution by Product Category. This visualization reveals the price variation within detergent and toothpaste categories, helping identify pricing patterns and potential market positioning opportunities. The box plots show median prices, quartiles, and outliers, enabling strategic pricing decisions based on category-specific market dynamics.}
    \label{fig:price_distribution}
\end{figure}

\begin{figure}[htbp]
    \centering
    \includegraphics[width=0.8\textwidth]{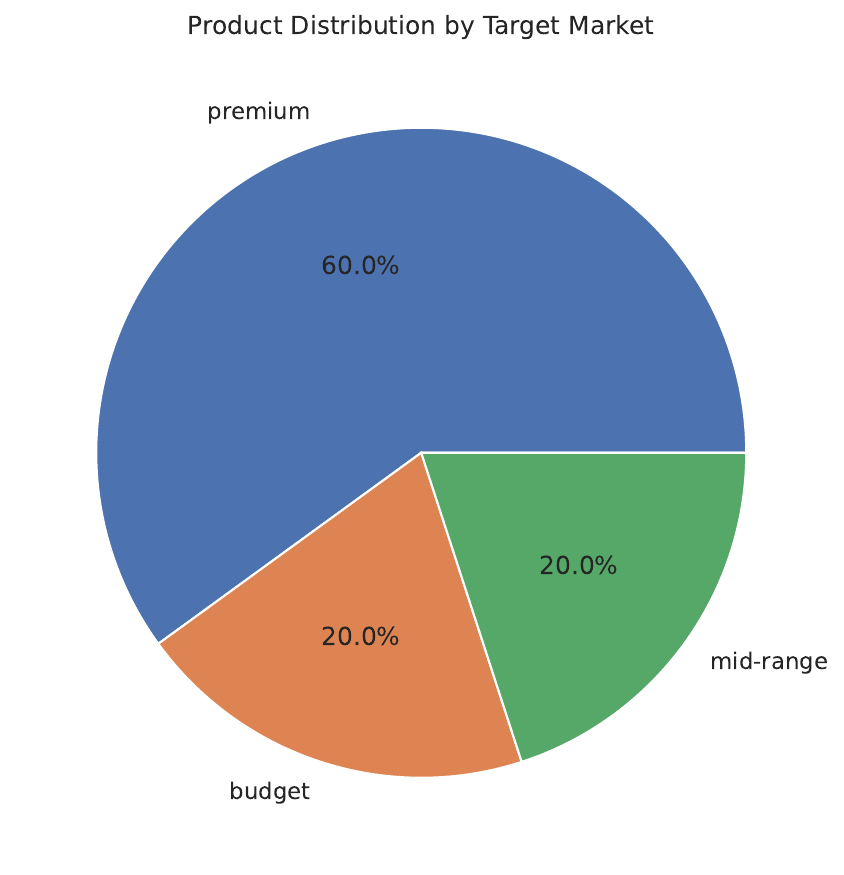}
    \caption{Product Distribution by Target Market. A pie chart showing the proportion of products targeting different market segments (premium, budget, mid-range). This visualization helps identify potential market gaps and assess the balance of product offerings across different consumer segments.}
    \label{fig:target_market}
\end{figure}

\begin{figure}[htbp]
    \centering
    \includegraphics[width=0.8\textwidth]{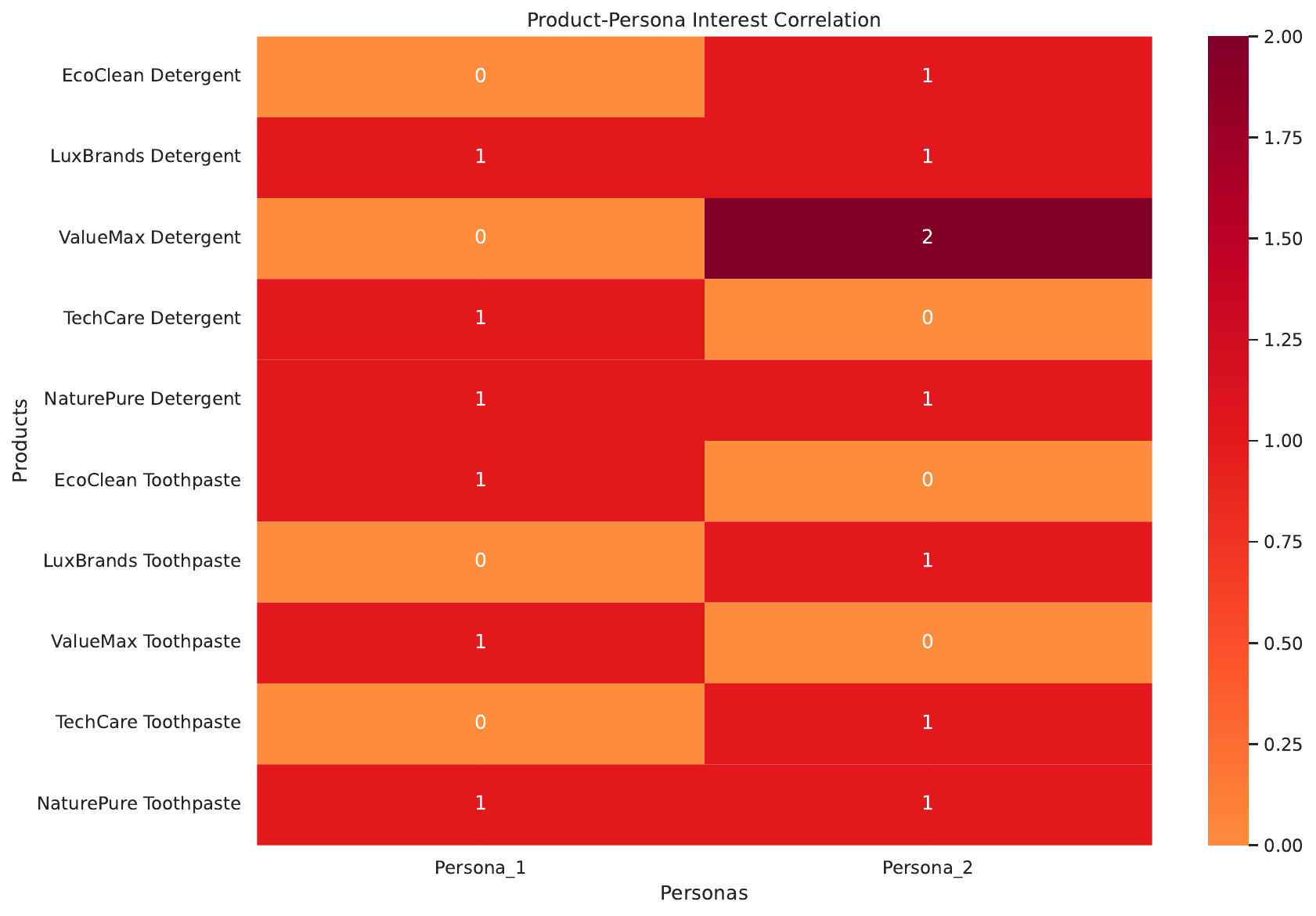}
    \caption{Product-Persona Interest Correlation Heatmap. This visualization maps the alignment between product target interests and persona interests, highlighting potential marketing opportunities and product-customer fit. Darker colors indicate stronger alignment between products and specific customer personas.}
    \label{fig:interest_correlation}
\end{figure}

\begin{figure}[htbp]
    \centering
    \includegraphics[width=0.8\textwidth]{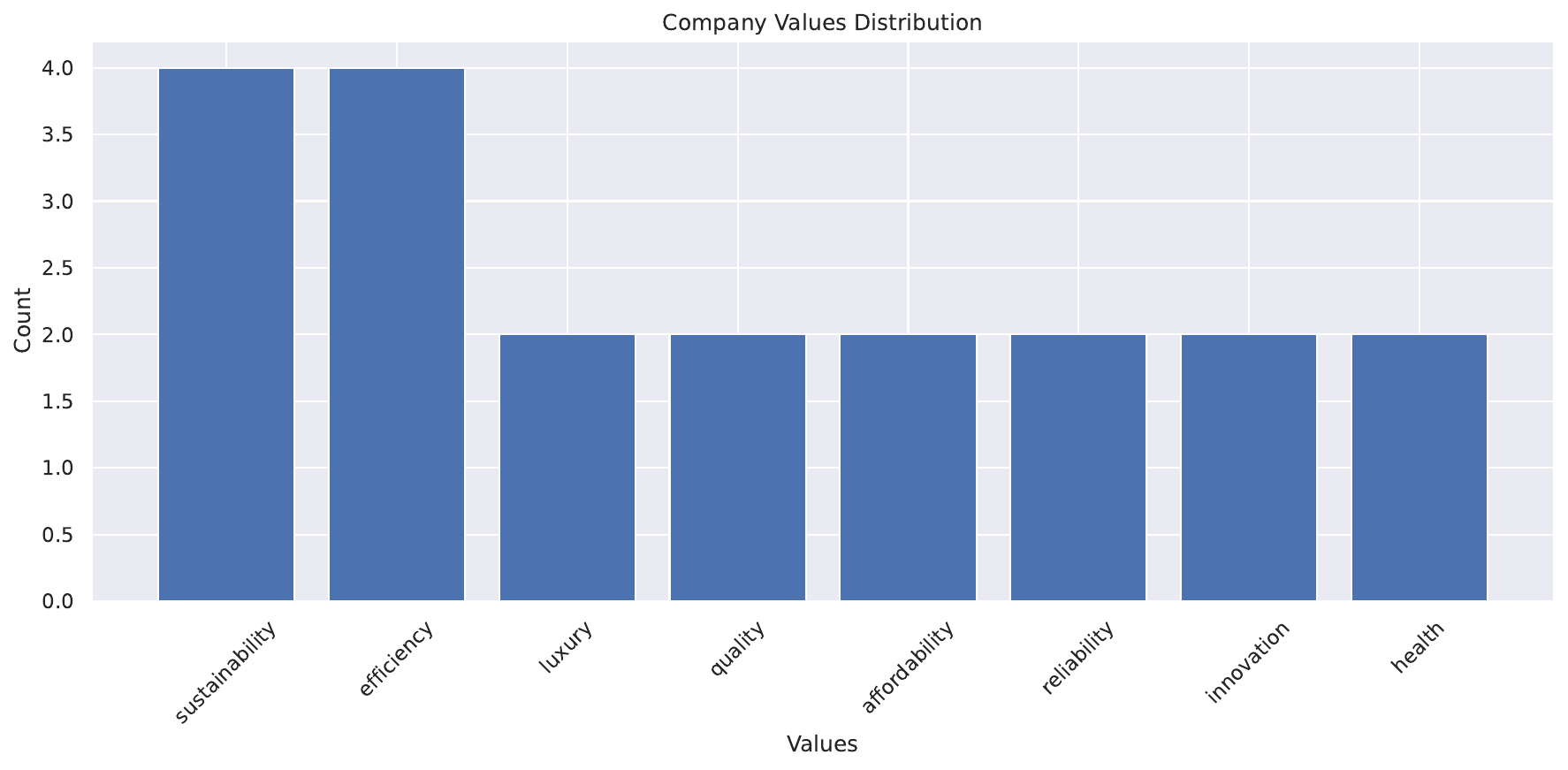}
    \caption{Company Values Distribution. Bar chart showing the frequency of different company values across the product portfolio. This visualization helps understand the overall brand positioning and identify potential gaps in value propositions, guiding future marketing and branding strategies.}
    \label{fig:value_proposition}
\end{figure}

\begin{figure}[htbp]
    \centering
    \includegraphics[width=0.8\textwidth]{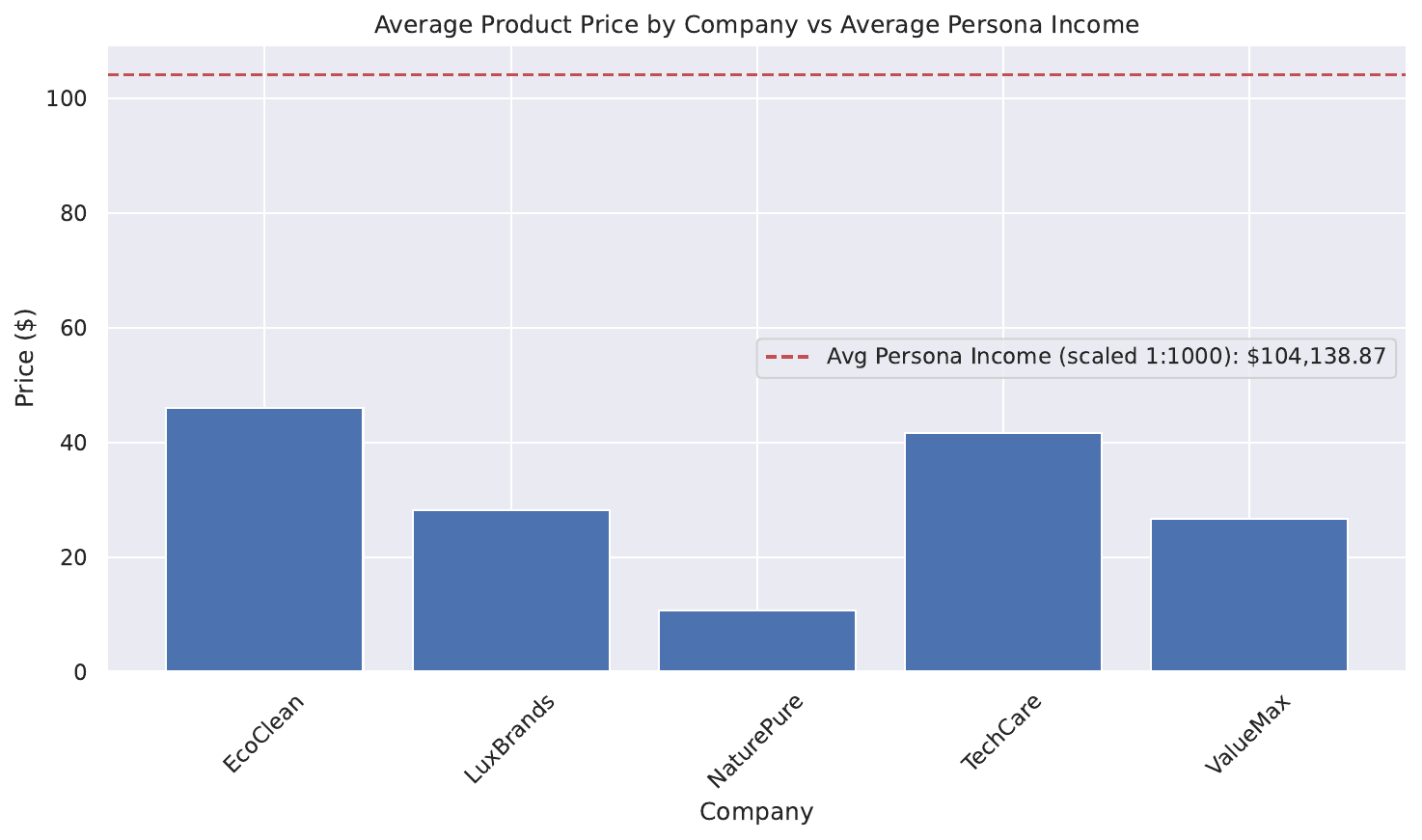}
    \caption{Average Product Price by Company vs Average Persona Income. This comparison between company pricing strategies and customer income levels helps assess product affordability and market positioning. The red dashed line represents scaled average persona income, providing context for pricing decisions.}
    \label{fig:price_vs_income}
\end{figure}

\begin{figure}[htbp]
    \centering
    \includegraphics[width=0.8\textwidth]{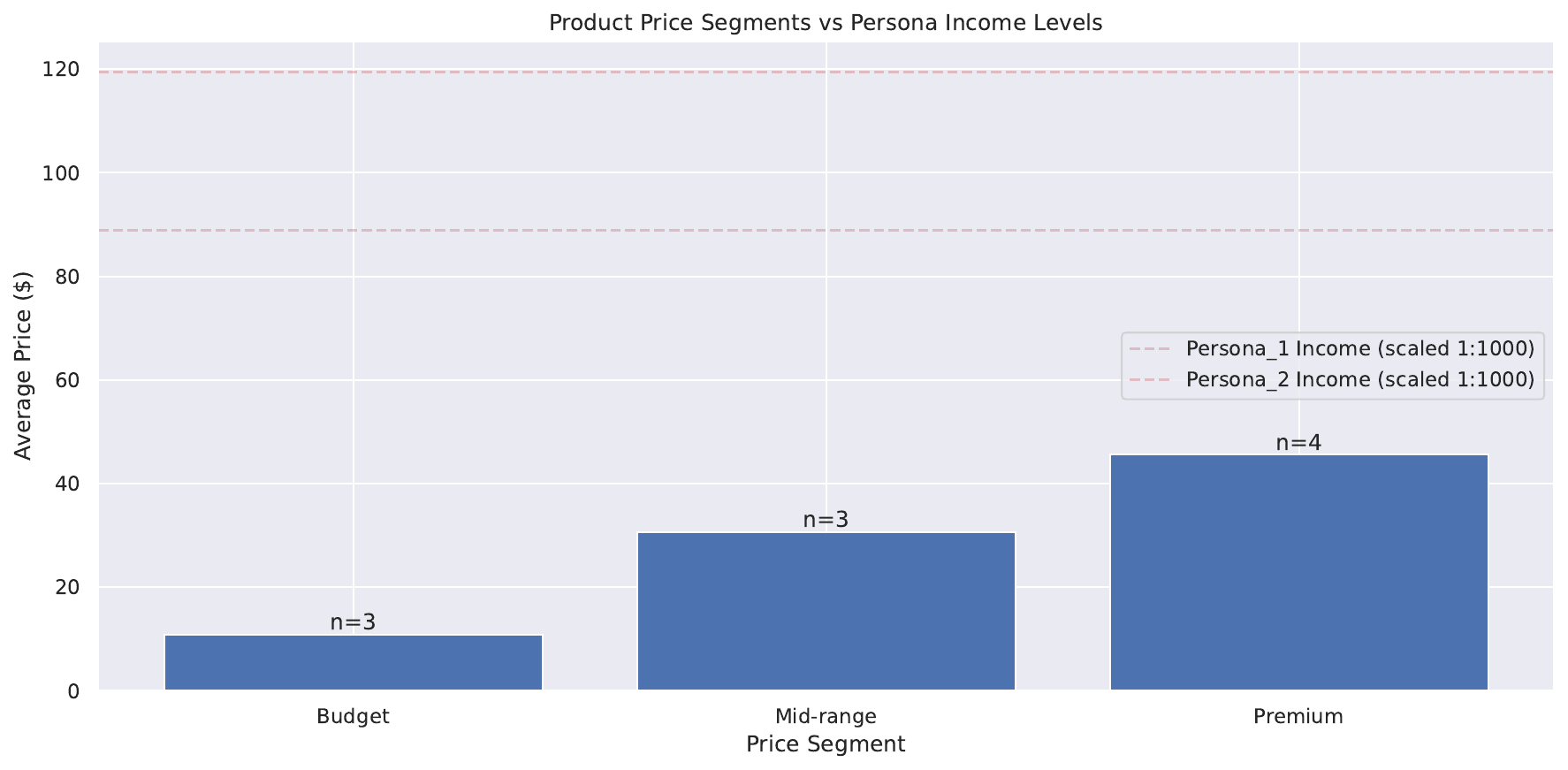}
    \caption{Price Segments vs Persona Income Levels. This visualization segments products into price tiers (Budget, Mid-range, Premium) and overlays persona income levels, revealing how well current pricing aligns with customer purchasing power. Bar annotations show the number of products in each segment.}
    \label{fig:price_segments}
\end{figure}

\begin{figure}[htbp]
    \centering
    \includegraphics[width=0.8\textwidth]{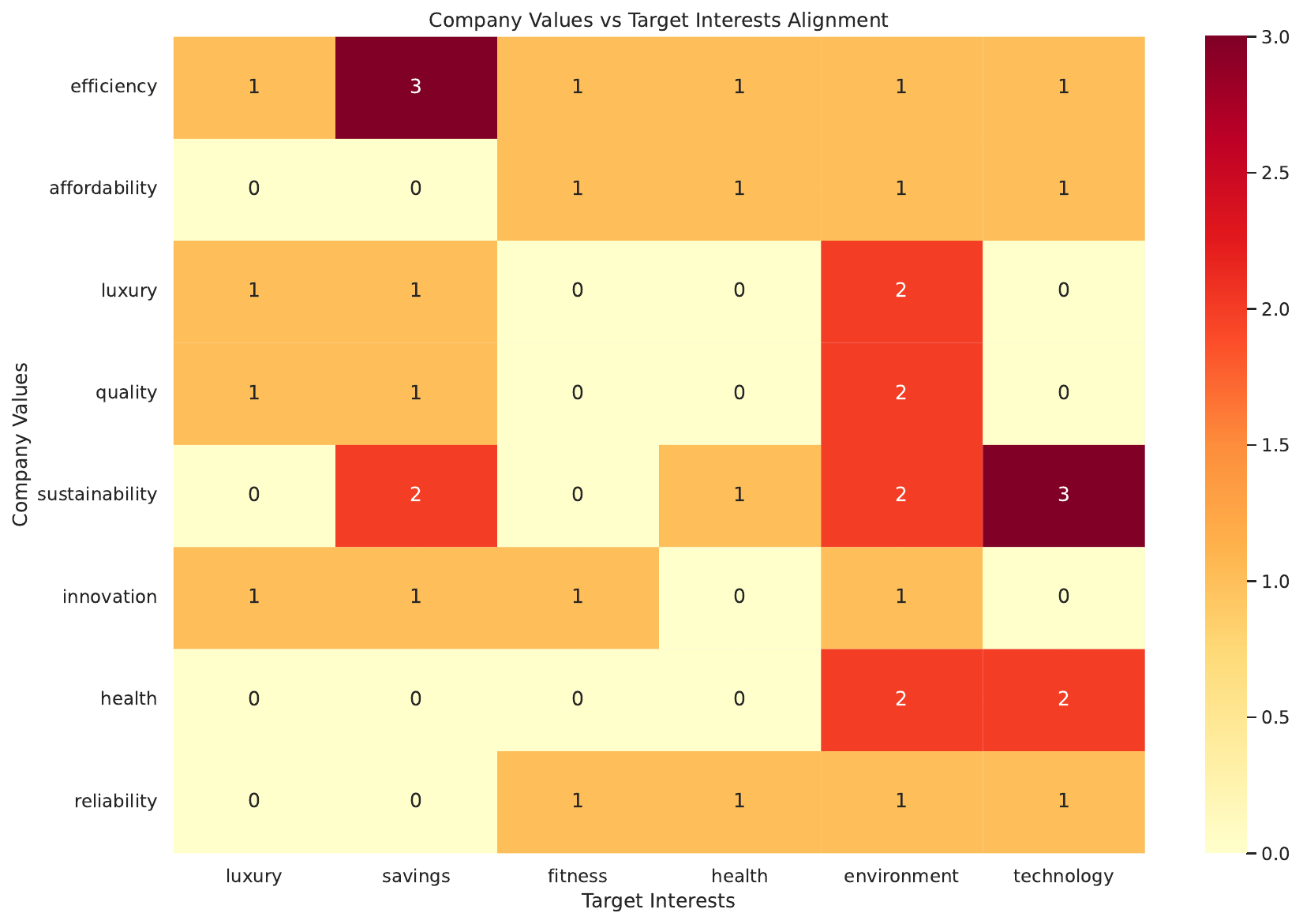}
    \caption{Company Values vs Target Interests Alignment Heatmap. This visualization shows how company values align with target customer interests, revealing opportunities for more effective marketing messaging. The intensity of colors indicates the strength of alignment between values and interests.}
    \label{fig:value_interest}
\end{figure}

\begin{figure}[htbp]
    \centering
    \includegraphics[width=0.8\textwidth]{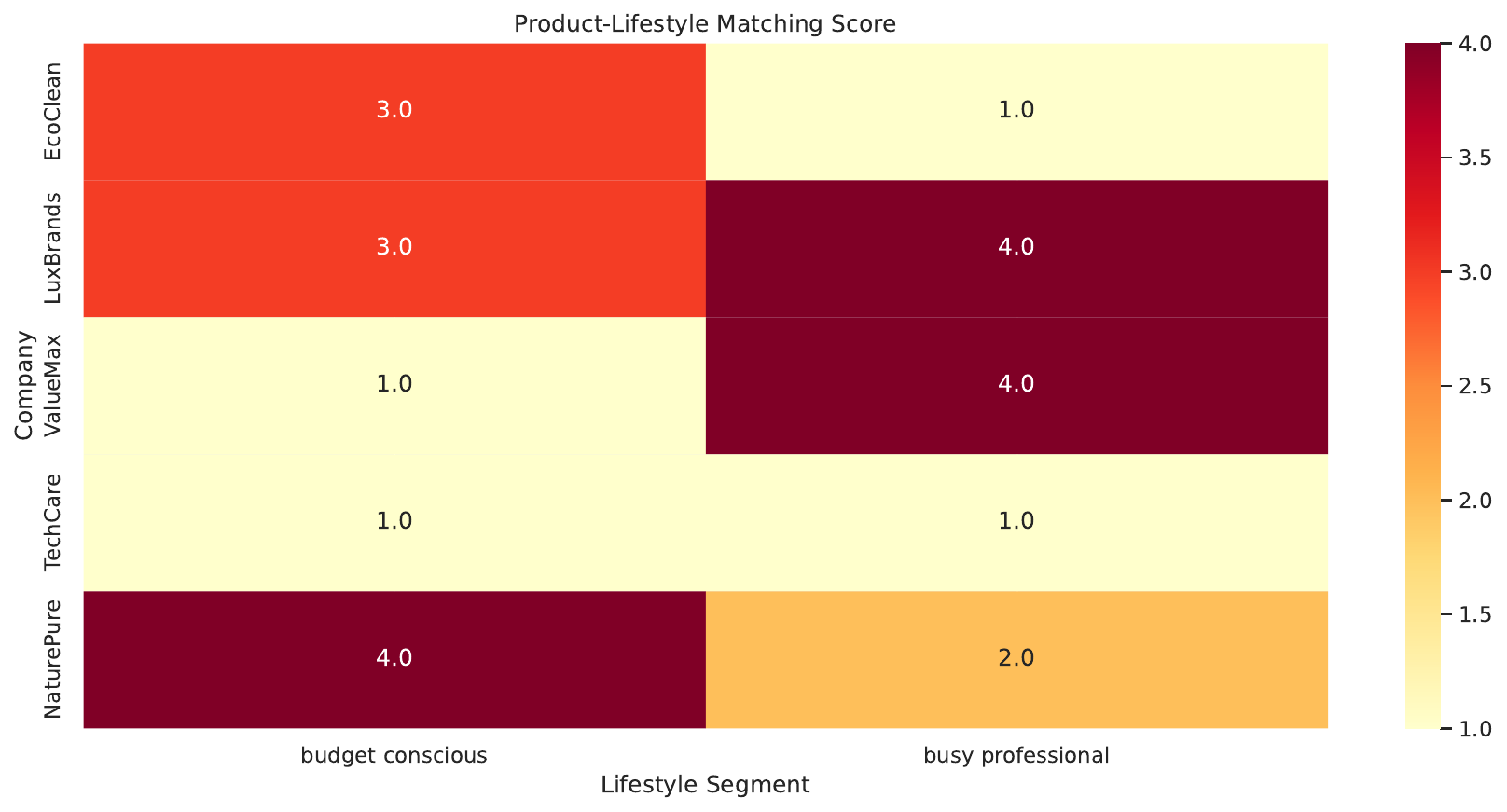}
    \caption{Product-Lifestyle Matching Score Heatmap. This visualization quantifies how well different companies' products match with various lifestyle segments, considering factors like price alignment, shared values, and interests. Higher scores (darker colors) indicate better product-lifestyle fit.}
    \label{fig:lifestyle_match}
\end{figure}

\begin{figure}[htbp]
    \centering
    \includegraphics[width=0.8\textwidth]{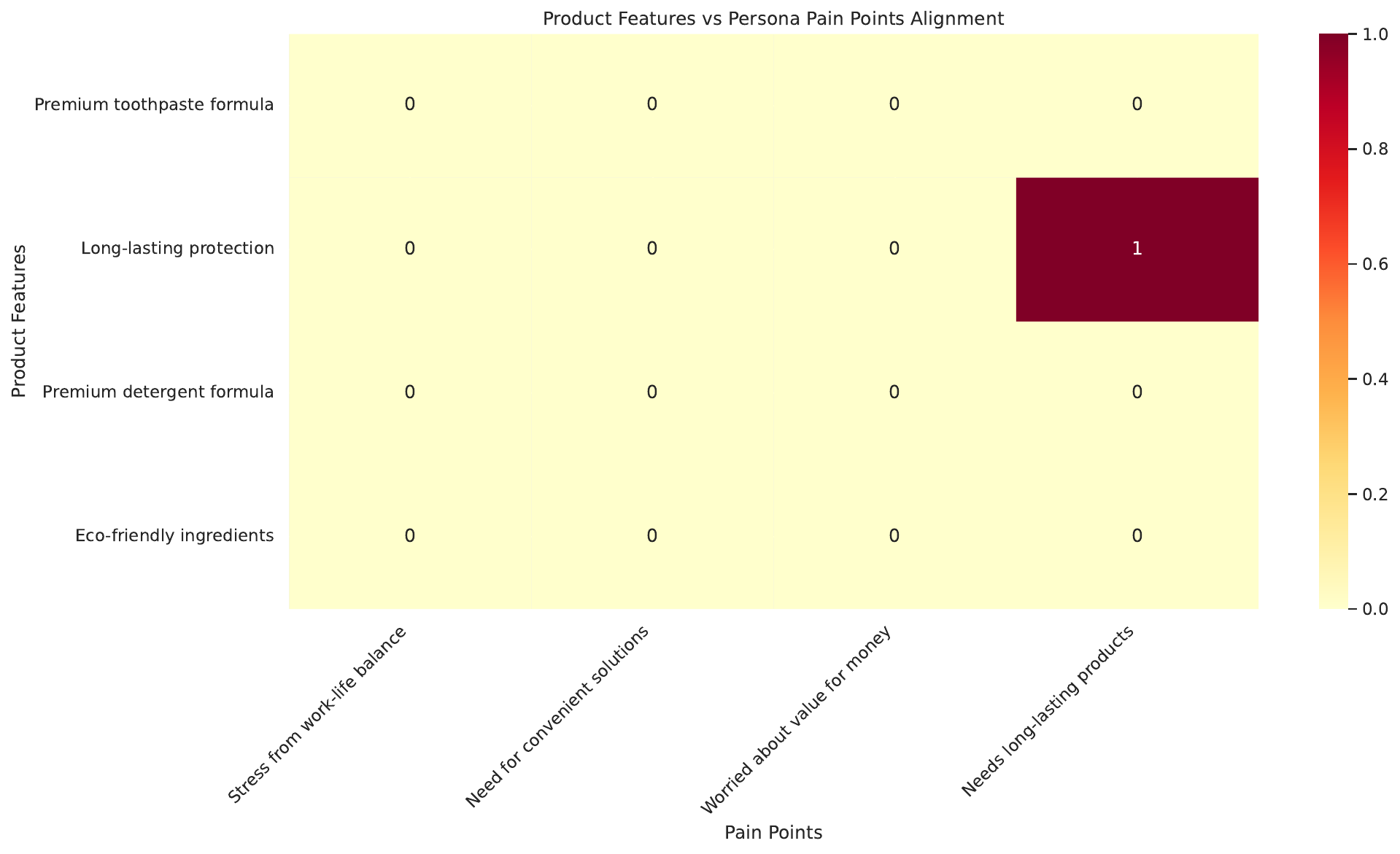}
    \caption{Product Features vs Persona Pain Points Alignment Heatmap. This visualization maps how well current product features address identified customer pain points. The heatmap helps identify gaps in product offerings and opportunities for product development to better meet customer needs.}
    \label{fig:pain_point_map}
\end{figure}

\begin{figure}[htbp]
    \centering
    \includegraphics[width=\textwidth]{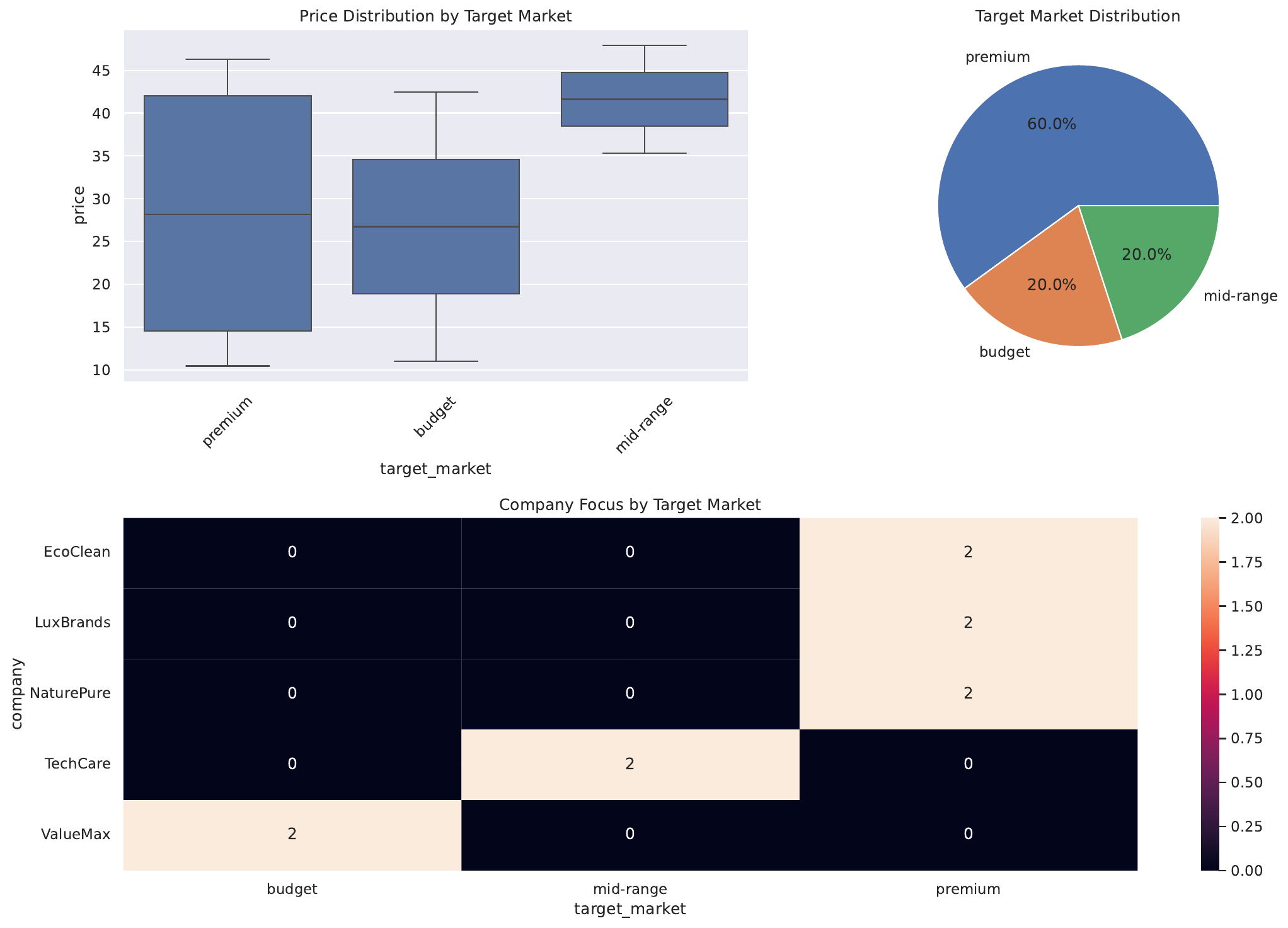}
    \caption{Market Segment Analysis Multi-plot. This comprehensive visualization combines three views: (1) price distribution across target markets, (2) market segment distribution, and (3) company-specific market targeting. Together, these plots provide insights into market segmentation strategies and competitive positioning.}
    \label{fig:market_segment}
\end{figure}

Table~\ref{tab:ad-comparison} illustrates personalized advertisements for EcoClean Detergent across three synthetic personas: Eco-Conscious Luxury, Active Lifestyle, and Premium Innovation consumers. The table breaks down messaging components (headlines, subheadlines, main body, calls-to-action, and targeted benefits) demonstrating how the same product is marketed differently to each persona. While Eco-Conscious Luxury messaging emphasizes sustainability and premium quality, Active Lifestyle focuses on affordability and performance, and Premium Innovation stresses advanced technology with environmental responsibility. This comparison showcases how synthetic persona generation informs targeted advertising strategies. The suite of visualizations (Figures~\ref{fig:price_distribution}, \ref{fig:target_market}, \ref{fig:interest_correlation}, \ref{fig:value_proposition}, \ref{fig:price_vs_income}, \ref{fig:price_segments}, \ref{fig:value_interest}, \ref{fig:lifestyle_match}, \ref{fig:pain_point_map}, and \ref{fig:market_segment}) provides comprehensive insights into the effectiveness of synthetic data generation for advertising analysis. Figure~\ref{fig:price_distribution} reveals distinct pricing patterns across synthetic product categories, while Figure~\ref{fig:target_market} demonstrates the simulated market segment distribution, showing how different products are positioned across market tiers. The correlation between synthetic products and generated customer personas is examined in Figure~\ref{fig:interest_correlation}, which highlights varying degrees of interest alignment across different synthetic customer segments. The distribution of company values is analyzed in Figure~\ref{fig:value_proposition}, providing insights into how different brand values can be positioned in marketing strategies. Figures~\ref{fig:price_vs_income} and~\ref{fig:price_segments} together offer a detailed view of pricing strategies relative to synthetic customer income levels, revealing potential market opportunities and gaps in the simulated environment. The alignment between company values and customer interests is visualized in Figure~\ref{fig:value_interest}, while Figure~\ref{fig:lifestyle_match} quantifies product-lifestyle fit across different synthetic customer segments. Figure~\ref{fig:pain_point_map} maps how effectively generated product features address synthetic customer pain points, identifying areas for potential product development and marketing focus. Finally, Figure~\ref{fig:market_segment} provides a comprehensive view of simulated market segmentation strategies, combining price distribution, market segment distribution, and company-specific targeting approaches. Together, these visualizations demonstrate the power of synthetic data in understanding market dynamics, customer preferences, and advertising opportunities, offering valuable insights for real-world marketing strategies without the need for extensive real-world data collection or testing.

\end{document}

%% file: math_commands.tex
\usepackage{amsmath,amsfonts,bm}

\def\eqref#1{equation~\ref{#1}}

\def\1{\bm{1}}

\DeclareMathAlphabet{\mathsfit}{\encodingdefault}{\sfdefault}{m}{sl}
\SetMathAlphabet{\mathsfit}{bold}{\encodingdefault}{\sfdefault}{bx}{n}

